%% file: acl_latex_arxiv.tex
\pdfoutput=1

\documentclass[11pt]{article}

\usepackage[final]{acl}

\usepackage{times}
\usepackage{latexsym}
\usepackage{caption}
\usepackage[labelformat=simple]{subcaption}
\usepackage{amsmath}
\usepackage{utfsym}
\usepackage{amssymb}
\usepackage[T1]{fontenc}
\usepackage[utf8]{inputenc}
\usepackage{microtype}
\usepackage{inconsolata}
\usepackage{graphicx}
\usepackage{xcolor}
\usepackage{multirow}
\usepackage{todonotes}
\usepackage {booktabs}
\usepackage{pifont}
\usepackage[inline]{enumitem}
\usepackage{breakurl}
\usepackage{adjustbox}
\usepackage{listings}
\usepackage{array}

\lstdefinelanguage{Jinja2}{
    morekeywords={for, endfor, if, endif},
    sensitive=true,
    morestring=[b]",
    morestring=[b]'
}

\newcolumntype{?}{!{\vrule width 1pt}}

\newcommand{\customfootnotetext}[2]{
  \renewcommand{\thefootnote}{#1}
  \footnotetext[0]{#2}}

\title{Rubrik's Cube: \\ Testing a New Rubric for Evaluating Explanations on the CUBE dataset \\\vspace{0.3cm} \includegraphics[width=1cm]{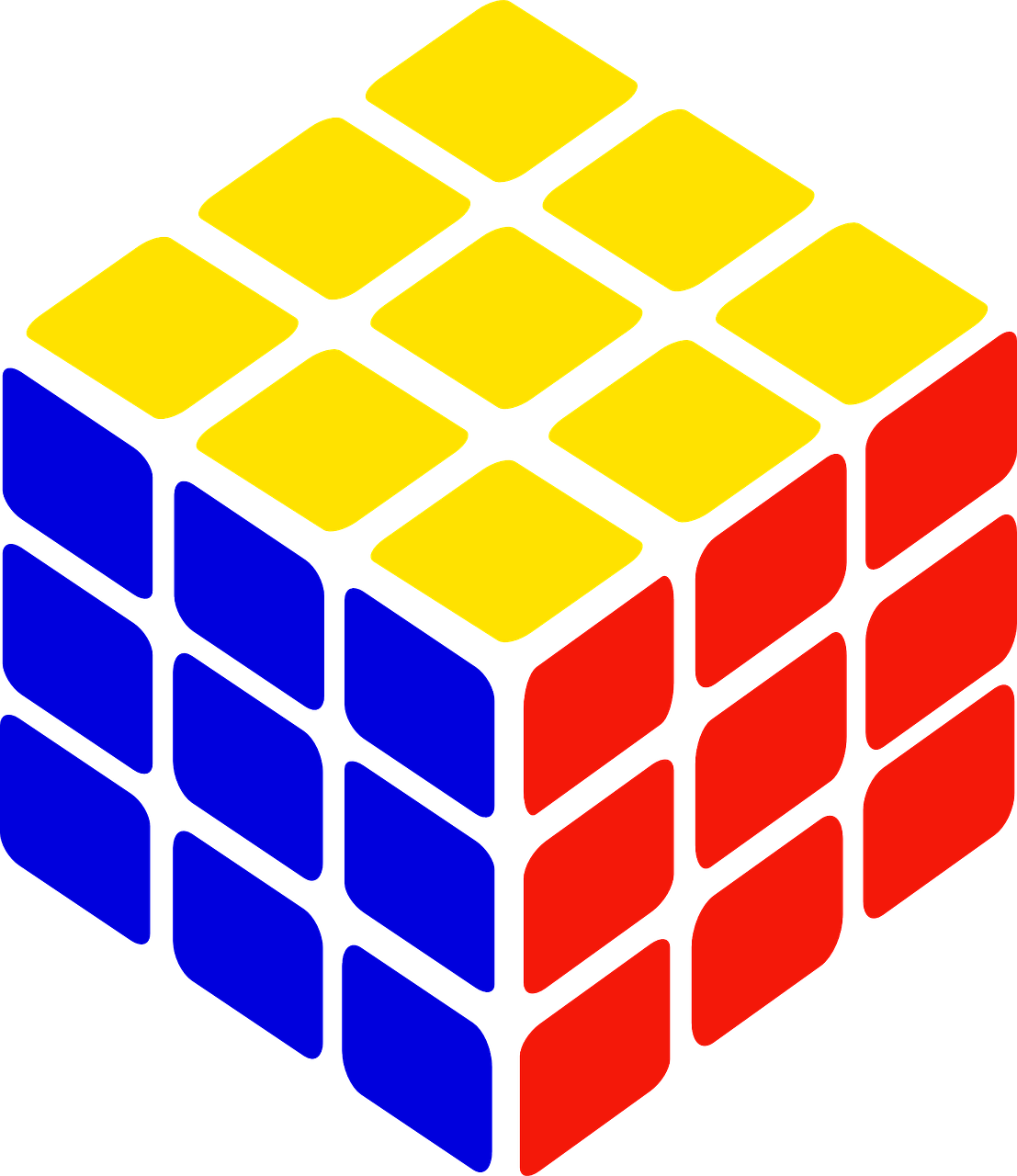}}

\author{
 \textbf{Diana Galvan-Sosa\textsuperscript{1,$\bigstar$}},
 \textbf{Gabrielle Gaudeau\textsuperscript{1,$\bigstar$}},
 \textbf{Pride Kavumba\textsuperscript{2,†}},
 \textbf{Yunmeng Li\textsuperscript{3}},
\\
 \textbf{Hongyi Gu\textsuperscript{5}},
 \textbf{Zheng Yuan\textsuperscript{6}},
 \textbf{Keisuke Sakaguchi\textsuperscript{3, 4}},
 \textbf{Paula Buttery\textsuperscript{1}}
\\
 \small\textsuperscript{1}ALTA Institute, Computer Laboratory, University of Cambridge,
 \textsuperscript{2}SB Intuitions,
 \textsuperscript{3}Tohoku University, \\
 \small\textsuperscript{4}RIKEN,
 \textsuperscript{5}NetMind.AI,
 \textsuperscript{6}The University of Sheffield
}

\begin{document}
\maketitle
\begin{abstract}

The performance and usability of Large-Language Models (LLMs) are driving their use in explanation generation tasks. 
However, despite their widespread adoption, LLM explanations have been found to be unreliable, making it difficult for users to distinguish good from bad explanations.
To address this issue, we present Rubrik’s CUBE--an education-inspired rubric and a dataset of 26k explanations, written and later quality-annotated using the rubric by both humans and six open- and closed-source LLMs.
The CUBE dataset focuses on two reasoning and two language tasks, providing the necessary diversity for us to effectively test our proposed rubric.
Using Rubrik, we find that explanations are influenced by both task and perceived difficulty.
Low quality stems primarily from a lack of conciseness in LLM-generated explanations, rather than cohesion and word choice. 
The full dataset, rubric, and code are available at \url{https://github.com/RubriksCube/rubriks_cube}.

\end{abstract}

\section{Introduction}

\customfootnotetext{$\bigstar$}{Equal contribution, contact: \{dg693,gjg34\}@cam.ac.uk.}

\begingroup  
  \renewcommand\thefootnote{†}
  \footnotetext{Most of the author’s contribution was performed while at LegalOn Technologies.}
\endgroup

\begin{figure}[t!]
\small
\centering
    \includegraphics[width=\columnwidth]{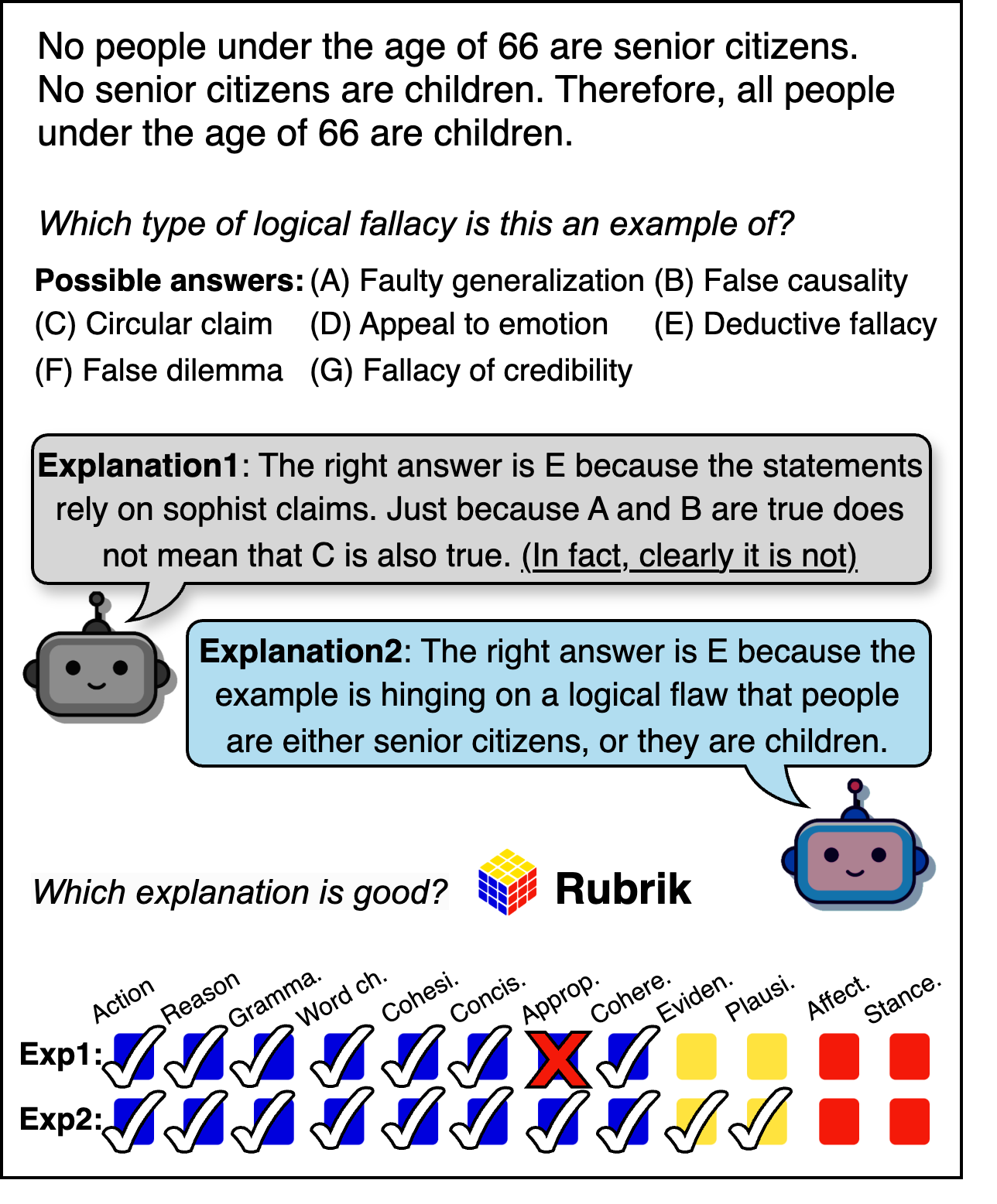}
    \caption{Two explanations of varying quality (in terms of \textsc{Appropriateness} and the provision of \textsc{Evidence}) which present the logic behind an answer choice.}
    \label{fig:examples}
\end{figure}

Explanations play a crucial role in the process of understanding why a decision was made. 
But, as illustrated in Figure~\ref{fig:examples}, there exist many ways of expressing the rationale behind a choice.
Large-Language Models (LLMs), with their inherent capacity for generating very different outputs given the same query, provide a compelling example of this phenomenon.
In fact, these models are increasingly being used in applications which expect a detailed breakdown explaining why a decision was made (e.g., automated scoring, question generation, problem resolution; ~\citeauthor{garcia2024review}, \citeyear{garcia2024review}). 

Unfortunately, LLM-generated explanations generally fall short of user expectations due to their unreliability~\citep{kim2024understanding}. 
Indeed, they are known to occasionally hallucinate, produce incorrect or misleading information, and struggle to back
up their responses to queries, highlighting an overall deficiency in their reasoning capabilities~\citep{huang-chang-2023-towards, saxena2024evaluating}. As noted by~\citet{zhang2023language}, these issues remain unaddressed, even by prompting strategies like \textit{``Let's think step by step.''}
As a result, LLM-generated explanations lack
transparency, and are a source of misinformation and limited knowledge~\citep{sallam2023chatgpt, kabir_is_2024}. Consequently, the challenge has shifted from generating text to assessing its quality, a difficulty that has led some sites to temporarily ban the use of any generative AI (GenAI)\footnote{See StackOverflow's policy on the use of ChatGPT and other LLMs: \url{https://meta.stackoverflow.com/questions/421831/policy-generative-ai-e-g-chatgpt-is-banned}}.

The most common practice in GenAI to determine the quality of a text is to rely on human evaluators.
However, because such evaluators typically lack specific training, the exact evaluation criteria are left to their discretion~\citep{clark2021all}. Inspired by the use of rubrics in education for the qualitative evaluation of complex and subjective tasks like essay writing (e.g., the IELTS writing rubric; \citeauthor{arnold_ielts_2023}, \citeyear{arnold_ielts_2023}), we design our very own rubric following~\citet{dawson2017assessment}'s best practices. 
In doing so, we align ourselves with the human-grounded evaluation proposed by~\citet{doshi-velez_towards_2017}, which identifies and evaluates the ``general notions'' of the quality of an explanation without having a specific end goal.

We thus introduce Rubrik's CUBE\footnote{Short for \textbf{C}ommonsense reasoning, \textbf{U}sual logical fallacies, \textbf{B}asic reading comprehension, and \textbf{E}ssay scoring.}, a task-independent rubric and a dataset to help evaluate the quality of LLM-generated explanations. 
Rubrik identifies the core components and features of a \textit{good} explanation, differentiated by explanation type; CUBE contains 26k explanations drawn from instances of four distinct tasks, generated by both humans and a set of open- (\texttt{Command R+}, \texttt{Gemma 2}, \texttt{Llama 3.1}, \texttt{Mixtral})
and closed-source (\texttt{GPT-4o}, \texttt{Claude Sonnet 3.5}) models. 
We additionally include two custom agreement metrics that account for the hierarchical and nested nature of our rubric.
Rubrik enables valuable insights on output quality, allowing us to identify distinct patterns in the explanations of all annotators. 
We observe that the explanation type depends on the task and its perceived difficulty. 
Specifically, our rubric revealed that low-quality LLM explanations are primarily due to not being concise and only rarely because of word choice or cohesion.

\begin{table*}
  \centering
  \small
  \begin{tabular}{l l p{5.5cm} c c}
    \toprule
      &  & \centering{\textsc{\textbf{Components}}} & \multicolumn{2}{c}{\textsc{\textbf{Dimensions}}} \\\vspace{0.3cm}
     & \hspace{1.5em}\raisebox{-0.2em}{\includegraphics[width=1.5em]{images/cube.png}} & \centering{necessary parts of an explanation} & \multicolumn{2}{p{5.5cm}}{\centering{necessary qualities of a \textit{good} explanation}} \\
    \multicolumn{2}{c}{\textbf{Typology of Explanations}} & & \textbf{Language} & \textbf{Content} \\\midrule
    
    \multirow{3}{*}{Typ1.} & \multirow{3}{*}{\textsc{\textbf{\includegraphics[width=1em]{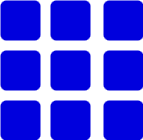} Commentary}}} & 1.a) \centering{Action} &  Grammaticality & Conciseness \\
    & & 1.b) \centering{Reason} & Word Choice & Appropriateness \\
    & & & Cohesion & Coherence \\
    \midrule
    Typ2. & \textsc{\textbf{\includegraphics[width=1em]{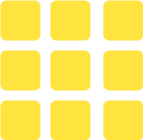} Justification}} & 2.a) \centering{Evidence} & & Plausibility \\\midrule
    Typ3. & \textsc{\textbf{\includegraphics[width=1em]{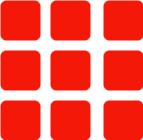} Argument}}  & 3.a) \centering{Affective appeal(s) and Qualifier(s)} & & Stance Clarity \\
    \bottomrule  
  \end{tabular}
  \caption{Overview of our evaluation rubric which identifies three hierarchical types of explanations, their necessary parts (\textsc{Components}), and the features that distinguish the \textit{good} from the \textit{bad} ones (\textsc{Dimensions}).}
  \label{table:rubric_final}
\end{table*}

\section{Background}
\label{sec:background}

We summarise different bodies of literature on the nature and qualities of explanations which, alongside insights from the education assessment literature, informed the design of our rubric.

\subsection{Cognitive Science and Social Sciences}
\label{sec:cogsci_background}

There is an open discussion in philosophy and other social sciences like psychology about what an explanation is and what makes the best explanation~\citep{doshi-velez_towards_2017,gilpin2018explaining,miller2019explanation}.
From the psychology and cognitive science perspective, an explanation is something ubiquitous, diverse, and fundamental to humans' sense of understanding.
They come in a variety of forms and formats and are used for a variety of purposes~\citep{keil2006explanation}, including: \begin{enumerate*}[label=(\arabic*)]
    \item understanding a \textit{decision process} \label{item:goal1}
    \item understanding and predicting an \textit{unexpected event}\label{item:goal2}, and
    \item filling a gap in knowledge (i.e., \textit{learning}).\label{item:goal3}
\end{enumerate*} It follows that a \textit{good} explanation is inherently related to its purpose, which some suggest is shaped by what is being asked~\citep{bromberger1992we}.
In particular, authors like~\citet{lombrozo2006structure} and~\citet{miller2019explanation} argue that an explanation's relation to cognition comes from an attempt to answer a \textit{why-question}.
Miller investigated the criteria that people use to evaluate explanations, finding that the most important are: \textsc{probability}, \textsc{simplicity}, \textsc{generalise}, and \textsc{coherence} with prior beliefs.
The truth of \textsc{likelihood} is also identified as an important criterion.
However, Miller notes that an explanation that includes this attribute is not always the \textit{best} explanation.

\subsection{Explainable AI} 

In the context of Explainable AI (XAI) and Machine Learning (ML) interpretability, an explanation should be able to reflect the internal decision process of a system.
\textit{Introspective} systems output this kind of explanation, while \textit{justification} systems output evidence supporting a decision \cite{park2018multimodal}. 
The most studied properties of explanation systems include \textsc{fidelity}, \textsc{stability}, \textsc{comprehensibility}, \textsc{generalisability} and \textsc{consistency}~\citep{fel2022good}.
According to~\citet{Wiegreffe2021TeachMT}, explanations are implicitly or explicitly designed to answer the why-question \textit{``why is <input> assigned <label>''}.
They identified \textsc{highlights} (subsets of the input elements that explain a prediction) as one type of explanation in the Explainable NLP (EXNLP) literature, where \textsc{compactness}, \textsc{sufficiency} and \textsc{comprehensiveness} are the main attributes.

\subsection{Natural Language Generation}
In an attempt to find a consensus about how human evaluations of generated text should be designed and reported,~\citet{howcroft-etal-2020-twenty} examined twenty years of NLG papers that reported some form of human evaluation. 
Some of the most common criteria used to assess quality include \textsc{fluency}, \textsc{appropriateness} and \textsc{clarity}.

The Multidimensional Quality Metrics (MQM) framework~\cite{burchardt-2013-multidimensional, mariana2014multidimensional, freitag-etal-2021-experts} has been widely applied to machine translation studies in recent years.\footnote{An updated version (MQM 2.0) is available from \href{https://themqm.org/the-mqm-full-typology/}{https://themqm.org/the-mqm-full-typology/}.} This hierarchical typology of quality issues provides a detailed and flexible approach for evaluating translation tasks. 
It has been applied to different domains of machine translation, such as literary translation~\citep{karpinska-iyyer-2023-large} and chat translation~\citep{li-etal-2025-mqm}.
It could be used as a metric for human evaluators to evaluate translation models, and could also be used to prompt models as evaluators~\citep{park-pado-2024-multi, li-etal-2025-mqm}.
It identifies seven high-level error types (namely \textsc{terminology}, \textsc{accuracy}, \textsc{linguistic conventions}, \textsc{style}, \textsc{locale conventions}, \textsc{audience appropriateness}, and \textsc{design and markup}), which can be broken into multiple subtypes to enable fine-grained assessment.
In the design of our rubric, we similarly arranged the significant features of explanations hierarchically to allow for both coarse and granular evaluations.

\subsection{Education}
Education, and specifically science education, has long focused on teaching students how to construct explanations, and assessing them (e.g., \citeauthor{sandoval_conceptual_2003}, \citeyear{sandoval_conceptual_2003}; \citeauthor{mcneill_supporting_2006}, \citeyear{mcneill_supporting_2006}; \citeauthor{mcneill_scientific_2008}, \citeyear{mcneill_scientific_2008}; \citeauthor{zangori_fostering_2013}, \citeyear{zangori_fostering_2013}). For them, explanations ``make sense of a phenomenon based on other scientific facts''~\citep{ohlsson_generating_2002}. They should begin with a statement of the \textit{explanandum} (i.e., the phenomenon to be explained). Then, what makes a \textit{good} explanation differs is ``explanatory adequacy''~\citep{brigandt_why_2016} which consists in providing an understanding of how or why a phenomenon occurs~\citep{chin_learning_2000}.

In practice, assessing explanations is difficult~\citep{berland_for_2012}, so teachers generally rely on rubrics, like the one proposed by~\citet{mcneill_middle_2007}, which provide clear, consistent, and objective sets of criteria for evaluation. More generally, rubrics are firmly established evaluation tools in written assessment and widely advocated in books by~\citet{walvoord_effective_1998, huba_learner-centered_2000, dunn_student_2003, stevens_introduction_2004, freeman_planning_2016}. Unfortunately, these practices are not currently being used beyond education, and no equivalent rubric exists for evaluating LLM explanations on a variety of tasks (beyond scientific explaining). To address this gap, we propose to draw on this literature to come up with our very own rubric.

\section{A Systematic Quality Assessment Framework}

This section introduces our proposed assessment framework in three parts. 
First, we detail the design decisions taken to develop the rubric, drawing upon the key principles outlined by~\citet{dawson2017assessment}.
Second, we provide a comprehensive overview of the rubric itself, outlining its key elements and their hierarchical relationships.
Finally, we provide practical guidance on how to effectively use the rubric for the task of explanation quality assessment.

\subsection{Designing an Assessment Rubric}

Recognising that the foundation of an effective evaluation lies in its instrument, we carefully considered the design elements suggested by~\citet{dawson2017assessment}.
A key advantage of adhering to their framework is the streamlined design process and the enhanced transparency of the resulting rubric, facilitating easier comparisons with other instruments.

\begin{table*}
  \centering
  \small

  \begin{tabular}{p{8.5cm}p{5.5cm}}
    \toprule
    \textbf{Design element} & \textbf{Decision} \\\midrule
    \textit{Specificity}: the particular object of assessment & Assess the quality of explanations. \\
    \textit{Secrecy}: who the rubric is shared with, and when it is shared & It should be secret to the annotators. It is only shared with the evaluators. \\
    \textit{Exemplars}: work samples provided to illustrate quality & Examples of acceptable and not acceptable instances. \\
    \textit{Scoring strategy}: procedures used to arrive at marks and grades & A series of binary judgments (yes/no) all amounting to a binary decision (good/bad). \\
    \textit{Evaluative criteria}: overall attributes required of the explanation & Components and dimensions. \\
    \textit{Quality levels}: the number and type of levels of quality & Two quality levels (\usym{1F60A} good or \usym{1F641} bad). \\
    \textit{Quality definition}: explanations of attributes of different levels of quality & Motivated by different bodies of literature (social sciences, XAI, and NLG). \\
    \textit{Judgment complexity}: the evaluative expertise required of users of the rubric & Should be simple enough for \textbf{anyone} to use. \\
    \textit{Users and uses}: who makes use of the rubric, and to what end & Evaluators use for summative assessment. \\
    \textit{Creators}: the designers of the rubric & NLP researchers. \\
    \bottomrule
  \end{tabular}
  \caption{Summary of the design decisions taken to develop our proposed rubric. The design elements are those suggested by~\citet{dawson2017assessment}. The ``annotators'' are the humans or LLMs who write the explanations.}
  \label{table:rubric_design}
\end{table*}

\subsection{A Task-Agnostic Quality Rubric}
\label{subsec:eval_criteria}

A fundamental assumption underlying this work is that it is possible to account for the diverse nature of explanations (which can serve a wide range of goals as highlighted in Section~\ref{sec:cogsci_background}) whilst also being able to recognise common
features that generally characterise them.
Through Section~\ref{sec:background}, we showed that different bodies of literature identify shared attributes of a \textit{good} explanation. Using these attributes, our proposed rubric (henceforth Rubrik) classifies explanations into three goal-driven types.
Each type is defined by the presence of specific \textsc{Components}. The typology is hierarchical and nested, with subsequent types inheriting the \textsc{Components} of preceding types and adding to them.
Each explanation type also comes with its own set of attributes called \textsc{Dimensions}: together, these capture the quality of an explanation of that type (\textit{good} or \textit{bad}). Much like \textsc{Components}, \textsc{Dimensions} are inherited and accumulate across the type hierarchy.
This ``building block''-like structure provides a robust framework for understanding how the form and features of explanations evolve alongside the distinct goals of each type. Table~\ref{table:rubric_final} presents an overview of our proposed rubric (see Table ~\ref{table:rubric_extended} in Appendix \ref{app:rubric_creation} for the full-sized, illustrated rubric) and Table~\ref{table:rubric_design} shows the design considerations and choices we made in developing it.

\subsubsection{Components}
\label{subsubsec:components}

The three hierarchical and nested explanation types in Rubrik are: \textsc{Commentary}, \textsc{Justification}, and \textsc{Argument}. 
The \textsc{Commentary} is the foundational level and consists of two \textsc{Components}: an \textsc{Action} and a \textsc{Reason}. 
The \textsc{Justification} extends this base by incorporating an additional \textsc{Component}: an \textsc{Evidence}. 
Finally, the \textsc{Argument} includes the elements of both the \textsc{Commentary} and the \textsc{Justification}, as well as an additional unique element: the \textsc{Affective appeal(s) and Qualifier(s)}. 
This progression, where each higher-level type nests the elements of the lower-level ones, results in an increasing richness of information.
Providing an \textit{understanding} of a decision process is the central goal of a \textsc{commentary} and a \textsc{justification}. 
An \textsc{argument}, while also considering the same goal, is more focused on \textit{persuasion}.
Formally, \textsc{Commentary} $\subseteq$ \textsc{Justification} $\subseteq$ \textsc{Argument}.
See Appendix \ref{app:components_creation} for a more in-depth understanding of the reasoning that led to the definition of types and components.

\subsubsection{Dimensions}
\label{subsubsec:dimensions}

\textsc{Components} provide the necessary structural elements of different types of explanations; \textsc{dimensions} are their requisite qualities. This distinction ensures that our rubric accounts for both what is being said (through the \textsc{components}) and how well it is communicated (through the \textsc{dimensions}). 

The eight \textsc{dimensions} shown in Table~\ref{table:rubric_final} were chosen from a wider set of explanation qualities (see Table~\ref{table:rubric_dim} in Appendix~\ref{app:dimensions_creation}) that have been studied, annotated or evaluated in the bodies of literature introduced in Section~\ref{sec:background}. 
We filtered out those that were too task-specific for our goal of creating a general-purpose rubric (e.g., \textsc{Fidelity}, \textsc{Consistency}, \textsc{Transparency} and \textsc{Interpretability} specifically focus on the internal workings of AI models) or too vague (for e.g., \textsc{Clarity}; see Section \ref{app:clarity}). 
The eight remaining \textsc{dimensions} were then put in one of two categories. \textbf{Language} assesses whether the explanation is well-formed; \textbf{Content} evaluates the ideas expressed by the explanation. 
This design choice was motivated by the fact that LLMs sometimes produce text that is only \textit{good} on the surface but factually incorrect, inappropriate, or misleading~\citep{huang_survey_2025}. 
We describe our process in more detail in Appendix \ref{app:dimensions_creation}.

These \textsc{dimensions} were then related to the \textsc{components} and explanation types introduced in the previous section. \textsc{Action} and \textsc{Reason} are pre-requisites for a \textsc{commentary} to be considered complete; but for it to be \textit{good}, we must enforce certain linguistic requirements: it needs to be grammatical, cohesive, and use \textit{context}-appropriate language. On the other hand, its content should be coherent and concise and match the expectations imposed by the defined \textit{context}.
Further, a \textsc{justification} is contingent on the presence of \textsc{evidence}. Ensuring it is plausible and consistent with human reasoning is a further requirement for a \textit{good} \textsc{justification}. Finally, the presence of argumentative markers generally betrays the explainer's intent to persuade the audience of their \textit{stance} (i.e., their personal feelings towards the task). Whether this stance is clearly and unambiguously conveyed distinguishes a \textit{good} from a \textit{bad} \textsc{argument}.

\subsection{Scoring Strategy}
\label{subsec:scoring_strategy}
To use Rubrik, evaluators must first establish the \textit{context} of the explanations:
\begin{itemize}
    \item What is the task? In our case, we will be looking at two reasoning and two language tasks (Section \ref{subsec:data_collection}).
    \item Who is the target audience? In our case, NLP researchers (i.e., formal academic setting)
    \item What is their intended goal?
\end{itemize}

Once the \textit{context} is defined, we can proceed with the evaluation. Given an explanation, the outcome of an evaluation with Rubrik is a \textbf{Type} for that explanation (\textsc{None}, \textsc{Commentary}, \textsc{Justification}, \textsc{Argument}) and a related \textbf{Quality label} (\usym{1F60A} \textit{good} or \usym{1F641} \textit{bad}). The evaluation process follows our hierarchical typology: starting from the foundational level--the \textsc{Commentary}--and going all the way to the \textsc{Argument}. We describe this process in detail below:

\begin{itemize}
    \item First, we start by checking whether the two \textsc{Components} of the \textsc{Commentary} (namely \textsc{Action} and \textsc{Reason}) are present (\ding{51}) or absent (\ding{55}) in the explanation. If either \textsc{Component} is missing (\ding{55}), then the explanation is incomplete and classified as \textsc{None}, and the evaluation ends there. If, on the other hand, both are present (\ding{51}), then the explanation's \textbf{Type} is at least a \textsc{Commentary}. 
    \item Next, we check whether the explanation satisfies (\ding{51}) each of the \textsc{Commentary}’s six \textsc{Dimensions} or not (\ding{55}). If the explanation fails to meet any of these (\ding{55}), then the explanation is a \usym{1F641} \textit{bad} \textsc{Commentary} and the evaluation ends there. If however, \underline{all} six \textsc{Dimensions} are satisfied (\ding{51}), then the explanation is at least a \usym{1F60A} \textit{good} \textsc{Commentary}. 
    \item Continue this procedure with the \textsc{Components} and \textsc{Dimensions} of the \textsc{Justification}. Specifically, if the explanation does not have \textsc{Evidence} (\ding{55}), then the explanation is only a \usym{1F60A} \textit{good} \textsc{Commentary} and the evaluation ends there. If it does (\ding{51}), then it is at least a \textsc{Justification}. Whether it is a \usym{1F60A} \textit{good} or  \usym{1F641} \textit{bad} \textsc{Justification} will depend on whether the \textsc{Evidence} is judged as \textsc{Plausible} (\ding{51}) or not (\ding{55}). If it is the latter, then the evaluation ends there; otherwise, the explanation is at least a \usym{1F60A} \textit{good} \textsc{Justification}. 
    \item Repeat this process with the \textsc{Argument}’s \textsc{Component} and \textsc{Dimension}. 
\end{itemize}

Notice that for each explanation type, we performed two validation steps: (1) Structure validation (determined by the \textsc{components}) and (2) Attribute validation (determined by the \textsc{dimensions}). At each step, the evaluator makes a series of binary judgements based on the presence (\ding{51}) or absence (\ding{55}) of \textsc{Components}, and whether \textsc{Dimensions} are satisfied (\ding{51}) or not (\ding{55}), using the definitions and examples included in the full rubric (Table~\ref{table:rubric_extended}) as reference.

\begin{table*}[h]
\centering
\small
\resizebox{\textwidth}{!}{
\begin{tabular}{l?llll|lllll|l?lll|llll|l}
\toprule
 &  & \multicolumn{3}{c}{\bf Single annotations} &  & \multicolumn{4}{c}{\bf Joint annotations} &  & \multicolumn{3}{c}{\bf Single evaluations} & \multicolumn{4}{c}{\bf Joint evaluations} &  \\
\rule{0cm}{0.3cm}  & & Inst. & LLM  & Total & Inst. & H & LLM & Total & Total & Total & Inst. & E & LLM & Inst. & E & H & LLM & Total \\\hline
\textbf{T1} & \underline{\underline{1000}} & \underline{890} & 6 & 5340 & \underline{110$^{\ddagger}$} & 4 & 6 & 10 & 1100 & 6440 & 90$^{\ddagger}$ & 900 & 1 & 20$^{\ddagger}$ & 200 & 2 & 1 & \\
\textbf{T2} & \underline{\underline{1000}} & \underline{890} & 6 & 5340 & \underline{110$^{\ddagger}$} & 4 & 6 & 10 & 1100 & 6440 & 90$^{\ddagger}$ & 900 & 1 & 20$^{\ddagger}$ & 200 & 2 & 1 & \\
\textbf{T3} & \underline{\underline{1000}} & \underline{890} & 6 & 5340 & \underline{110$^{\ddagger}$} & 7 & 6 & 13 & 1430 & 6770 & 90$^{\ddagger}$ & 1170 & 1 & 20$^{\ddagger}$ & 260 & 2 & 1 & \\
\textbf{T4} & \underline{\underline{1000}} & \underline{890} & 6 & 5340 & \underline{110$^{\ddagger}$} & 7 & 6 & 13 & 1430 & 6770 & 90$^{\ddagger}$ & 1170 & 1 & 20$^{\ddagger}$ & 260 & 2 & 1 & \\[0.1cm] \hline
\rule{0cm}{0.3cm}\textbf{Total} & 4000 & 3560 & & 21360 & 440 & & & & 5060 & \textbf{26420} & 360 & 4140 & & 80 & 920 & & & \textbf{5060} \\
\bottomrule 
\end{tabular}
}
\caption{Instances and explanations (E) in CUBE. Double-underlined numbers represent the initial pool, divided into subsets (single-underlined) based on the annotators assigned. A ($\ddagger$) denotes variations in evaluator assignment.}
\label{table:cube_stats}
\end{table*}

\section{Rubric Validation}
The main motivation behind our proposed rubric is to allow for a more systematic evaluation of an explanation's quality.
In order to determine the effectiveness of our proposal, we designed a validation process aimed at addressing the following question: \textit{Does the rubric effectively discriminate between high-quality and low-quality explanations, while simultaneously providing clear and concise guidance for evaluators?}
Given the absence of existing datasets for explanation assessment, the validation of this rubric required a tailored approach. 
This began with identifying an appropriate source of data, followed by gathering explanations, evaluating them using the rubric with three raters, and finally, measuring the inter-rater reliability to determine the consistency of the rubric's application.
The effectiveness of our rubric was evaluated by measuring the level of inter-rater agreement for each explanation.

\subsection{Data Collection}
\label{subsec:data_collection}
We assume a decision-making scenario involving a set of choices, where one is selected.
Thus, our data collection process required instances from tasks that could be framed as a series of multiple-choice questions (MCQ) with a single correct answer.
To ensure a diverse set of explanations, we chose four different tasks, drawn from reasoning and language assessment.
The reasoning tasks are: (\textbf{T1}) commonsense reasoning and (\textbf{T2}) fallacy detection. 
The language tasks are: (\textbf{T3}) reading comprehension and (\textbf{T4}) essay scoring.
From an initial pool of $1000$ instances from each task, we curated an \textit{annotation set} of $440$ total instances for annotation ($110$ from each dataset). 
A brief description of the datasets follows. 
Detailed selection criteria are described in Appendix~\ref{sec:selection}.

\textbf{Reasoning tasks}. For T1 and T2, we selected instances from the HellaSwag~\citep{zellers-etal-2019-hellaswag} and Logic~\citep{jin_logical_2022} datasets, respectively. 
Each instance in HellaSwag has a \textbf{context} and a set of four \textsc{endings}; the task is to select the most likely follow-up sentence. 
Logic consists of common logical fallacy examples collected from various online educational materials. 

\textbf{Language tasks}. For T3 and T4, we selected instances from RACE~\citep{lai-etal-2017-race} and the Write\&Improve (W\&I)~\citep{bryant_bea-2019_2019} corpus, respectively. RACE consists of a series of passages and questions taken from English exams that evaluate a student's ability in understanding and reasoning. Write\&Improve\footnote{\url{https://writeandimprove.com/}.} is an online web platform that assists English Language Learners with their writing~\citep{yannakoudakis_developing_2018}.
The dataset contains submissions (defined as ``essays'') that were annotated with a coarse CEFR\footnote{\label{foot:cefr} Common European Framework of Reference for Languages~\citep{north_common_2020} levels correspond to language proficiency levels ranging from A1 (elementary) to C2 (complete proficiency) from a second-language learner's perspective.} level (A, B or C) by trained raters. 

\begin{figure*}[t!]
\centering
\small
  \includegraphics[width=\linewidth]{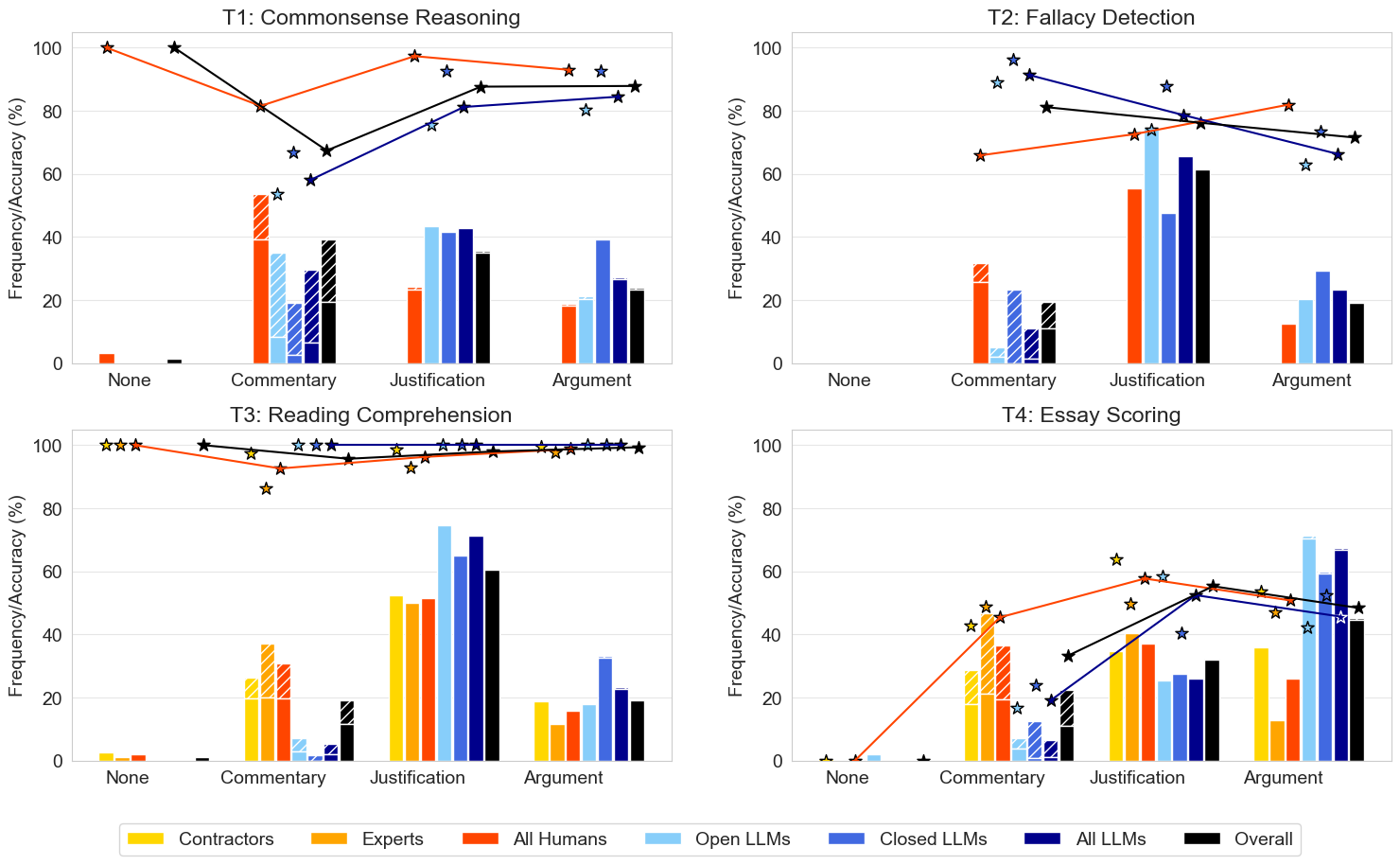}
  \caption {The bar plots show the frequencies (\%) of the different explanation types in each group of annotators as judged by and averaged across the three evaluators (two humans and \texttt{GPT-4o}). The patterned fill indicates the proportion of \textit{bad} explanations of each type; the solid fill shows the proportion of \textit{good} explanations of each type. The scattered stars represent the accuracy (\%) of each group of annotators (i.e., did they select the correct answer out of the possible multiple choices to a question) related to the type of explanation they produced as judged by and averaged across the three evaluators. We plot the accuracy lines for the following three groups: all human annotators, all LLMs, and all annotators (``Overall'').}
\label{fig:quality_freq}
\end{figure*}

\subsubsection{Annotation}
\label{subsubsec:annotation}
Two key decisions shaped the annotation process.
First, we retained all annotations, regardless of the correctness of the chosen answer. 
This decision was driven by the need to explore the explanations associated with correct and incorrect answers, allowing for a more nuanced understanding of the explanatory quality.
Second, human explanations were not treated as the gold standard.  
This allowed for a more objective comparison of human and LLM explanations, avoiding potential bias towards human responses.
Below, we give a brief overview of the annotation process, but we refer the reader to Appendix~\ref{sec:annotation} for more information.

\textbf{Human}. We recruited seven annotators: four general annotators (contractors) and three professionals with experience in language assessment.
They were asked to answer a series of multiple-choice questions and explain their choices. While contractors covered all four tasks, experts focused on the language tasks. This process resulted in $880$ explanations for T1 and T2, $1,540$ for T3 and T4.

\textbf{LLM-based}. We worked with six LLMs, including four open-source: 
\texttt{Llama 3.1}~\citep{dubey_llama3_2024},
\texttt{Gemma 2}
\citep{team_gemma2_2024}, 
\texttt{Mixtral}
\citep{jiang_mixtral_2024}
\texttt{Command R+}, 
\citep{cohere_c4ai_2024} and two closed-source models:
\texttt{GPT-4o}~\citep{openai_gpt4o_2024} and
\texttt{Claude 3.5 Sonnet}~\citep{anthropic_claude_2024}. See Appendix \ref{subsec:llm_annotation} for model versions. Models were prompted using a few-shot setting (see Appendix~\ref{subsubsec:prompt_design}).
Explanations were generated for all instances, yielding a total of $24,000$ explanations.
Table \ref{table:cube_stats} shows a more detailed breakdown of the number of annotations and evaluations.

\subsubsection{Evaluation}
\label{subsubsec:evaluation} 
Data evaluation was performed by two expert evaluators and the same six LLMs on a subset of the \textit{annotation set}: namely, $20$ instances for each task.
Thus, our \textit{evaluation set} has a total of $920$ explanations derived from $80$ instances.
Using two custom agreement metrics, we identified that out of the LLMs, \texttt{GPT-4o} most closely matched our human evaluators. 
As was previously done by \citet{brassard_acorn_2024} and \citet{sottana-etal-2023-evaluation}, we took \texttt{GPT-4o} to act as our third evaluator to enhance the robustness of our analysis, and used it to automatically evaluate the $4,140$ explanations from the remaining $360$ instances of the \textit{annotation set}. 
For details on the preliminary experiment and metrics, see the Appendix ~\ref{sec:agreement_metric}. 

The raters followed the scoring strategy specified in Section \ref{subsec:scoring_strategy}. 
Unlike the human raters, \texttt{GPT-4o} limited its role to validating only the structure and attributes of the explanations. 
In other words, it did not render a final judgment on an explanation's quality. 
This approach mitigated the risk of the model's self-bias (as reported in \citeauthor{panickssery2024llm},\citeyear{panickssery2024llm}); further details on this potential source of bias are provided in Appendix \ref{app:rubric_scoring}.

\begin{figure*}[ht]
    \includegraphics[width=0.9\linewidth]{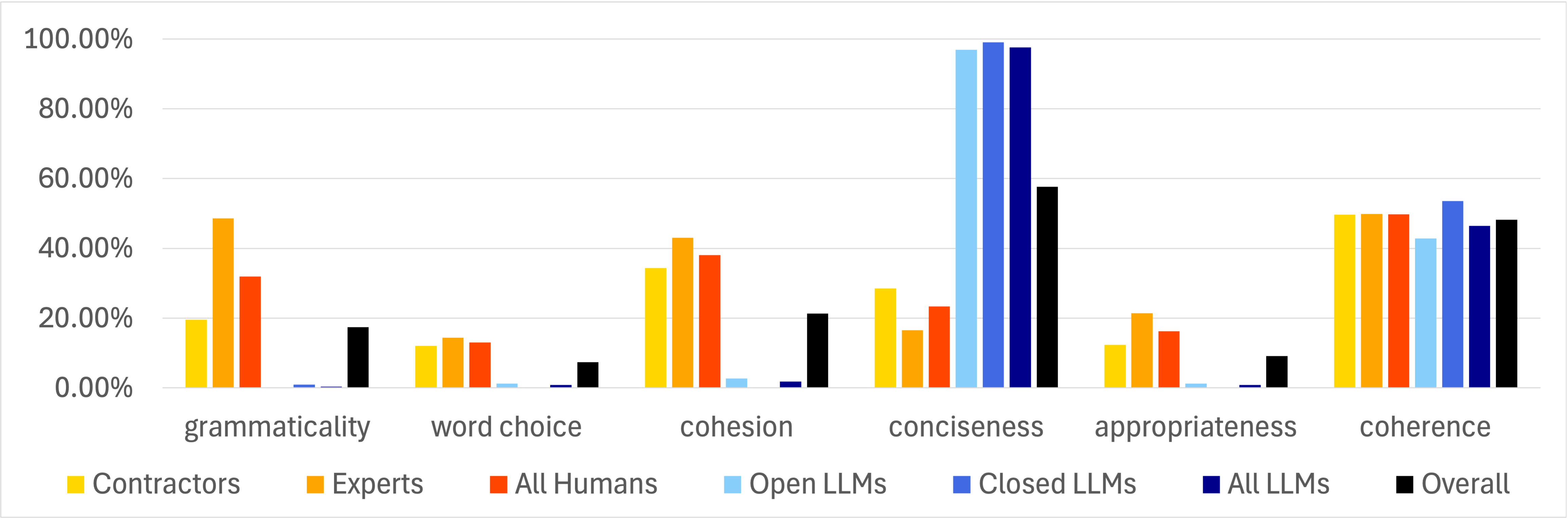}
    \caption{Plot showing the source of \protect\usym{1F641} \textit{bad} \textsc{commentaries} (i.e., which of the \textsc{commentary}'s \textsc{dimensions} was not met \ding{55}) in the \textit{evaluation set}. We average the frequencies across all three evaluators (two humans and \texttt{GPT-4o}).}
    \label{fig:sublabels}
\end{figure*}

\section{Discussion}
\label{sec:discussion}

\begin{table*}
  \centering
  \small
  \begin{tabular}{p{2.5cm}p{12cm}}
    \toprule
    \textbf{Type} & \textbf{Example} \\\midrule
    \textsc{\includegraphics[width=1em]{images/cube_blue_face.png} Commentary} & The right answer is D because the reason given to encourage Luke to eat is subjective and has nothing to do with his taste in food or any potential benefits. It is not a valid reason.\\
    \textsc{\includegraphics[width=1em]{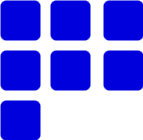} Commentary} & The right answer is D because this statement is trying to make Luke \textbf{eat the sheep's brains with chopped liver and brussel sprouts} by making him feel guilty \textbf{about the poor, starving children in a third world country}. It's an appeal to his emotions, rather than presenting a logical argument.\\\midrule
    \textsc{\includegraphics[width=1em]{images/cube_yellow_face.png} Justification} & The right answer is A because the woman in the video is demonstrating how to make or destroy lipsticks, which is a process that involves using one's mouth and lips. This is the only option that describes an action that would require the use of the mouth and lips.\\
    \textsc{\includegraphics[width=1em]{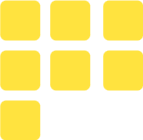} Justification} & The right answer is D because it is the most reasonable answer of the options. A is incorrect because \textbf{demonstrating how to make and destroy lipsticks is an unusual subject to demonstrate: the making of lipsticks in particular is complicated and challenging.} B is incorrect because one's tongue does not fall out when speaking. C is incorrect because you do not describe ordinarily describe women as walking into a classroom with ``both [their] boobs'' as this implies their boobs are separable. D is correct because sucking from a hookah is a fairly ordinary activity.\\\midrule
    \textsc{\includegraphics[width=1em]{images/cube_red_face.png} Argument} & The right answer is B because the essay is written in a somewhat coherent and understandable manner, but it lacks clarity, coherence, and proper sentence structure. The writer's emotions and thoughts are expressed, but the writing is not sophisticated or polished. The essay does not demonstrate a clear understanding of the topic or the ability to express complex ideas. The writer's use of language is simple, and the essay lacks depth and analysis. Therefore, it is best graded as Intermediate (grade B).\\
    \textsc{\includegraphics[width=1em]{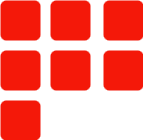} Argument} & The right answer is B because the essay demonstrates a basic understanding of grammar and sentence structure, \textbf{but there are some errors in word choice and sentence construction. The vocabulary used is relatively simple, and the ideas are expressed in a straightforward manner. However, the essay lacks a clear focus and organization, and the conclusion is abrupt.}\\
    \bottomrule
  \end{tabular}
  \caption{Pairs of \textit{good} and \textit{bad} explanations by type. From top to bottom, the source of low-quality is \textsc{conciseness}, \textsc{Plausibility}, and \textsc{Stance Clarity}. }
  \label{table:good_bad_examples}
\end{table*}

\textbf{Agreement.} A key indicator of the utility of Rubrik is the level of agreement observed between the human evaluators who used it.
Standard inter-rater agreement metrics are often inadequate for nested hierarchical data.
Therefore, we designed a custom metric that accounts for both \textit{superlabels} (explanation types) and \textit{sublabels} (\textsc{components} and \textsc{dimensions}) in Rubrik, penalising discrepancies based on the difference in hierarchical level.  
Using this novel metric, we found an average inter-rater agreement of 0.86 and 0.878 for superlabels and sublabels, respectively, among humans.
In selecting the third evaluator, our preliminary experiments revealed that LLMs tended to favour \textsc{justifications}, potentially inflating agreement scores on this first metric.  
To address this, we designed a second metric that weights the evaluations based on a comparison with both human and LLM judgments, providing a more accurate measure of performance. 
Using both custom metrics, we obtained scores of 0.841 (superlabel) and 0.86 (sublabel) for metric one, and 0.476 for the second. 
The second metric led to the selection of \texttt{GPT-4o} as the third evaluator.

\textbf{Task Performance.} As mentioned in Section~\ref{subsubsec:annotation}, we decided to keep explanations, even if they are associated with an incorrect answer.
Just as explanations are inherently tied to their goal, we hypothesised that they might vary depending on the task, and how successful the annotators were.
To explore this, we started by looking at the average performance of each annotator across tasks.
Humans showed an average accuracy of T1: 70.46\%, T2: 69.09\%, T3: 80.78\%, T4: 55.06\%; LLMs showed T1: 78.94\%, T2: 69.24\%, T3: 87.42\%, T4: 47.58\% (as reported in Figure \ref{fig:acc_grouped}). Overall, closed-source LLMs outperformed humans and open-source models.
Interestingly, not only did T2 and T4 have the lowest accuracies, annotators also reported lower confidence on these tasks in comparison to T1 and T3 (see Appendix \ref{subsubsec:survey}). In fact, T4 proved to be the most challenging task for all annotators, while T3 was the least challenging. 
See Figure \ref{fig:acc_individual} in Appendix \ref{app:results_acc} for a breakdown of these accuracies per annotator.

\textbf{Frequency of Explanation Types.} The bar plots in Figure \ref{fig:quality_freq} show the frequencies of each explanation type as judged by and averaged across the three evaluators (\texttt{GPT-4o} and two humans) in the \textit{evaluation set.}
Overall, the evaluators judged explanations to be mostly \textsc{justifications}.
A notable observation is the low frequency of negative types (i.e., \textsc{None}). A closer look at the data revealed that these assignments were predominantly made by human evaluators.
Furthermore, we found that T4 had a much higher proportion of \textsc{arguments} than other tasks, whereas T3, the easiest task, had comparatively fewer.
These results reveal insights into the tendencies of humans and LLMs to generate \textsc{justifications}, whilst also highlighting the influence of task characteristics on the nature of generated explanations. T4 is a notoriously complex task that requires evaluators to go beyond simply recognising correct language use. 
They must also assess the effectiveness of the writing in achieving its intended purpose, which involves subjective judgments about argumentation, organisation, and style.
While some interpretation might be involved in understanding the context in T1, T2 and T3 the range of acceptable interpretations is much narrower.
Thus, our results suggest that the presence of \textsc{arguments} is correlated with the subjectivity of the task.
The relationship between \textsc{arguments} and task subjectivity is reinforced by the findings of our follow-up survey, where human annotators expressed lower confidence in T4.
Upon further inspection of the frequency of \textsc{arguments} across tasks, we found that \texttt{Sonnet 3.5}, while similar in terms of accuracy to \texttt{GPT-4o}, is more likely to produce this type of explanation.
Figure \ref{fig:quality_freq_detailed} in Appendix \ref{app:detailed_analysis} provides a more granular view of these findings.

\textbf{Accuracy Across Types.} The scatter plot in Figure \ref{fig:quality_freq} relates the types of explanations produced by the annotators and their accuracy in each task (\textbf{Task Performance}). We observe an interesting trend in T1, T3 and T4: the ``Overall'' line shows that lower accuracy in a task is associated with the lowest type in Rubrik's hierarchy.
In other words, annotators tended to generate a \textsc{Commentary} when their answers were incorrect whereas a \textsc{Justification} was primarily associated with correct answers and corresponded to the highest accuracy.
T2, however, shows the opposite trend.
Specifically, LLMs tend to generate an \textsc{Argument} (highest type in our hierarchy) whenever they answered incorrectly while humans generated a \textsc{Commentary}.
We hypothesise that the uneven behaviour on this task is due to the multi-label nature of T2. 
A similar variation was observed when we looked at the frequencies of the answer choices picked by the annotators (see Appendix \ref{app:answer_freqs}).

\textbf{Explanation Quality Breakdown.} Regarding the quality of the explanations, the number of \textit{bad} explanations was low and concentrated in \textsc{commentaries} across tasks.
The analysis of sublabel frequencies (plotted in Figure \ref{fig:sublabels}) showed that the main source of a bad explanation was the lack of \textsc{conciseness}, with open-source LLMs averaging 96.89\% and closed-source LLMs averaging 99.06\% on this sublabel. 
An example is shown in Table \ref{table:good_bad_examples}; the \textsc{commentary} is redundant, due to the repetition of details given in the question's context. 
This contrasts with the low frequency of \textsc{word choice}, \textsc{cohesion}, \textsc{appropriateness} and \textsc{grammaticality}.
On the other hand, \textsc{conciseness} is less of a problem to humans, whose explanations are mostly judged as bad due to poor \textsc{coherence}.
Human explanations were different between contractors and experts.
Bad explanations produced by experts were due to \textsc{grammaticality}, while contractors struggled with \textsc{coherence}. Figure \ref{fig:sublabels_detailed} in Appendix \ref{app:detailed_analysis} provides a more granular view of these findings. 

\section{Conclusion}
This work introduces Rubrik, a novel evaluation rubric for assessing the quality of explanations, and a dataset. CUBE, which includes diverse explanations across four tasks, served as the testbed for evaluating Rubrik's effectiveness. 
Rubrik's design, rooted in educational principles, applies insights from education, XAI, and NLG literature. 
Our work contributes to the responsible integration of GenAI into critical decision-making processes, providing a foundation for future advancements in explanation quality assessment.

\clearpage
\section*{Limitations}
\textbf{Scoring strategy.} 
Given the scope of this work, we opted for a binary evaluation strategy, categorising explanations as either \textit{good} or \textit{bad}.
The task of establishing criteria for a \textit{good} explanation presented a significant challenge, necessitating the identification and definition of relevant attributes.
A more nuanced scoring system that reflects varying degrees of quality would be desirable.
However, while a Likert scale might be a convenient choice, developing a valid and reliable graded scale specifically for explanations requires considerably more research.
Our primary goal in this initial study was to assess the viability of our proposed rubric in its simplest form, laying the groundwork for more nuanced evaluations in future work.
Furthermore, our approach does not explicitly assess the quality of reasoning itself. 
While a \textit{good} explanation is generally an indicator of a good reasoning, a poor explanation could stem from how the reasoning is communicated rather than from the reasoning process itself.
Although this is a complex problem, the development of methods for directly assessing reasoning quality is an interesting direction for future research. 

\noindent\textbf{Monolingual Data.} The different attributes (\textsc{dimensions}) of a \textit{good} explanation were taken from studies that exclusively considered English data.
In turn, our work only includes datasets in English as well.
In principle, the \textsc{dimensions} and definitions presented here should extend to other languages.
However, it is possible that some will change depending on the cultural heritage, literature, and history. 
Indeed, the very concept of explanations may differ depending on the linguistic community, which may influence how explanation types, \textsc{components} or \textsc{dimensions} are prioritised or understood.

\noindent\textbf{Annotators' Confidence Assessment.}
After completing the annotation tasks, human annotators were surveyed about their experience, including a self-assessment of their performance.
These responses provided valuable context for interpreting the data analysis results.
As for LLM annotators, they were prompted to assign probabilities reflecting their confidence in each answer option's correctness.
While logit analysis would have been ideal, we hypothesised that requesting that information in the prompt would be sufficiently accurate, especially given that logit access was not available across all models (due to some being closed-source).
However, the resulting probabilities often failed to sum to 100\%, indicating a lack of consistent or meaningful probability assignment. 
Consequently, these assigned probabilities were not considered in the data analysis.
Thus, we lack the means to make meaningful comparisons between human and LLM annotator confidence levels.

\section*{Ethical Considerations}
Prior to commencing the study, ethical approval was obtained from a relevant Ethics Committee. 
Informed consent was obtained from all participants, and their anonymity/confidentiality was ensured throughout the research process.

In light of \citet{baur-2020}'s critique of the current ``AI hype'', we acknowledge the potential for misinterpretation of GenAI capabilities, particularly the risk of users over-relying on automatic explanations in tasks where human oversight is crucial.
Our work aims to mitigate this risk by providing an objective evaluation framework for model outputs.  
This framework enables informed decision-making regarding the selection of the most appropriate resource—whether human or automated—for a given task. 
For instance, Rubrik can identify instances where a less complex model is sufficient, or conversely, when human expertise is required.

Finally, we also recognise the potential for misuse of our framework. 
Indeed, Rubrik could be exploited to deliberately generate misleading or poor-quality explanations. 
This could contribute to the spread of misinformation which poses a serious threat to informed decision-making. This risk highlights the importance of ensuring that the tool is used responsibly.

\section*{Acknowledgments}
We thank Øistein Andersen and Andrew Caines for their help recruiting annotators and their constructive suggestions and advice throughout the project. 
We also thank Camélia Guerraoui for her help conducting the preliminary experiments. 
Many thanks to the labmates at the NLIP Lab at the University of Cambridge, especially Marie Bexte and Iman Jundi for taking the time to review an earlier draft of this paper.
We are deeply grateful to the annotators whose meticulous work was crucial for building our dataset. Our thanks also go to Diane Nicholls and her skilled team of annotators at Cambridge University Press \& Assessment. Finally, we greatly appreciate the anonymous reviewers for their insightful feedback, which significantly strengthened this manuscript.

This paper reports on research supported by Cambridge University Press \& Assessment, by the JSPS KAKENHI Grant Numbers 25K03175, by JST Moonshot R\&D Program Grant Number JPMJMS2236, and by The Nakajima Foundation.

The third author’s contributions were primarily completed while employed at LegalOn Technologies. This paper has not undergone internal review or approval process of LegalOn Technologies.

\section*{Contributions of the Authors}

This project was a large collaboration that would not have happened without dedicated effort from every co-author.

The \textit{idea of the project} originated in discussions among Pride Kavumba, Diana Galvan-Sosa and Keisuke Sakaguchi.
However, Gabrielle Gaudeau's entry as co-first author was essential in leading the project and \textit{designing Rubrik} with Diana Galvan-Sosa.
Paula Buttery, as an advisor, provided valuable input on its design.

The \textit{data selection} was primarily carried out by Diana Galvan-Sosa (T2), Gabrielle Gaudeau (T4), Pride Kavumba (T1) and Yunmeng Li (T3). 
For the \textit{collection of explanations}, Diana Galvan-Sosa and Gabrielle Gaudeau led the collection of human-generated explanations. 
Pride Kavumba and Hongyi Gu led the collection of LLM-generated explanations.

Pride Kavumba and Hongyi Gu led the \textit{experimental implementation}, with Diana Galvan-Sosa, Gabrielle Gaudeau and Yunmeng Li actively participating in the experimental design.
Zheng Yuan provided crucial expert advise and suggestions that shaped the final design.

\textit{Analysis} of the experimental results were first done by Yunmeng Li and Gabrielle Gaudeau, and later updated by Diana Galvan-Sosa.

All co-authors contributed to \textit{writing the paper}, especially Diana Galvan-Sosa, Gabrielle Gaudeau, Pride Kavumba, Yunmeng Li and Hongyi Gu.

\bibliography{bibliography}

\clearpage
\appendix

\section{Rubric Creation}
\label{app:rubric_creation}

\subsection{Components}
\label{app:components_creation}

As the foundational type in Rubrik, a \textsc{commentary} embodies the most basic type of explanation, with its primary objective being to provide an understanding of a decision-making process.
Throughout this work, we assume a situation where there is an explicit set of choices, and one choice is selected over the others.
Then, a decision is the behavioural \textsc{action} of choosing among alternative options \cite{brust2021judgment} and it is complemented by the \textsc{reason} that guided that choice.

If there is \textsc{evidence} to support the decision, a \textsc{commentary} then transitions to a \textsc{justification}.
Note that in either case, the underlying principle of objectivity remains consistent across both types.
A subjective approach to presenting a decision process shifts the main goal of \textit{understanding} the underlying rationale to \textit{persuading} the audience.
This idea aligns with the definition of an \textsc{argument}, which is the result of an activity aimed at convincing a reasonable critic of the acceptability of a standpoint~\citep{lunsford2008sage}.

When considering the nature of argumentation, it is common to refer to the seminal work of~\citet{toulmin-1958-arguments}, who provided a framework for constructing, analysing, and evaluating arguments.
However, we adopt a different perspective, drawing upon the principles of rhetoric.
Although there are some similarities between \textsc{warrant}\textendash \textsc{reason} 
and \textsc{backing}\textendash \textsc{evidence}, this does not hold for the relationship between \textsc{claim}\textendash \textsc{action}. 
In Toulmin's framework, a warrant supports the claim and the backing further supports the warrant, but a claim is always assumed to be linked to a \textit{standpoint}.
Rhetorical argumentation, on the other hand, commonly refers to Aristotle's trio \textit{ethos-logos-pathos}~\citep{braet1992ethos}, where \textit{ethos} refers to the credibility of the speaker, \textit{pathos} refers to the emotional state of the audience and \textit{logos} refers to what is true.
We can identify a relationship between \textsc{logos}\textendash \textsc{commentary} through the \textsc{reason} component and \textsc{ethos}\textendash \textsc{justification} through \textsc{evidence}.
It is then left to \textsc{pathos} to introduce the elements of \textit{persuasion}.
Considering that a stance is usually implicit in discourse, we focus on linguistic markers: metadiscourse features used by writers to express stance~\citep{barbara2024corpus}.
Thus, we merge into one component the essence of \textit{pathos}, usually expressed in discourse through \textsc{affective appeal(s)}, and features from Hyland’s Interpersonal Model of Metadiscourse~\citep{amiryousefi2011metadiscourse}: hedges, boosters, attitude and engagement markers (i.e., \textsc{qualifiers}).

\subsection{Dimensions}
\label{app:dimensions_creation}
We conducted an extensive review of NLP literature including work in Natural Language Generation (NLG) such as Machine Translation (MT) and Educational NLP (including Grammatical Error Correction and Automated Essay Scoring), but also in Linguistics and Cognitive Science. In doing so, we recorded the names of qualities (or \textsc{dimensions}) that people have looked for in explanations or argumentative writing more generally, and, when present, their definitions. We also kept note of how these qualities have been evaluated in a target text, using either human annotators or automated methods. See Table~\ref{table:rubric_dim} for the exhaustive list.

\begin{table}[h]
  \centering
  \small
  \begin{tabular}{p{4cm}}
    \toprule
    \textsc{Dimension Name} \\\midrule
     \textsc{Appropriateness}  \\ 
     Adequacy \\ 
     Clarity \\
     \textsc{Coherence} \\
     \textsc{Cohesion} \\
     Completeness\\
     \textsc{Conciseness}  \\
     Consistency  \\
     Comprehensibility  \\
     Comprehensiveness \\
     Correctness\\
     Factuality \\
     Faithfulness \\
     Fidelity  \\
     Fluency  \\
     \textsc{Grammaticality} \\
     Interpretability  \\
     Organisation  \\
     Persuasiveness \\
     \textsc{Plausibility} \\
     Readability \\
     Reasonableness  \\
     Transparency  \\
     Truth of likelihood\\
     Usefulness \\
     \textsc{Word choice} \\
    \bottomrule
  \end{tabular}
  \caption{Exhaustive list of the quality \textsc{dimensions} of explanation we found when surveying the literature. We highlight in \textsc{capital letters} the names of the \textsc{dimensions} we included in our rubric \textit{verbatim}.}
  \label{table:rubric_dim}
\end{table}

Below we describe how we defined and chose the eight \textsc{dimensions} that are represented in Rubrik. We also introduce a few of the many qualities that were considered and explain why they were excluded, as a demonstration of our overall process. Though we cannot be exhaustive at this time, we rigorously researched each and every one of the dimensions mentioned in Table~\ref{table:rubric_dim}. The final definitions we used in the automated evaluation prompts are provided in Appendix \ref{app:rubric_scoring}. 
The full rubric with examples is shown in Section~\ref{app:full_rubric}.

\subsection{Grammaticality}

\textsc{Grammaticality}, though essential, was surprisingly hard to define. This was largely due to the fact that grammar has a long-standing tradition in a variety of fields---including Linguistics, Psychology, Education, and Cognitive Science---which have each contributed different perspectives and theories over time. As a result there is no single, universally accepted definition. Definitions which originate from the field of Linguistics tend to be highly theoretical, and as a result, quite impractical. A classic example is~\citet[Chapter 1, p.2]{chomsky_aspects_1965} for whom the ``grammar of a language purports to be a description of the ideal speaker-hearer's intrinsic competence'', which has been criticised for being too abstract and disconnected from actual language use~\citep[Chapter 18]{pride_sociolinguistics_1972}. On the other hand, most NLP studies assume that the definition of \textsc{grammaticality} is common knowledge and avoid going through the trouble of formally defining it in the context of their work (e.g., \citeauthor{wei_evaluating_2018}, \citeyear{wei_evaluating_2018}). In fact, it is openly admitted that ``Grammatical Error Correction'' is something of a misnomer as it is now commonly understood to encompass errors that are not always strictly grammatical in nature''~\citep{bryant_grammatical_2023}. 

However, to avoid relying on our intuition of what a grammatical explanation is, we needed to bridge the gap between theory and practice, and find a definition that could be both pragmatic and grounded in the literature. We did find one in a paper by~\citet[Table 10]{hu_are_2024}, similarly focused on the evaluation of LLM outputs, which defines \textsc{grammaticality} as measuring ``whether the target text is grammatically correct without any lexical or syntax errors, regardless of its content and meaning. Consider whether the target text itself complies with the English standard usage and rules of grammar, such as tense errors, misspellings, incorrect prepositions, collocation misusages, and so on.'' In using this definition, it is quite straightforward to classify \textsc{Grammaticality} as a \textbf{Language} \textsc{dimension} as it in no way attends to the content of the text.

\subsection{Conciseness}

In contrast, we found \textsc{Conciseness} to be well-documented across many literatures and much less controversial. In Education, ``concise writing gets to the point quickly and does not introduce unnecessary information''
~\citep[p.25]{long_college_2007} and requires you to ``cut fat'' into your writing by ``eliminating redundancies, eliminating writing zeroes, reducing sentences to simplest form, and cutting bureaucratic waste''~\citep[Chapter 8]{alley_language_1996}. Similarly, in NLP,~\citet{cao_automatic_2022} define it as a measure of ``non-redundancy'' in text, sometimes through the number of repeated words~\citep{peyrard_simple_2019} or through computing sentence similarities~\citep{wan_manifold-ranking_2007}. 

We finally opted for~\citet{kabir_is_2024}'s comprehensive taxonomy of three conciseness issues:
\begin{quote}
    \textit{Redundant} sentences reiterate information stated in the question or in other parts of the answer. \textit{Irrelevant} sentences talk about concepts that are out of the scope of the question being asked. And lastly, \textit{Excess} sentences provide information that is not required to understand the answer.
\end{quote} Not only were these issues identified when evaluating ChatGPT answers, a task closely related to ours, we additionally felt that they encompassed all the elements that were individually picked out in previous definitions. Note that since this definition is concerned with redundant, irrelevant or excess information, not just language, we decided to classify \textsc{Conciseness} as a \textbf{Content} dimension.

\subsection{Fluency}

For a while, we considered fluency, an important notion in Machine Translation, which is generally evaluated by humans (e.g., \citeauthor{callison-burch_meta-_2007}, \citeyear{callison-burch_meta-_2007}; \citeauthor{ graham_continuous_2013}, \citeyear{graham_continuous_2013}; \citeauthor{bojar_findings_2016}, \citeyear{bojar_findings_2016}), or using automated metrics (e.g., \citeauthor{toral_multifaceted_2017}, \citeyear{toral_multifaceted_2017}; \citeauthor{martindale_identifying_2019}, \citeyear{martindale_identifying_2019}; \citeauthor{feng_modeling_2020}, \citeyear{feng_modeling_2020}). In the first case, we found that human annotators were almost never provided with a proper definition of fluency and expected to use their intuition of what the word meant via prompts like ``\textit{how do you judge the fluency of this translation?}'' in~\citet{callison-burch_meta-_2007} or ``\textit{read the text below and rate it by how much you agree that: the text is fluent English}'' in~\citep{graham_continuous_2013}. In the latter case, the metrics used were only considered to be proxies for fluency which was never actually defined.

As with \textsc{Grammaticality},~\citet[Table 9]{hu_are_2024} provided the following definition: ``[fluency] measures the quality of individual sentences, are they grammatically correct, non-repetitive, and in accord with common English usage, with clear meanings'', which seemed to overlap both our definitions for \textsc{Conciseness} and \textsc{Grammaticality}. Since our goal was to reach a set of well-delineated, atomic dimensions, we chose to discard it.

\subsection{Cohesion}

\textsc{Cohesion} is a very important notion in Linguistics and is classically defined by~\citet[p.4]{halliday_cohesion_2014} as:
\begin{quote}
    occur[ring] where the \textsc{interpretation} of some element in the discourse is dependent on that of another. The one \textsc{presupposes} the other, in the sense that it cannot be effectively decoded except by recourse to it. When this happens, a relation of cohesion is set up, and the two elements, the presupposing and the presupposed, are thereby at least potentially integrated into the text.
\end{quote}
Unfortunately, as with \textsc{Grammaticality}, this definition is not accessible to most people and is far too theoretical.

However, \textsc{Cohesion} is also widely present in Education, particularly in writing assessment and teaching literature, due to the common idea that a written text's quality is highly related to its level of \textsc{Cohesion}~\citep{mcnamara_cohesion_2010}. This belief is reflected in the literature about writing (e.g., \citeauthor{collins_strategies_1998}, \citeyear{collins_strategies_1998}, \citeauthor{devillez_writing_2003}, \citeyear{devillez_writing_2003}) and the rubrics that teachers use to assess writing (e.g., \citeauthor{arnold_ielts_2023}, \citeyear{arnold_ielts_2023}; \citeauthor{crossley_english_2024},\citeyear{crossley_english_2024}). It is notably defined by~\citet{mcnamara_cohesion_2010} as follows:
\begin{quote}
    Cohesion refers to the presence or absence of explicit cues in the text that allow the reader to make connections between the ideas in the text. For example, overlapping words and concepts between sentences indicate that the same ideas are being referred to across sentences. Likewise, connectives such as `because', `therefore', and `consequently', inform the reader that there are relationships between ideas and the nature of those relationships.
\end{quote} Or more simply as the ``appropriate use of transition phrases'' by~\citet[Table 1]{ke_automated_2019}. For our purposes, we prefer these pragmatic definitions to those offered by Linguistics. 

From these definitions, it seems that \textsc{Cohesion} is only concerned with \textbf{Language} not the content of a text. In fact, the dimension has also been examined through automated tools like Coh-Metrix~\citep{mcnamara_automated_2014} or TAACO~\citep{kyle_automatically_2015}, which use a compound of linguistic metrics like the Type Token Ratio (TTR; \citeauthor{mccarthy_vocd_2007}, \citeyear{mccarthy_vocd_2007}) as proxies for \textsc{Cohesion}. 

\subsection{Coherence}

A related notion to \textsc{Cohesion} is \textsc{Coherence}. It has been defined in Linguistics as a ``continuity of sense'' by~\citet[p.84]{beaugrande_introduction_1981}, or more concretely as ``the state of being logically consistent and connected''~\citep{jaszczolt_textual_2012}.  It is also an important notion in Document Summarisation, where \textsc{Coherence} is similarly defined as ``what makes multiple sentences semantically, logically and syntactically coherent''~\citep{yao_recent_2017}. It is also frequently evaluated writing assessment either by humans (e.g., \citeauthor{higgins_evaluating_2004}, \citeyear{higgins_evaluating_2004}) or via automated methods (e.g., \citeauthor{higgins_evaluating_2004}, \citeyear{higgins_evaluating_2004}; \citeauthor{miltsakaki_evaluation_2004}, \citeyear{miltsakaki_evaluation_2004}; \citeauthor{wu_learning_2018}, \citeyear{wu_learning_2018}).

Where \textsc{Cohesion} is an ``overt (or explicit) linguistic-surface phenomenon, [...] coherence is a covert (or implicit) deep-structure phenomenon''. But while \textsc{Coherence} is more concerned with meaning (i.e., \textbf{Content}) than form ~\citep{jaszczolt_textual_2012}, it also ``depends on a number of factors, including explicit cohesion cues, implicit cohesion cues (which are more closely linked to text coherence than are explicit cues), and nonlinguistic factors such as prior knowledge and reading skill''~\citep{kyle_automatically_2015}. They are thus ``interdependent'' notions~\citep{zhang_text-based_2006}. To portray this in our rubric, we chose to similarly relate both \textsc{Dimensions}: an explanation should thus not be labelled as coherent without first being judged as cohesive. 

\subsection{Clarity}
\label{app:clarity}

We first encountered this quality while looking at writing education papers, where clarity generally ``refers to how clearly an author explains the thesis of her essay, i.e., the position she argues for with respect to the topic on which the essay is written''~\citep{persing_modeling_2013}. It also appears in the ICLE++ corpus of persuasive student essays~\citep{granger_international_2009, li_icle_2024}, an important dataset in the field of Automated Written Assessment. However, the definitions we found were far too vague and we struggled to find more formal or practical descriptions of the term which seemed to support~\citet[Chapter 2]{beaugrande_introduction_1981}'s claim that clarity is ``too vague and subjective to be reliably defined and quantified''. We ultimately decided to drop this \textsc{dimension}. 

\subsection{Word Choice}

The \textsc{Word Choice} \textsc{dimension} is broadly defined as ``the choice and aptness of the vocabulary used''~\citep{mathias_asap_2018}. It is frequently included in written assessment rubrics (e.g, see the very detailed 6-point rubric for this dimension in the ASAP\footnote{\label{foot:asap} The original dataset and annotation guidelines can be downloaded from \href{https://www.kaggle.com/c/asap-aes/data}{https://www.kaggle.com/c/asap-aes/data}.} corpus) and the focus of automated assessment research (e.g., \citeauthor{kyle_automatically_2015}, \citeyear{kyle_automatically_2015}; \citeauthor{kyle_tool_2018}, \citeyear{kyle_tool_2018}; \citeauthor{kristoffersen_where_2019}, \citeyear{kristoffersen_where_2019}).

We also came across~\citet{stede_lexical_2002}'s work on lexical choice for NLG:

\begin{quote}
    Generally speaking, the point of ``interesting'' language generation (that is, more than merely mapping semantic elements one-to-one onto words) is \textbf{to tailor the output to the situation at hand}, where ``situation'' is to be taken in the widest sense, including the regional setting, the topic of the discourse, the social relationships between discourse participants, etc.
\end{quote}

Though not explicitly defining \textsc{Word Choice}, the above citation introduces the idea that every ``interesting'' or \textit{good} utterance (or in our case, explanation) is made within a given ``situation'' and thus evaluating the language of that utterance should be context-dependent. It is this \textbf{context} that dictates what is ``apt''~\citep{mathias_asap_2018}. Realising that it is necessary to define an evaluation \textit{context} before starting any kind of evaluation (see Section~\ref{subsec:scoring_strategy}) was a turning point for our rubric. 

Now, \textit{context}-appropriateness relies on both form and content. However, due to the strong emphasis on evaluating \textsc{Word Choice} as a surface-level feature, not a content one, in automated assessment research, we chose to classify it as a \textbf{Language} \textsc{dimension}.

\subsection{Appropriateness}

\textsc{Appropriateness} defined in Linguistics by~\citet{canale_communicative_1983} as ``the extent to which particular communicative functions [...] and \textbf{ideas} are judged to be proper in a given situation'' or as ``an optimal mapping between context and speech, or as `natural speech,' is also connected intrinsically with the sociocultural notions of politeness and impoliteness'' by~\citet{fetzer_appropriateness_2018}. This term also occasionally appears in AI literature as something we must ensure in the systems we develop, and thus, evaluate (e.g., \citeauthor{spitale_appropriateness_2024}, \citeyear{spitale_appropriateness_2024}; \citeauthor{javidan_evaluating_2024}, \citeyear{javidan_evaluating_2024}; \citeauthor{balta_evaluating_2025}, \citeyear{balta_evaluating_2025};). There, it is more often related to other qualities such as safety, consistency, and readability. Hence, \textsc{Appropriateness} is a complex, multi-faceted dimension which also relies on \textit{context}. 

For our purpose, we needed to relate this \textsc{dimension} to \textsc{Word Choice}. For this, we turned to the prominent sociolinguist, Dell Hymes who ``pointed out that appropriateness [depend] both on linguistic and sociocultural competence'' \citep{dewaele_appropriateness_2008}, and defined it as ``what to say to whom in what circumstances and how to say it'' in \citet[p.277]{hymes_communicative_1972}. We deem that this last part, ``how to say it'' is already encompassed by our definition of \textsc{Word Choice}. Further, ``to whom in what circumstances'' refers to our very own definition of the \textit{context}, which leaves us with the ``what to say'' for \textsc{Appropriateness}, that is, the \textbf{Content}. 

\subsection{Plausibility}

In reading around the topic of explanations in AI, we came across the following trait: ``the \textbf{truth of likelihood} of an explanation is considered an important criterion of a good explanation'' in a paper by \citet{miller_explanation_2019}. The term was used to refer to facts that were judged as ``either true or likely to be true by the explainee.'' We note that in no way is our rubric intended to evaluate the truth condition of explanations. However, we felt that it was important that our rubric allows for \textsc{justification} to be evaluated as \textit{bad} or of \textit{bad} quality if their \textsc{evidence} was deemed implausible by the evaluator. After some research, we could not find any other mention of the ``truth of likelihood'' and sought a more general name for our \textsc{dimension}.

A related notion was \textsc{Plausibility} which was present in similar literature and already being used to evaluate explanations. For instance, \citet{agarwal_faithfulness_2024} who define plausible explanations as being ``seemingly logical and coherent to human users'' or as ``being convincing towards the model prediction, regardless of whether the model was correct or whether the interpretation is faithful'' by \citet{jacovi_aligning_2021}. Though not exactly similar, the latter introduces the idea that using \textsc{Plausibility} as criteria for a \textit{good} explanation might encourage deception. As a result, the authors advise against pursuing this \textsc{dimension}.

Taking this warning into consideration, it was important to us to centre our definition of \textsc{Plausibility} around the \textsc{evidence} component (2.a), and we modified \citet{agarwal_faithfulness_2024}'s Definition 1, substituting the word ``explanation'' with ``evidence'':
\begin{quote}
    An evidence* is considered plausible if it is coherent with human reasoning and understanding.
\end{quote}

\subsection{Stance Clarity}

Whenever we found a mention of \textsc{Arguments} in the literature, the concept of persuasiveness was almost always mentioned. It thus seemed natural that it would be included in our rubric. We first looked at the notion of ``argument strength'' in persuasive writing which is defined, in an admittedly very circular fashion, as ``the strength of the argument an essay makes for its thesis'' and evaluated by \citet{persing_modeling_2015}. In a similar vein, we discovered work by \citet{song_applying_2014} and \citet{stab_annotating_2014} which designed argument schemes for annotating arguments manually in student essays. Yet, none of the definitions we found seemed right.

We then turned to persuasiveness in rhetoric, and found \citet[Table 5]{connor_linguisticrhetorical_1990}'s Persuasive Appeals Scale. Though very useful, we struggled to see whether these were in fact \textsc{components} or indeed a \textsc{dimension}, and where to fit them in our rubric. After some iterations, we arrived at the fact that the presence of \textsc{affective appeals} and \textsc{qualifiers} in an argument help us understand what the explainer's ``stance'' is, that is, their personal ``feeling, attitude, perspective, or position as enacted in discourse'' \citep{strauss_discourse_2013}. By that point, it felt like persuasiveness was too vague and we coined the term ``Stance Clarity'' for our last \textsc{dimension}.

\subsection{Full Rubric}
\label{app:full_rubric}

\begin{table*}
  \centering
  \small
  \begin{tabular}{l | p{3.5cm} | p{4cm} p{4cm}}
    \toprule
    & & & \\
    & \centering{\textsc{\textbf{Components}}} & \multicolumn{2}{c}{\textsc{\textbf{Dimensions}}}  \\
      & & \multicolumn{1}{p{5cm}|}{\centering{\textbf{Language}}} & \multicolumn{1}{c}{\textbf{Content}} \\\hline
    & & \multicolumn{1}{p{5cm}|}{} & \\
    \includegraphics[width=1em]{images/cube_blue_face.png} \textbf{\textsc{Commentary}} &  \textbf{\textsc{Action}} (1.a): does the explanation clearly indicate the decision or choice being made (e.g., specifying the selected answer)? For e.g., 
    \begin{itemize}[leftmargin=0.5cm]
        \item Acceptable: ``\textbf{The correct answer is A}.''
        \item Not acceptable: ``Because it is the final part of the sequence.''
    \end{itemize}
    &  \multicolumn{1}{p{5cm}|}{\textbf{\textsc{Grammaticality}}: is the explanation grammatically correct, free of lexical or syntax errors? \textit{Small typos are acceptable, but the errors should \textbf{\underline{not}} impede comprehension in any way.} For e.g., 
    \begin{itemize}[leftmargin=0.5cm]
        \item Acceptable: ``The correct answer is A because nowadays our \textbf{socity} is based on consumerism and the way in which we are producing is contaminating the \textbf{word}.''
        \item Not acceptable: ``The correct answer is A because \textbf{now a day} our \textbf{socity it is bassed in consumer}, \textbf{so that become} the \textbf{word more contaminate} to produce the products \textbf{that we demanding}.''
    \end{itemize}} & \textbf{\textsc{Conciseness}}: is the explanation free of any redundant, irrelevant, or excess sentences (that is, not required to understand the answer)? For e.g., given that the answer choice D is ``next she explains how to use the lawnmower and other tools and then she cuts the grass,''
    \begin{itemize}[leftmargin=0.5cm]
        \item Acceptable: ``The correct answer is D because it accurately reflects the sequence of events.''
        \item Not acceptable: ``The correct answer is D because \textbf{she explains how to use the lawnmower and other tools, and then she cuts the grass}.''
    \end{itemize} \\
     & \textbf{\textsc{Reason}} (1.b): does the explanation provide reasoning or insight into why the decision or choice was made, explaining the underlying logic or rationale for the Action? For e.g.,
     \begin{itemize}[leftmargin=0.5cm]
        \item Acceptable: ``The right answer is C, \textbf{because it is the final part of the sequence}.''
        \item Not acceptable: ``The correct answer is A.''
     \end{itemize} & \multicolumn{1}{p{5cm}|}{\textbf{\textsc{Word Choice}}: is the language used in the explanation tailored to the given \textit{context} (task, audience, purpose)? And are the sentences in the explanation well-formed? For e.g., 
    \begin{itemize}[leftmargin=0.5cm]
        \item Acceptable: ``The correct answer is A because the essay lacks fluency, has many incorrect clauses and missing words. And while the overall meaning can be deduced, the essay does not demonstrate an accurate grasp of language.''
        \item Not acceptable: ``\textbf{Answer A. l}ack of fluency, incorrect clauses and missing words, \textbf{meaning} can be found but does not demonstrate an accurate grasp of language.''
    \end{itemize}} & \textbf{\textsc{Appropriateness}}: is the explanation culturally appropriate, matching expectations for the given \textit{context}? For e.g., 
    \begin{itemize}[leftmargin=0.5cm]
        \item Acceptable: ``The right answer is B because the tenses are properly used and the story makes sense.''
        \item Not acceptable: ``The right answer is B  because the tenses are properly accorded and \textbf{(within the slightly odd context)} the story makes sense.''
    \end{itemize} \\
     & & \multicolumn{1}{p{5cm}|}{\textbf{\textsc{Cohesion}}: does the explanation make appropriate use of 
transition phrases (e.g., connectives like ``because'', ``therefore'', and ``consequently'', overlapping words across sentences, etc.)?  For e.g., 
\begin{itemize}[leftmargin=0.5cm]
    \item Acceptable: ``The correct answer is C because the man is on roller blades, not on a skateboard. Further, he is not talking to anyone and therefore cannot possibly `continue speaking'.''
    \item Not acceptable: ``The correct answer is C, \textbf{because} the man is on roller blades, not a skateboard, \textbf{and} is not talking to anyone in the example \textbf{so cannot} `continue speaking'.''
    \end{itemize}} & \textbf{\textsc{Coherence}}: does the explanation appropriately transition between ideas, i.e., does it make sense as a whole (e.g., good context-relatedness, semantic consistency, and inter-sentence causal and temporal dependencies, etc.)? For e.g., given the start of explanation ``The correct answer is D, because no information about Liu's relationship to science subjects specifically is given in the passage,''
    \begin{itemize}[leftmargin=0.5cm]
        \item Acceptable: ``therefore the fact that they like chemistry is implied and ambiguous.''
        \item Not acceptable: ``therefore the fact that they like \textbf{cheese} is implied and ambiguous.''
    \end{itemize} 
  \end{tabular}
  \caption{Extended rubric with definitions and illustrative examples for each of the \textsc{Components} and \textsc{Dimensions} (continued on next page).}
  \label{table:rubric_extended}
\end{table*}

\setcounter{table}{5}
\renewcommand{\thetable}{\arabic{table} (contd.)}

A concise overview of Rubrik is presented in Section~\ref{subsec:eval_criteria}, Table~\ref{table:rubric_final}.
This appendix provides the complete details of the full-sized, illustrated rubric in Table~\ref{table:rubric_extended} and Table~\ref{table:rubric_extended_continued}.

\begin{table*}
  \centering
  \small
  \begin{tabular}{l | p{3.7cm} | p{3.8cm} p{4cm}}
  
    \toprule
    & & & \\
    & \centering{\textsc{\textbf{Components}}} & \multicolumn{2}{c}{\textsc{\textbf{Dimensions}}}  \\
      & & \multicolumn{1}{p{4.8cm}|}{\centering{\textbf{Language}}} & \multicolumn{1}{c}{\textbf{Content}} \\\hline
    & & \multicolumn{1}{p{4.8cm}|}{} & \\
    \includegraphics[width=1em]{images/cube_yellow_face.png} \textbf{\textsc{Justification}} &  \textbf{\textsc{Evidence}} (2.a): does the explanation provide concrete evidence (can be both explicit or implicit) that supports the reasoning, such as information from the question's context or general knowledge? For e.g., 
    \begin{itemize}[leftmargin=0.5cm]
        \item Acceptable: ``The right answer is C, because it finishes the sequence, \textbf{describing the effect of bowling the ball and what happens as a result}.''
        \item Not acceptable: ``The right answer is C, because is is the final part of the sequence.''
    \end{itemize}
    &  \multicolumn{1}{p{4.8cm}|}{} & \textbf{\textsc{Plausibility}}: is the provided \textsc{\textbf{Evidence}} plausible and consistent with human reasoning, considering the context and general world knowledge? For e.g., 
    \begin{itemize}[leftmargin=0.5cm]
        \item Acceptable: ``The correct answer is A (`Jack picks the cheese') because \textbf{we are told that he enjoys eating `mozzarella' in the morning}.''
        \item Not acceptable: ``The correct answer is A (`Jack picks the cheese') because \textbf{my name is also Jack and I personally love cheese for breakfast}.''
    \end{itemize} \\\hline
     & & \multicolumn{1}{p{4.8cm}|}{} & \\
     \includegraphics[width=1em]{images/cube_red_face.png} \textbf{\textsc{Argument}} &  \textbf{\textsc{Affective appeal(s)}} (3.a): does the explanation use vivid, or emotionally charged language (e.g., metaphors) to evoke feelings in the audience? For e.g., 
    \begin{itemize}[leftmargin=0.5cm]
        \item Acceptable: ``The expression in the final section is very \textbf{heartfelt}; the tone is \textbf{excitable} and \textbf{keen} throughout.''
        \item Not acceptable: ``The final section reflects the writer's strong feelings on this issue.''
    \end{itemize}
    &  \multicolumn{1}{p{4.8cm}|}{} & \textbf{\textsc{Stance clarity}}: is the explainer's stance (their personal feelings towards the task) clearly and unambiguously conveyed through affective appeals or qualifiers? Note that the stance can be implicit unlike the \textsc{\textbf{Action}}. For e.g., 
    \begin{itemize}[leftmargin=0.5cm]
        \item Acceptable: ``The correct answer is A (beginner) because this text is \textbf{undeniably} of a low English level.''
        \item Not acceptable: ``The correct answer is A (beginner) because this text is \textbf{clearly} of a low English level although the final section is \textbf{incredibly} well written.''
    \end{itemize} \\
    & \textbf{\textsc{Qualifiers(s)}} (3.a): does the explanation make use of hedges, boosters, attitude markers, self-mentions, or engagement markers? For e.g., 
    \begin{itemize}[leftmargin=0.5cm]
        \item Acceptable: ``The right answer is B, because the text is keeping with what is \textbf{presumably} a tour guide's voice: \textbf{intentionally} using clunky and \textbf{overly} expressive words.''
        \item Not acceptable: ``The right answer is B, because the text is keeping with the original tour guide's voice.''
    \end{itemize} & \multicolumn{1}{p{4.8cm}|}{} & 
    \\ \bottomrule
  \end{tabular}
  \caption{Extended rubric with definitions and illustrative examples for each of the \textsc{Components} and \textsc{Dimensions}.}
  \label{table:rubric_extended_continued}
\end{table*}

\renewcommand{\thetable}{\arabic{table}}

\section{Data Selection}
\label{sec:selection}

Considering the fact that the four datasets we chose to work with were all of different sizes, we chose to only work with a subset of each dataset: namely $n = 1000$ instances for each task.
Thus, our \textit{base set} has a total of $4000$ instances.

We collected a set of human-written (see Section~\ref{subsec:human_annotation}) and LLM-generated explanations (see Section~\ref{subsec:llm_annotation}).
Due to limitations in time and resources, only a subset of the $1000$ instances was shown to the annotators: namely $n = 110$ instances for each task.
Thus, our \textit{annotation set} has $440$ instances.
The following subsections detail the subset selection criteria.

\subsection{Commonsense Reasoning}
\label{subset_task1}

\textbf{Base set}. Each \textsc{context} in the \textsc{hellaswag} dataset is taken either from ActivityNet's video captions or WikiHow's how-to-articles.
During the annotator's training (see Section~\ref{subsubsec:annotators-training}), questions whose context made reference to a video were constantly flagged as \textit{``not clear or ambiguous''}.
Thus, we filtered instances that include the word \textit{``camera''}, \textit{``video''} or \textit{``clip''}. 
After that, instances were selected randomly, making sure that the correct answers were distributed as evenly as possible across the four options (A-D), with roughly 25\% assigned to each.

\begin{table}[h]
\centering
\small
\begin{tabular}{lll}
\toprule
{\bf Correct answer} & {\bf Base set}& {\bf Ann set}\\\midrule
A & 267 & 27\\
B & 228 & 28\\
C & 266 & 27\\
D & 239 & 28\\\midrule
Total & 1000 & 110\\
\bottomrule 
\end{tabular}
\caption{\label{hellaswag_ans} Distribution of questions across each possible correct answer for T1's \textit{base set} and \textit{annotation set}.}
\end{table}

\textbf{Annotation set}. Since the \textit{base set} already had an even distribution of the four answer choices, we selected a proportionally representative subset of 110 instances.
See Table~\ref{hellaswag_ans} for a summary of this selection process.

\subsection{Fallacy Detection}
\textbf{Base set}.~\citet{jin_logical_2022} classified fallacies in the \textsc{logic} dataset into 13 fallacy types.
Due to potential overlap between some of the initial types and dataset imbalance, we focused on a subset of 7 types.

Selecting instances within the 30-300 character range effectively eliminated instances requiring specialised political or religious knowledge, ensuring consistent annotation based on general knowledge.
After manual inspection, we removed some duplicated instances and statements that were not exactly fallacies, but rather someone's opinion on a topic.
We also identified a few instances that were incorrectly labelled (i.e., were assigned the wrong fallacy type).
Those were re-labelled and kept in the final subset.
Table~\ref{fallacy_types} shows the final distribution of our subset.

\begin{table}[h]
\centering
\small
\begin{tabular}{llll}
\toprule
{\bf Logical Fallacy} & {\bf Inc}& {\bf Base set} & {\bf Ann set}\\\midrule
Faulty Generalisation & \ding{51} & 289 & 17\\
Ad Hominem & \ding{55} \\
Ad Populum & \ding{55} \\
False Causality & \ding{51} & 154 & 15\\
Circular Claim & \ding{51} & 112 & 15\\
Appeal to Emotion & \ding{51} & 109 & 15
\\
Fallacy of Relevance & \ding{55} \\
Deductive Fallacy & \ding{51} & 120 & 15\\
Intentional Fallacy & \ding{55} \\
Fallacy of Extension & \ding{55} \\
False Dilemma & \ding{51} & 118 & 17\\
Fallacy of Credibility & \ding{51} & 95 & 16\\
Equivocation & \ding{55} \\ \midrule
Total & & 1000 & 110\\
\bottomrule 
\end{tabular}
\caption{\label{fallacy_types} Distribution of instances across each fallacy type for T2's \textit{base set} and \textit{annotation set}.}
\end{table}

\textbf{Annotation set}. This task was originally framed as a classification task.
For the purposes of this research, we adapted the task to follow an MCQ format, where the \textsc{context} was the fallacy statement, and each of the fallacy types was listed as \textsc{answer choices}.
We aimed for a balanced distribution of correct answers across the seven options (A-G).
Instances were selected randomly from the \textit{base set}.
See Table~\ref{fallacy_types} for a summary of this selection process.

\subsection{Reading Comprehension}
\textbf{Base set}. \textsc{race} data is grouped by difficulty (\textsc{race-m}: middle school; \textsc{race-h}: high school).
To better understand the dataset, authors subdivided questions into five reasoning categories.
Since the \textit{Passage Summarization} and \textit{World Knowledge} do not fully require students to carefully read the passage to answer, we focused on the other three question types: \textit{Detail Reasoning}, \textit{Whole Picture Reasoning}, and \textit{Attitude Analysis}.
Specifically, answers to \textit{Detail Reasoning} questions cannot simply be found by matching the questions to the reading passages and require test-takers to provide reasons for their choices.
For \textit{Whole Picture Reasoning} questions test the students' overall understanding of a story. 
\textit{Attitude Analysis} questions ask about the opinions or attitudes of the author or characters of the reading passages.

Unfortunately, the questions have not been labelled with these reasoning categories in the published dataset; hence, we manually selected the data based on the description and examples given by~\citet{lai-etal-2017-race} and reviewed them to ensure quality.

\begin{table}[h]
\centering
\small
\begin{tabular}{llll}
\toprule
{\bf Question type} & {\bf Inc}& {\bf Base set}& {\bf Ann set}\\\midrule
Detail reasoning & \ding{51} & 400 & 36\\
Whole-picture reasoning & \ding{51} & 400 & 37\\
Passage summarization & \ding{55} \\
Attitude analysis & \ding{51} & 200 & 37\\
World knowledge & \ding{55} \\ \midrule
Total & & 1000 & 110\\
\bottomrule 
\end{tabular}
\caption{\label{question_types} Distribution of text passages across each question type for T3's \textit{base set} and \textit{annotation set}. }
\end{table}

\textbf{Annotation set}. Each question in \textsc{race} has four answer choices (A-D).
We aimed for a balanced distribution of instances of correct answers across options within each question type.
Instances were randomly selected from the \textit{base set}, targeting a proportion of approximately 25\% per option.
See Table~\ref{question_types} for a summary of this selection process.

\subsection{Essay Scoring}
\label{subsec:selection_essayscoring}

\begin{figure}
    \centering
    \subcaptionbox{) W\&I corpus word count distribution. We highlight in orange the region from which the \textit{base set} essays were selected.\label{fig:sub-wandi-full}}{\centering\includegraphics[height=6cm]{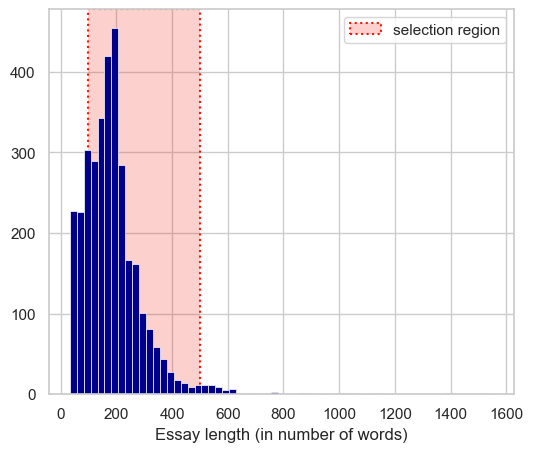}}
    \subcaptionbox{) \textit{Base set} word count distribution.\label{fig:sub-wandi-base}}{\centering\includegraphics[height=6cm]{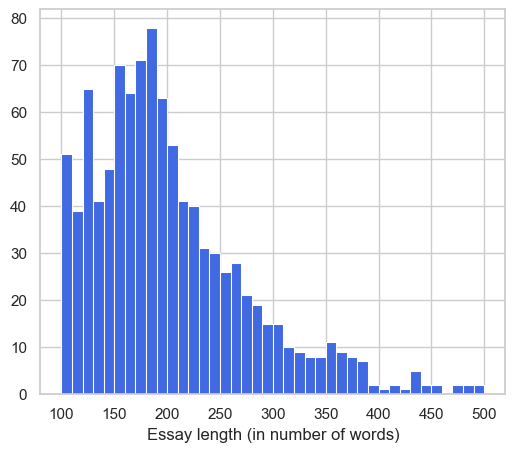}  }
    \subcaptionbox{) \textit{Annotation set} word count distribution.\label{fig:sub-wandi-ann}}{\centering\includegraphics[height=6cm]{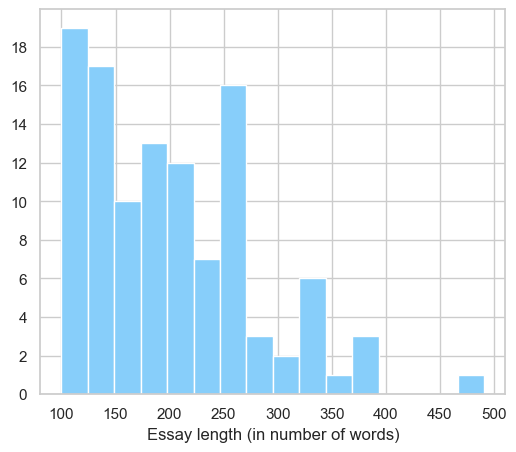}}
    \caption{Plotting the word count distributions }
    \label{fig:wandi-dist}
\end{figure}

\textbf{Base set}. In the W\&I corpus, essays range between 33 and 1,551 words in length. Figure~\ref{fig:sub-wandi-full} plots this distribution. We chose to exclude essays of less than 100 words, and more than 500 words, to avoid selecting essays sitting on either extreme of this distribution. Indeed, essays that are too short might contain too little information to be interesting to evaluate; essays that are long might exceed the limits of LLM contexts or prove too time-taking to annotate for humans. This step left us with a remaining total of 2,598 essays (833 A-scored essays, 1,039 B-scored essays, and 726 C-scored essays). 
Then, we randomly sampled 333 essays from each CEFR level group (334 for the B level) to obtain our \textit{base set} of 1000 essays. We additionally randomly selected 3 essays (one of each CEFR level) from the remaining pool of essays to be used as examples in our experiments.

\textbf{Annotation set}. For our \textit{annotation set}, we again selected randomly from the \textit{base set}, aiming for a balanced distribution of essays across the three CEFR levels. See Table~\ref{wandi_ans} for a summary of this selection process.

\begin{table}[h]
\centering
\small
\begin{tabular}{llll}
\toprule
{\bf Essay Grade} & {\bf W\&I} & {\bf Base set} & {\bf Ann set}\\\midrule
A & 1430 & 333 & 36 \\
B & 1100 & 334 & 37 \\
C & 770 & 333 & 37 \\\midrule
Total & 3300 & 1000 & 110\\
\bottomrule 
\end{tabular}
\caption{\label{wandi_ans} Distribution of W\&I essays across each CEFR level for T4's \textit{base set} and \textit{annotation set}.}
\end{table}

\begin{table}[h]
\centering
\small
\begin{tabular}{lllllll}
\toprule
{\bf Essay Grade} & \multicolumn{2}{l}{\bf W\&I} & \multicolumn{2}{l}{\bf Base set} & \multicolumn{2}{l}{\bf Ann set} \\
& $\mu$ & $\sigma$ & $\mu$ & $\sigma$ & $\mu$ & $\sigma$ \\\midrule
A & 125 & 70 & 163 & 56 & 150 & 51 \\
B & 211 & 100 & 207 & 73 & 205 & 71 \\
C & 262 & 132 & 235 & 71 & 245 & 77 \\\midrule
Overall & 186 &
113 & 201 & 73 & 201 & 78 \\
\bottomrule 
\end{tabular}
\caption{\label{wandi_wordcount} Mean ($\mu$) and standard deviation ($\sigma$) word count of the essays in the W\&I corpus, the \textit{base set}, and the \textit{annotation set} (rounded to the nearest integer).}
\end{table}

\section{Data Collection}
\label{sec:annotation}
\subsection{Human Annotators}
\label{subsec:human_annotation}

We recruited seven human annotators: four research assistants (contractors) and three professional annotators (experts). 
One of the main authors, along with a senior researcher, led the contractors' recruiting efforts, which included conducting interviews with potential candidates.
We selected individuals who appeared to have strong abilities in \textbf{attention to detail}, \textbf{assessment}, and \textbf{strong language skills.}
These skills were essential for completing the assigned reasoning and language tasks.
The PA's were annotators who were specially trained EFL (English as a Foreign Language) teachers and examiners.
The annotators were paid an hourly rate of £22.59 for their work.
We anonymised the annotations by removing any personally identifiable information.
Each annotator was identified with a randomly assigned ID (e.g., 000005FB, 000004E4)

\subsubsection{Training}
\label{subsubsec:annotators-training}
All annotators received a detailed annotation guide that introduced the four tasks and provided a number of annotated examples (question + answer choices + correct answer) for each task.
The examples were intended to help them familiarise themselves with the tasks.
Since T2 necessitates some familiarity with fallacious reasoning, this task was further supported by an appendix with definitions of all fallacy types.\footnote{Specifically, the information provided by~\citet{jin_logical_2022} in their Appendix D.}
We did \underline{not} include explanations to avoid biasing the annotators as to what a \textit{good} explanation should look like.
The annotation guide also included a series of guidelines they should abide by during the annotation process.

Upon reading the annotation guide, the annotators were asked to write explanations for each of the annotated examples contained in the guide.
Their explanations were then reviewed by two of the main authors to ensure they were acceptable in terms of format and length.\footnote{Since the guide does not specify a minimum length for the explanations, we made sure annotators wrote complete sentences as opposed to disjointed notes.}
Unless absolutely necessary, annotators did \underline{not} receive any \underline{feedback} on their explanations.

Subsequently, each annotator received an invitation-only Google Spreadsheet with a set of 15 to 40 examples per task.\footnote{The number varied according to the difficulty of each task. For example, the questions in T2 were short but required more specific knowledge while T3 questions contained longer but easier-to-read texts.}
Before beginning their annotation work, the annotators were reminded that:
\begin{enumerate}
    \item They were asked to dedicate exactly 20 minutes per task (for a total of 1h20min) and should \underline{not} necessarily aim to complete all the questions provided in the allocated time. 
    \item At the end of each 20 minute set, the annotators were told to move onto to the next task without delay and asked not to go back to any previous task (even if they had time to spare).
    \item They were asked to select only \underline{one single answer} per question from the set of potential answers, and to \underline{not} explain their decision process during the training phase.
    \item Within one task, they were allowed to attempt the questions in any given order. However, they were asked \underline{not} to spend more than 5 minutes on a single question. In order to manage their time more efficiently, it was also recommended that they (1) flag difficult questions as they found them, moving immediately to the next one. In other words, they should first \textbf{focus on answering the questions where they felt confident} and only if they had time to spare, (2) go back to the flagged questions and try to solve them. Questions could be \textbf{flagged as either ``too difficult'' or ``not clear or ambiguous''}.
    \item Finally, they were allowed to consult the annotation guide at any time.
\end{enumerate}

When the training was complete, their work was marked by two of the main authors of this paper and sent back to the annotators who were then asked to review their answers in order to learn from their mistakes.

\subsubsection{Annotation Process}

As shown in Table~\ref{ann_phase_info}, we followed a two-phase iterative approach.
Phase 1 included a small batch from the T2, T3 and T4's \textit{annotation set}.
Note that T1 data was excluded due to necessary revisions based on training feedback (see Section~\ref{subset_task1}).
Once completed, explanations underwent the same review process as those used during the annotation training.
Our training scheme proved to be effective, resulting in minimal necessary corrections to the annotations.
Phase 2 included the remaining instances in the \textit{annotation set}.

\begin{table}[h]
\centering
\small
\begin{tabular}{lllll}
\toprule
{\bf Phase} & {\bf T1}& {\bf T2}& {\bf T3}& {\bf T4}\\\midrule
1 & 0 & 28 & 28 & 28\\
2 & 110 & 82 & 82 & 82\\\midrule
Total & 110 & 110 & 110 & 110\\
\bottomrule 
\end{tabular}
\caption{\label{ann_phase_info} Distribution of task instances across each annotation phase.}
\end{table}

Annotators generally adhered to the allocated time frame of 5 minutes per instance, which translated to approximately 7 hours of annotation in Phase 1 and 30 hours in Phase 2.
Upon completion, their files were marked and formatted as a JSON file.

\subsubsection{Follow-up Survey}
\label{subsubsec:survey}

After completing the annotation, we asked the annotators to take a brief follow-up survey. We collected task load data for each of the four tasks using all six NASA-TLX items on a 9-point scale (1-10)~\citep{hart1988development, hart2006nasa}. We considered the items individually, as well as their sum, as has been done in prior work (e.g., \citeauthor{quinn_cost-benefit_2016},\citeyear{quinn_cost-benefit_2016}; \citeauthor{arnold_predictive_2020}, \citeyear{arnold_predictive_2020}). 

\begin{figure*}
    \centering
    \includegraphics[width=\linewidth]{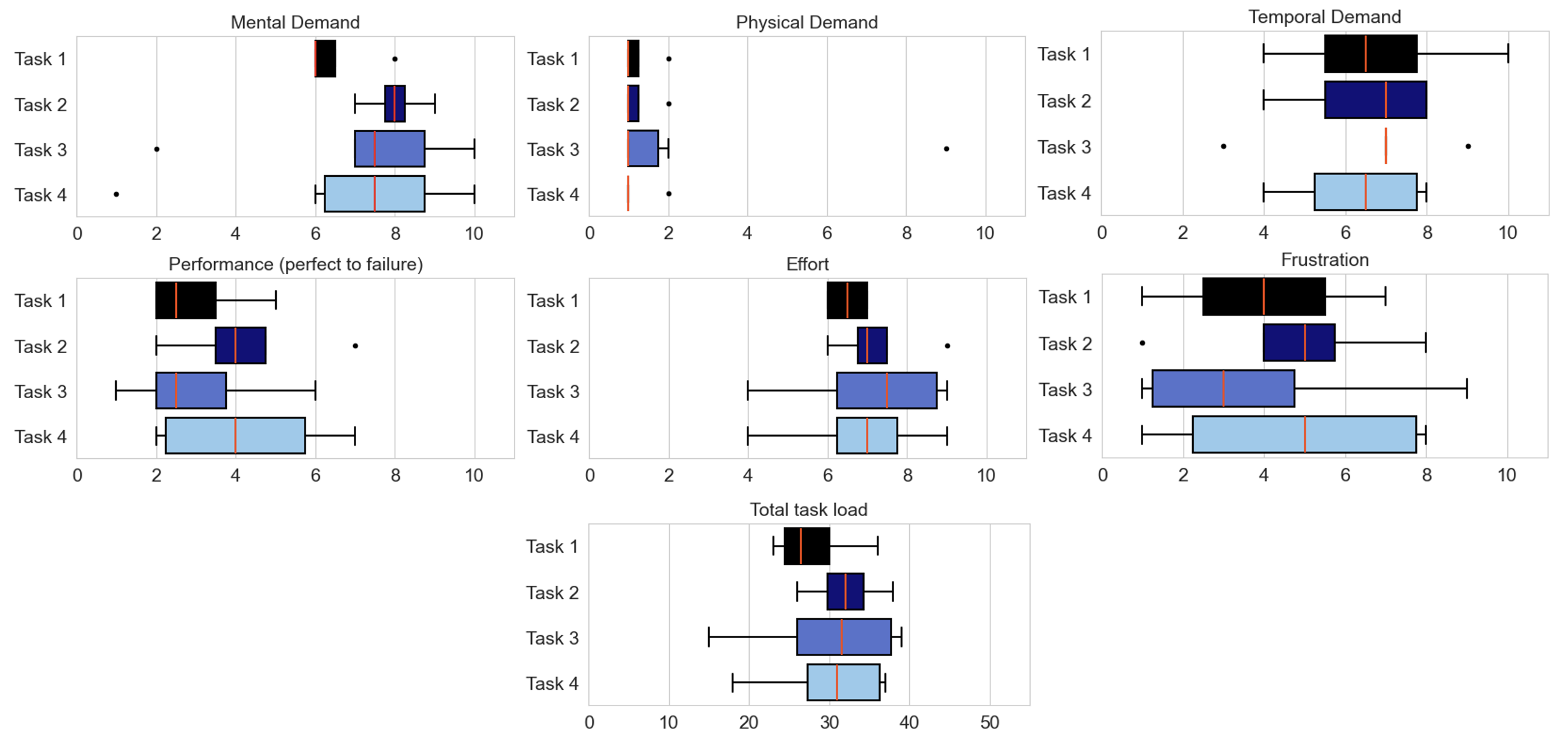}
    \caption{Box-plots of the six NASA-TLX items on a 9 point scale and their sum total. The median is shown in red.}
    \label{fig:workload}
\end{figure*}

Figure~\ref{fig:workload} shows box-plot representations of the responses from the NASA-TLX surveys, on which we performed Friedman tests \cite{friedman_comparison_1940} using the \texttt{friedmanchisquare} function of the \texttt{scipy} Python library \cite{virtanen_scipy_2020}. Taking the accepted standard $\alpha = 0.05$ as the significance threshold~\citep{exposito-ruiz_statistical_2010}, we found significant differences for performance ($\chi^2 = 8.11$, $p$-value $= 0.044$) only. Note that the performance item in the NASA-TLX survey is framed as follows: ``How successful do you think you were in accomplishing the goals of the task set by the experimenter (or yourself)? How satisfied were you with your performance in accomplishing these goals?''. 
Hence, annotators generally reported a lower sense of accomplishment and satisfaction in T2 and T4, than in T1 and T3. 

In the survey, we also included the two open-ended questions to learn more about the annotators' individual approaches to writing the explanations: specifically, whether they had a particular audience in mind, and what they thought the purpose of the explanations was. We include the exact wording of the questions below:
\begin{itemize}
    \item[\textbf{Q1:}] \textit{The intended recipient of our writing shapes our choice of language and style. Different audiences have different expectations, knowledge levels, and interests.
    When writing your explanations, did you have a specific audience in mind, or were you writing for a general audience?}
    \item[\textbf{Q2:}] \textit{Explanations can serve a range of purposes: (1) provide an understanding of why a choice was made, (2) justify how that choice was made by providing some evidence, (3) convince others that the choice was correct, and (4) other. When writing your explanations, what were you trying to achieve?}
\end{itemize}

In response to \textbf{Q1}, some annotators reported targeting a ``specific'' audience, such as researchers or students. On the other hand, one annotator explicitly aimed for a general audience. Others assumed an educated readership with basic linguistic knowledge of English without necessarily being specific about who they might be. Notably, one annotator expressed frustration towards the lack of clarity regarding the intended readership. The diversity in the annotators' conceptual audiences is very much echoed in the variety of tones used and the level of depth of the explanations we collected (refer to Table \ref{table:good_bad_examples} for example).

In response to \textbf{Q2}, five out of the six annotators that completed the survey chose (1) as their intended purpose which roughly matches our idea of what a \textsc{commentary} should do. The remaining annotator sought to justify their choice with evidence (2). While annotators assumed similar strategies, it is interesting to see that they in fact often went well beyond simply providing an understanding of why a choice was made and provided a majority of \textsc{justifications} instead (see Figure \ref{fig:quality_freq}).

\subsection{LLM Annotators}
\label{subsec:llm_annotation}

Six different models were used to generate annotations. They were chosen based on coverage of different model sizes, architectures and diversity of sources:
\begin{itemize}
    \item \textit{Llama-3.1-8B-Instruct}\footnote{\url{https://huggingface.co/meta-llama/Llama-3.1-8B-Instruct}} belongs to the family of Llama3.1 models published by Meta AI under the Llama3 community license. It incorporates a context window of 128k length and is pre-trained on a corpus of about 15 trillion tokens. 
    \item \textit{gemma-2-9b-it},\footnote{\url{https://huggingface.co/google/gemma-2-9b-it}} a lightweight open-source model from Google that also supports a 128k length context window. It was trained on 8 trillion tokens of data covering web documents, code, mathematics and more.
    \item \textit{Mixtral-8x7B-Instruct-v0.1},\footnote{\url{https://huggingface.co/mistralai/Mixtral-8x7B-Instruct-v0.1}} a pre-trained generative, sparse, mixture of experts model from Mistral AI. It has a context window of 32k tokens and is pre-trained on data extracted from open web.
    \item  \textit{c4ai-command-r-plus-08-2024}\footnote{\url{https://huggingface.co/CohereForAI/c4ai-command-r-plus-08-2024}} is a 104B parameter multilingual model released from Cohere For AI. It supports a context length of 128K.
    \item \textit{GPT-4o},\footnote{\url{https://openai.com/index/hello-gpt-4o/}} a multimodal model from OpenAI capable of processing and generating text, images, and audio. The parameter count of \texttt{GPT-4o} has not been publicly disclosed.
    \item \textit{Claude 3.5 Sonnet (claude-3-5-sonnet-20240620)},\footnote{\url{https://www.anthropic.com/news/3-5-models-and-computer-use}} an LLM model from Anthropic with improvements in reasoning, language understanding, and coding. The parameter count of Claude 3.5 Sonnet has not been publicly disclosed.

\end{itemize}

All open-source models were run on NVIDIA A100 GPUs using bf16 precision. We used the latest checkpoints of all open-weight models available at the time of the experiment, along with the default pretrained tokenizers provided for each model. A temperature of 0 was used for all models, including \texttt{Sonnet 3.5} and \texttt{GPT-4o}, which we accessed via API (for some HuggingFace models, we used 0.01 or set \texttt{do\_sample=False} due to implementation constraints).

\subsubsection{Prompts for Eliciting Explanations}
\label{subsubsec:prompt_design}
To elicit explanations from the model, we use a structured prompting approach. Each dataset is associated with a specific prompt designed to guide the model in generating explanations. Additionally, all prompts are preceded by a common system prompt:

\input{prompts/evaluation/system}

Below, we present the prompts used for each dataset.

\subsection{HellaSwag Prompt}
Each model was given \texttt{4} examples to guide its responses. For brevity, these examples are omitted from the prompt shown below.
\input{prompts/evaluation/hellaswag}

\subsection{RACE Prompt}
We provided \texttt{4} examples per query to improve model performance. The prompt format is shown below, excluding the examples for \textsc{conciseness}.
\input{prompts/evaluation/race}

\subsection{W\&I Prompt}
Models received \texttt{3} examples as part of the prompt structure. The displayed prompt excludes these examples for clarity.
\input{prompts/evaluation/wandi}

\subsection{Logic Prompt}
Each model was given \texttt{7} examples to guide its responses. For brevity, these examples are omitted from the prompt shown below.
\input{prompts/evaluation/logic}

\section{Custom Agreement Metric}
\label{sec:agreement_metric}
\textbf{First metric.} 
Cohen's $\kappa$~\citep{Cohen1960ACO} and Krippendorff's $\alpha$~\citep{Krippendorff2011ComputingKA} are among the most frequently used inter-rater reliability metrics.
However, their direct application is best suited to nominal or categorical data.
Even with adaptations like weighted kappa, these coefficients struggle to capture the full inter-relationship of hierarchical nested data.
To bridge this gap, we introduced a custom metric that specifically accounts for the nested dependencies in CUBE.
Our custom metric accounts for the \textit{superlabels} (\textsc{none}, \textsc{commentary}, \textsc{justification}, \textsc{argument}) and \textit{sublabels} (i.e., all \textsc{dimensions}) in Rubrik.
In both cases, the metric penalises discrepancies between ratings, with the penalty proportional to the difference in the hierarchical level.
For example, consider the cases shown in Table~\ref{table:superlabel_metric} and Table~\ref{table:sublabel_metric}.

\begin{table}[h]
\centering
\begin{adjustbox}{width=1\columnwidth}
\begin{tabular}{lllll}
\toprule
{\bf Case} & {\bf Rater 1} & {\bf Rater 2} & {\bf Diff.} & {\bf Agree. (\%)}\\\midrule
1 & \textsc{commentary} & \textsc{justification} & 1 & 67\\
2 & \textsc{commentary} & \textsc{argument} & 2 & 50\\
3 & \textsc{none} & \textsc{argument} & 3 to 4 & 0 to 25\\
\bottomrule 
\end{tabular}
\end{adjustbox}
\caption{Superlabel agreement. \textsc{none} denotes the case where either of the \textsc{commentary}'s \textsc{components} are missing, namely Action (1.a) and Reason (1.b).}
\label{table:superlabel_metric}
\end{table}

From the \textit{superlabel} point of view, there is a partial agreement in Case 1 since a \textsc{justification} has the two components (\textsc{action} and \textsc{reason}) of a \textsc{commentary} and an additional one (namely, \textsc{evidence}).
Thus, the difference in the raters' judgement is 1.
From the \textit{sublabel} point of view, the agreement range is higher as it takes into consideration all the elements of a \textsc{commentary} (8: 2 \textsc{components}, 6 \textsc{dimensions}) and a \textsc{justification} (10: 3 \textsc{components}, 7 \textsc{dimensions}).

\begin{table}[h]
\centering
\begin{adjustbox}{width=1\columnwidth}
\begin{tabular}{lllll}
\toprule
{\bf Case} & {\bf Rater 1} & {\bf Rater 2} & {\bf Diff.} & {\bf Agree. (\%)}\\\midrule
1 & \textsc{commentary} & \textsc{justification} & 1-8 of 10 & 90-20\\
2 & \textsc{commentary} & \textsc{argument} & 4-10 of 12 & 66-17\\
3 & \textsc{none} & \textsc{argument} & 11-12 of 12 & 8-0\\
\bottomrule 
\end{tabular}
\end{adjustbox}
\caption{Sublabel agreement. The difference (Diff.) column shows a range, taking both \textsc{components} and \textsc{dimensions} into consideration.}
\label{table:sublabel_metric}
\end{table}

As explained in Section \ref{subsec:scoring_strategy}, a good \textsc{commentary} is the base of a good \textsc{justification}.
This means that Rater 2 judged with met (\ding{51}) all the elements of a \textsc{commentary}.
The disagreement with Rater 1 comes from them judging with not met (\ding{55}) one or more of the six dimensions.
The same logic applies to Cases 2 and 3.\\

\begin{table*}[t]
\centering
\small
\resizebox{\textwidth}{!}{
\begin{tabular}{llr|rrrr|rr}
\toprule
 &  &  & \multicolumn{4}{c}{\bf Open models} & \multicolumn{2}{c}{\bf Closed Models} \\
\bf Task & Agreement & Humans & Llama 3.1 & \texttt{Gemma 2} & \texttt{Command R+} & \texttt{Mixtral} & \texttt{GPT-4o} & \texttt{Sonnet 3.5} \\\hline
\multirow{2}{*}{\bf T1} & Superlabel & 0.814 & 0.693 & 0.799 & 0.797 & \textbf{0.812} & 0.794 & \textbf{0.800}\\
 & Sublabel & 0.823 & 0.706 & 0.795 & 0.826 & \textbf{0.829} & 0.807 & \textbf{0.811}\\\hline
 \multirow{2}{*}{\bf T2} & Superlabel & 0.910 & 0.832 & 0.862 & \textbf{0.873} & 0.869 & 0.878 & \textbf{0.879}\\
 & Sublabel & 0.923 & 0.865 & 0.888 & \textbf{0.903} & 0.898 & \textbf{0.902} & 0.899\\\hline
 \multirow{2}{*}{\bf T3} & Superlabel & 0.830 & 0.830 & 0.838 & 0.843 & \textbf{0.847} & 0.844 & \textbf{0.854}\\
 & Sublabel & 0.869 & 0.862 & 0.866 & 0.881 & \textbf{0.887} & 0.872 & \textbf{0.881}\\\hline
 \multirow{2}{*}{\bf T4} & Superlabel & 0.887 & 0.797 & \textbf{0.817} & 0.810 & 0.774 & \textbf{0.846} & 0.833\\
 & Sublabel & 0.897 & 0.807 & 0.804 & \textbf{0.853} & 0.787 & \textbf{0.860} & 0.851\\\hline
 \multirow{2}{*}{\bf Overall} & Superlabel & 0.860 & 0.788 & 0.829 & \textbf{0.831} & 0.825 & 0.841 & \textbf{0.842}\\
 & Sublabel & 0.878 & 0.810 & 0.838 & \textbf{0.866} & 0.850 & \textbf{0.860} & \textbf{0.860}\\
\bottomrule 
\end{tabular}
}
\caption{\label{table:score_one_results} Overview of agreements scores, calculated with the first metric. In bold, the highest score by superlabel and sublabel, comparing the performance of open- vs. closed-source models.}
\end{table*}

 \begin{table*}[t]
   \centering
   \resizebox{\textwidth}{!}{
       \begin{tabular}{lrrrrrr}
         \toprule
         \textbf{Annotator}        &  \textbf{\textsc{None}} &  \textbf{\textsc{Commentary}} &  \textbf{\textsc{Justification}} &  \textbf{\textsc{Argument}} &  \textbf{Second-metric-score} & \textbf{Second-metric-rank}\\
         \midrule
         \texttt{Human\_annotator 1} & 0                     & 293          & 406              & 221           & \text{-}
             & \text{-}\\
         \texttt{Human\_annotator 2}  & 5                     & 264          & 229              & 422          & \text{-}
             & \text{-}\\
         \texttt{LLama 3.1}           & 87                    & 47           & 450              & 336         & 0.405
             & 5\\\midrule
         \texttt{Gemma 2}           & \textbf{9}                     & \textbf{222}          & \textbf{561}              & \textbf{128}         & \textbf{0.464}
             & \textbf{2}\\\midrule
         \texttt{Command R+}             & 4                     & 20           & 894              & 2           & 0.346
             & 6\\
         \texttt{Mixtral}          & 5                     & 240          & 654              & 21          & 0.427
             & 4\\\midrule
         \texttt{GPT-4o}            & \textbf{14}                    & \textbf{107}          & \textbf{685}              & \textbf{114}         & \textbf{0.476}
             & \textbf{1}\\\midrule
         \texttt{Sonnet 3.5}        & 5                     & 126          & 742              & 47          & 0.444
             & 3\\
         \bottomrule
       \end{tabular}
     }
 \caption{\label{table:annotator_class_distribution} Aggregated label counts for each annotator and metric score. In bold are the results from the two best-ranked LLM evaluators. In both cases, there is a better balance in the judgement of explanation types.}
 \end{table*}

\noindent\textbf{Second metric.} The first agreement metric accounts for partial agreement between LLMs and human annotators. 
We tested all LLMs as evaluators on the same subset judged by humans.
However, we observe that LLMs often rate an explanation as \textsc{justification} over the other options, compromising their ability to detect other types (see Table~\ref{table:annotator_class_distribution}).
This highlighted the need for an additional custom metric, which we designed based on a weighted F1 score to penalise over-centralization on a single label. 
The class weights are derived from both human evaluations and LLM evaluations from all six models. 
In our approach, we first calculate the distribution percentage of each superlabel in human evaluation $p_{i}^{human}$ for label $i$. 
We then calculate the average distribution percentage of each superlabel across all 6 LLM evaluations denoted as $p_{i}^{LLM}$. 
These two percentages are combined as the class weight:
$$w_i = \lambda p_{i}^{human} + (1-\lambda)p_{i}^{LLM}$$
where $\lambda$ is a hyperparameter representing the relative importance of human evaluations vs. LLM evaluations. 
The derived class weights are then incorporated into the calculation of the weighted F1 score. 

As shown in Table \ref{table:score_one_results}, our first metric points to \texttt{Command R+} as the model with higher agreement with human evaluators.
However, a closer look at the distribution of the explanation types assigned show that the high agreement is due to identifying an explanation as \textsc{justification} nearly always.
Our second metric penalises this behaviour, ranking \texttt{Command R+} as the least effective evaluator.

\section{Rubric Evaluation Prompts}
\label{app:rubric_scoring}

To evaluate explanations generated by the model, we use a structured prompting approach based on a rubric. Each dataset is associated with a specific prompt designed to guide the model in assessing explanations. 
Below is the prompt template that encodes the evaluation rubric.

Note that the prompt \textbf{does not} ask the model to judge whether an explanation is \usym{1F60A} \textit{good} or \usym{1F641} \textit{bad}.
This choice reflects the insights of \citet{panickssery2024llm}, who found that out-of-the-box LLMs, such as GPT-4 and Llama 2, have non-trivial (over 50\%) accuracy at distinguishing themselves from other LLMs and humans. 
As a result, these models tend to recognise and favour their own generations. 
Thus, our prompt only specifies the evaluation criteria to decide whether a given \textsc{component} or \textsc{dimension} is met (\ding{51}) or not met (\ding{55}).
This approach successfully mitigated self-preference; \texttt{GPT-4o}, our third evaluator, judged its own outputs as \textit{bad} at a comparable low rate to other models' outputs.
Recall from Section \ref{subsec:scoring_strategy} that an explanation is deemed good if, and only if, it meets \underline{all} the criteria.
While this condition establishes a rigorous baseline for assessment, this scoring strategy is flexible. 
The specific conditions can be adjusted to fit varying research objectives or contextual needs.

\input{prompts/rubric/base}

\section*{Dataset-Specific Evaluation Prompts}
In the above template, the main difference between datasets is the format of the question and the options. Below, we show how each dataset-specific question and option block is customised.

\subsection{HellaSwag}
\input{prompts/rubric/hellaswag}

\subsection{RACE}
\input{prompts/rubric/race}

\subsection{WANDI}
\input{prompts/rubric/wandi}

\subsection{Logic}
\input{prompts/rubric/logic}

\section{Detailed Analysis Results}
\label{app:detailed_analysis}

This section delves deeper into the data, offering additional insights to complement the summary provided in Section \ref{sec:discussion}.

\subsection{Answer Frequencies}
\label{app:answer_freqs}

\begin{figure*}
    \centering
    \includegraphics[width=1\linewidth]{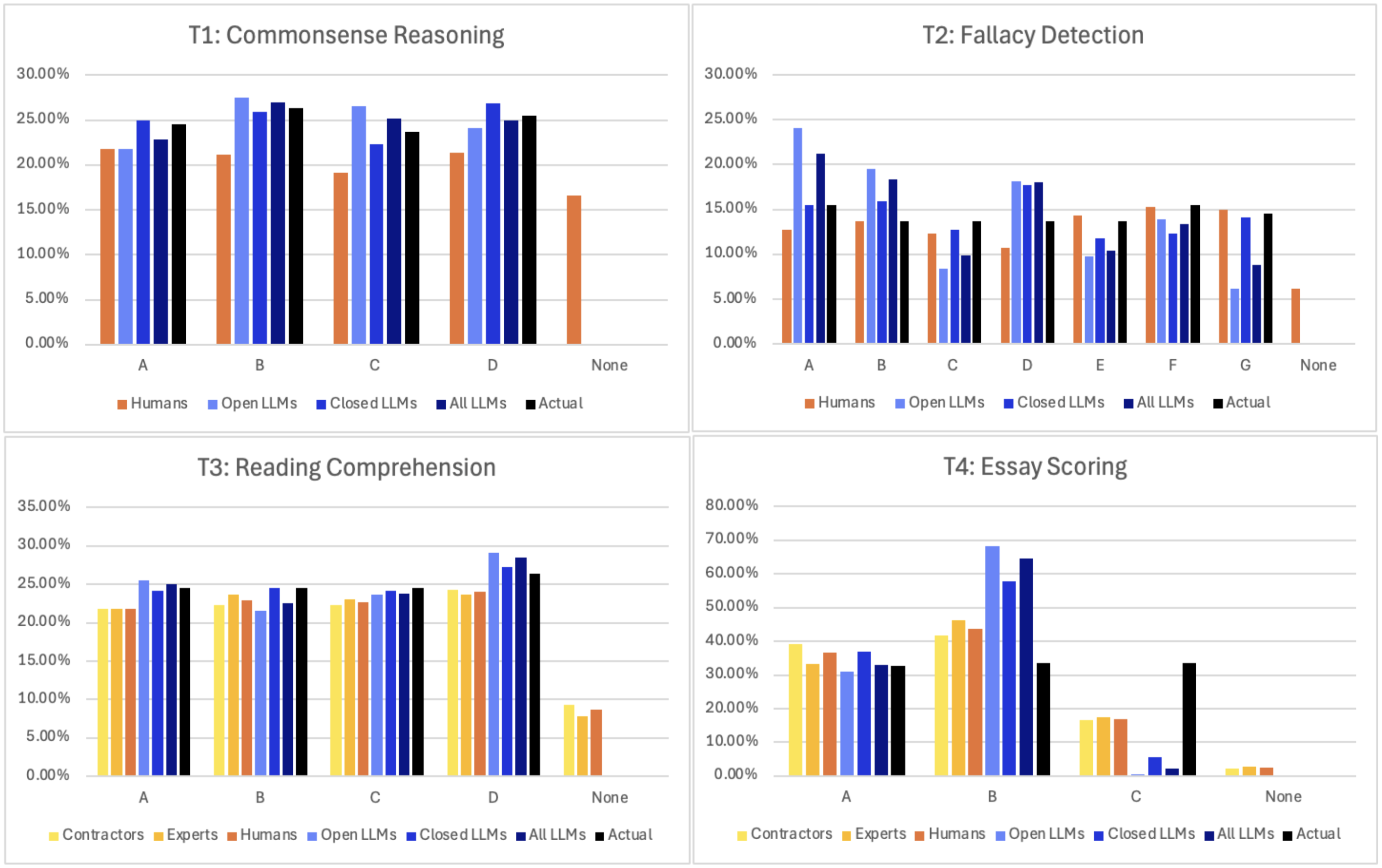}
    \caption{Frequencies of the answers picked by the different groups of annotators during the annotation phase. We also show the \textbf{Actual} distribution of correct answers in black in the \textit{annotation set.}}
    \label{fig:answer_frequencies}
\end{figure*}

First, we report the frequencies of the answer choices picked by different groups of annotators during the annotation phase, and compare these to the actual distribution of correct answers in each task on the \textit{annotation set} in Figure \ref{fig:answer_frequencies}. Recall that we explicitly tried to get as uniform a distribution across the different answer choices as possible in the \textit{annotation set} (as described in Appendix \ref{sec:selection}). 

Overall, we note that while human annotators sometimes refused to choose an answer between those provided (\textsc{None}), the LLMs almost never refused to answer. This may be because LLMs have a tendency to overestimate their ability to answer questions \citep{zhang_sirens_2023}.

In T1 and T3, the answer frequencies of all annotators seem fairly balanced, with the only notable difference being that human annotators also responded \textsc{None}. 
In T2, however, we can see that the grouped Open LLMs (\texttt{Command R+}, \texttt{Mixtral}, \texttt{Llama 3.1} and \texttt{Gemma 2}) seem to significantly favour answers A, B and D at the expense of answers C and G, while the other groups of annotators remain relatively close to the actual frequency distribution. 
We should note that despite the fact that the \textit{annotation set} is more or less balanced, in \citet{jin_logical_2022} authors state that more than a single fallacy type may apply to a single instance.
This may explain the variation observed. 
Specifically, they identified ``common among incorrect but reasonable predictions'' in their task, which ``are debatable cases where multiple logical fallacy types seem to apply''.

In T4, we notice a stark difference between humans and LLMs annotators. 
On one hand, LLMs almost never assign C (advanced) scores to essays, and overwhelmingly assign B (intermediate) scores around 65\% of the time. 
While human annotators use the whole range of the scale, though still showing signs of a strong central tendency or severity by only assigning around half the actual proportion of advanced scores. 
Interestingly, experts annotators, that are professionally trained to assess the work of language learners, did not distinguish themselves from the contractors we hired who had very similar frequency distributions in the two language tasks. Overall, evaluators failed to identify advanced essays, focusing most of their attention on the middle of the rating scale. 
Essay scoring is a notoriously complex and subjective task \citep{brown_validity_2010}, and we intentionally did not provide any scoring rubric to the annotators. 
They thus lacked a proper point of reference for the scale, which seems to be the source of the frustration reported by one annotator (see Section \ref{subsubsec:survey}).

\subsection{Detailed Accuracy}
\label{app:results_acc}

\begin{figure*}
\centering
\begin{subfigure}{0.5\textwidth}
\centering
    \includegraphics[width=\linewidth]{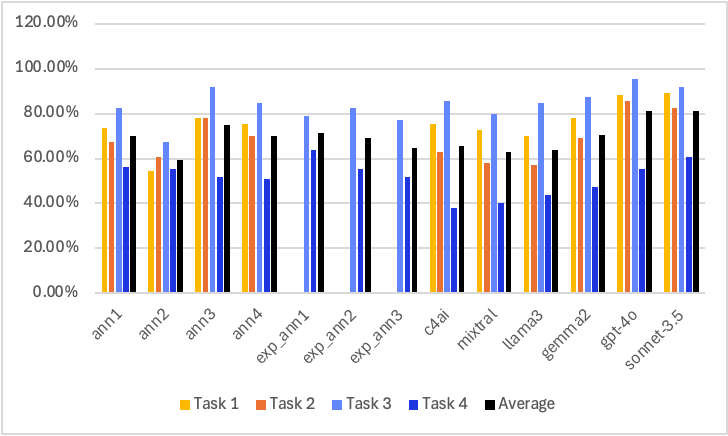}
    \caption{}
    \label{fig:acc_individual}
\end{subfigure}
\begin{subfigure}{0.5\textwidth}
\centering
    \includegraphics[width=\linewidth]{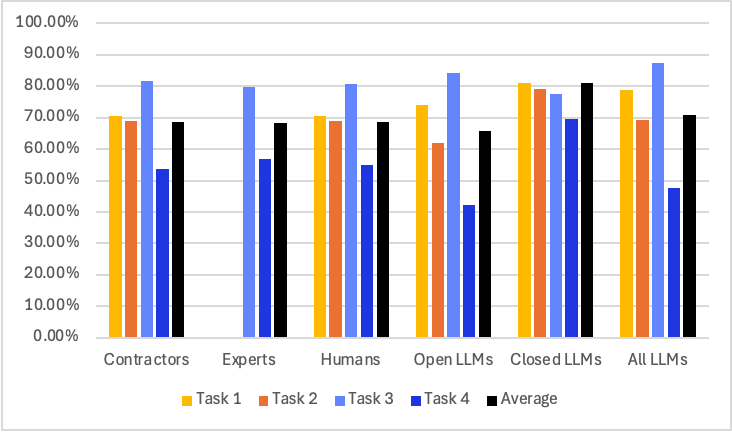}
    \caption{}
    \label{fig:acc_grouped}
\end{subfigure}
\caption{Accuracy results of the different annotators in each of the tasks. On the left, \ref{fig:acc_individual} shows the individual annotator performance, and on the left, \ref{fig:acc_grouped} shows the performance by group of annotators. We also include the \textbf{Average} accuracy across the four tasks of each annotator or group in black.}
\label{fig:accuracy_res}
\end{figure*}

Next, in Figure \ref{fig:accuracy_res} we report the performance or accuracy (\%) of the individual annotators and their groups, in each of the tasks, as well as their overall average performance across the four tasks.

Looking at the average performance across the four tasks, closed LLMs seem to perform the best, while open LLMs perform the worst, with humans (contractors and experts) performing just slightly better than the open models. 
The two closed models exhibited comparable average performance across the four tasks, but \texttt{Sonnet 3.5} is more consistently good across the four tasks, whereas \texttt{GPT-4o} is very good at Reading Comprehension (T3) and less good at Essay Scoring (T4). 

Overall, these graphs make it apparent that Essay Scoring (T4) was the hardest with an average accuracy of roughly 52\% (across all annotators), while Reading Comprehension (T3) was by far the easiest with an average accuracy reaching almost 84\%.

As in the previous section, we note that humans were overall quite consistent. The experts were ever so slightly better at Essay Scoring (T4) than the contractors, but this difference is very small. We had expected them to do much better due to being professionally trained to perform language assessment tasks. Further, while this background should have directly impacted their capacity to do well in T4, we also expected them to do better than the contractors in T3 given the language-related nature of their day-to-day work. 
However, contractors were in fact ever so slightly better at Reading Comprehension (T3). 
These findings suggest that we do not always necessarily need to hire professionals, and that professional expertise can be matched by a rigorous selection process and sufficient training of annotators.

\begin{figure*}
    \includegraphics[width=\textwidth]{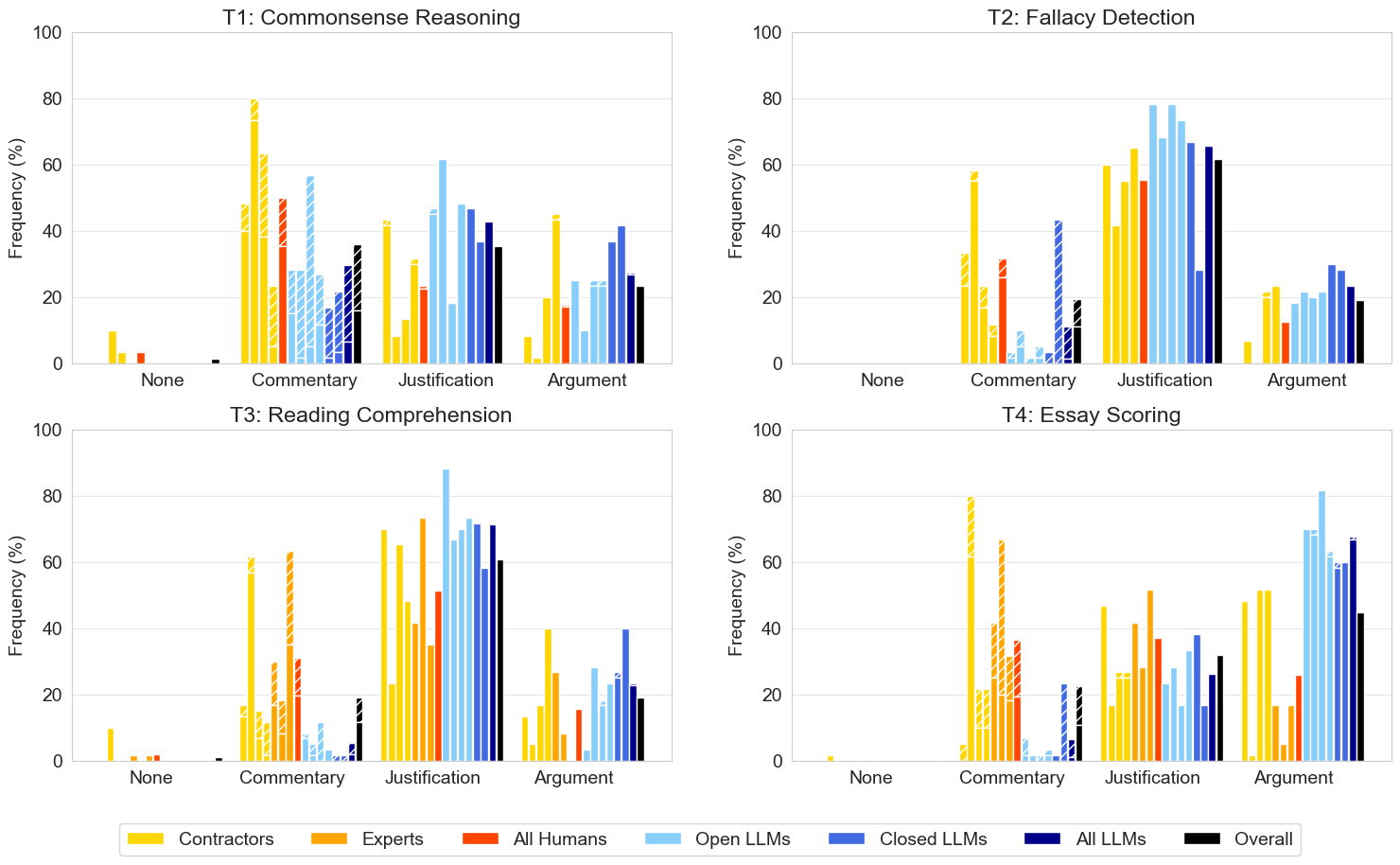}
    \caption{A breakdown of the bar plots in Figure \ref{fig:quality_freq} which shows the frequencies (\%) of the different explanation types for each individual human and LLM annotator (4 contractors, 3 experts in \textbf{T3} and \textbf{T4}, 4 open-models---\texttt{Command R+}, \texttt{Mixtral}, \texttt{LLama 3.1}, \texttt{Gemma 2}---and 2 closed-models---\texttt{GPT-4o} and \texttt{Sonnet 3.5}---in this order) in the \textit{evaluation set}. We also include the average frequencies across all annotators (in black). We average the frequencies across all three evaluators (two humans and \texttt{GPT-4o}).}
    \label{fig:quality_freq_detailed}
\end{figure*}

\begin{figure*}
    \includegraphics[width=\textwidth]{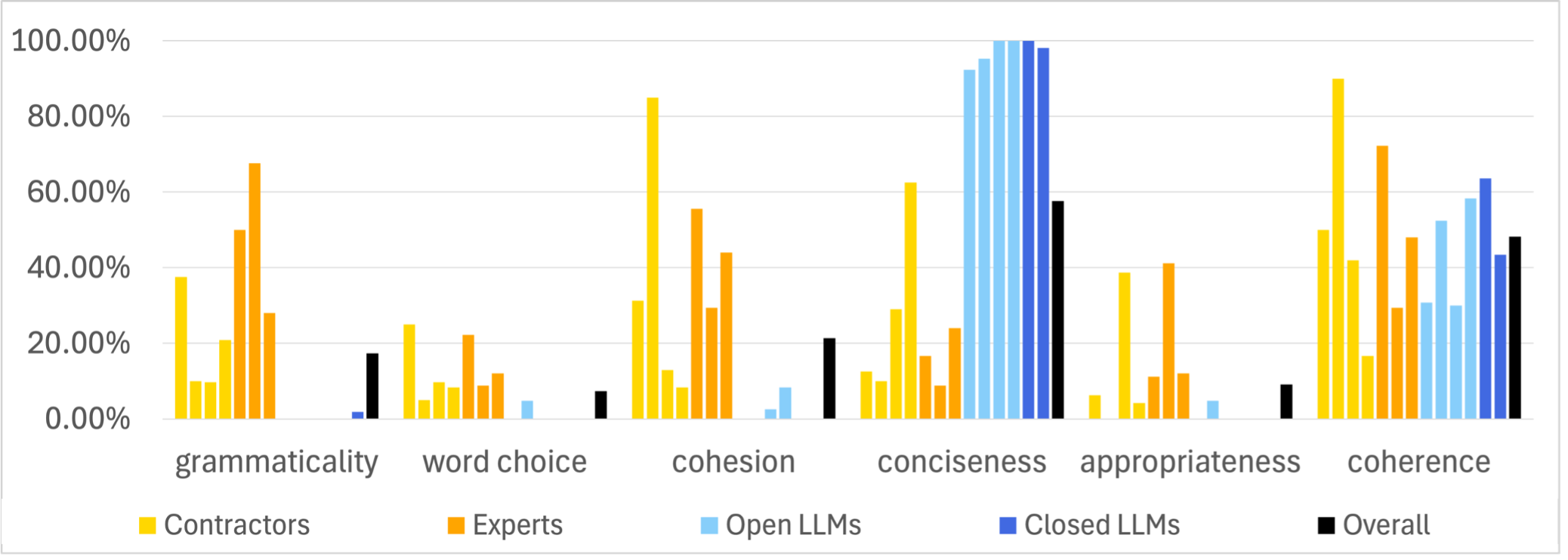}
    \caption{A breakdown of Figure \ref{fig:sublabels} which shows the sources of the \textit{bad} \textsc{commentaries} for each invidual human and LLM annotator (4 contractors, 3 experts in \textbf{T3} and \textbf{T4}, 4 open-models---\texttt{Command R+}, \texttt{Mixtral}, \texttt{LLama 3.1}, \texttt{Gemma 2}---and 2 closed-models---\texttt{GPT-4o} and \texttt{Sonnet 3.5}---in this order) in the \textit{evaluation set}. We also include the average frequencies across all annotators (in black). We average the frequencies across all three evaluators (two humans and \texttt{GPT-4o}).}
    \label{fig:sublabels_detailed}
\end{figure*}

\end{document}

%% file: prompts/evaluation/system.tex
\begin{quote}
You are a helpful, pattern-following assistant. Use the following instructions to respond to user inputs.
1. Start your answer with a prefix that says "The right answer is: ".
2. Explain the response given in Step 1, with a prefix that says "Because: ". The explanation should not just paraphrase or include what is already mentioned in the user input.
3. Show all the answer choices with their numeric probability of being the correct answer
\end{quote}

%% file: prompts/evaluation/hellaswag.tex
\begin{lstlisting}
## Examples

Please choose the most plausible ending (event) for the given context. There is only **one** correct answer. After selecting a correct answer, explain why you selected that option. The examples do not include an explanation but you will need to provide it when answering the question.

For reference, we provide below four examples that have already been solved for you.

{%
**Example {{loop.index}}**
{{example}}
{%

## Exercise

Context: {ctx_a}

Question: Choose the option that best completes the above story.

Options:
{%
{{ 'ABCD'[loop.index0] }}) {{ctx_b}} {{ ending }}
{%
\end{lstlisting}

%% file: prompts/evaluation/race.tex
\begin{lstlisting}
## Examples

In this task, you will be presented with a series of articles. Each is followed by a question which relates to the information provided in the text, and four possible answers. Select only **one** of these options as the correct answer, and explain your choice.

For reference, we provide below four examples that have already been solved for you.

{%
**Example {{loop.index}}**
{{example}}
{%

# Exercise

Article: {article}

Question: {question}

Options: 
{%
{{ 'ABCD'[loop.index0] }}) {{ option }}
{%
\end{lstlisting}

%% file: prompts/evaluation/wandi.tex
\begin{lstlisting}
# Task

In this task, you will be presented with a series of essays. Annotate each of these with exactly **one** of three grades: A (beginner), B (intermediate), C (advanced), and then explain your choice. 

For reference, we provide below three examples that have already been solved for you.

## Examples
{%
**Example {{loop.index}}**
{{example}}
{%

## Exercise

Essay: {{full_text}}

Question: If you were to assign a grade to this essay, what would it be?

Options: 

1. Beginner (grade A)  
2. Intermediate (grade B)  
3. Advanced (grade C)
\end{lstlisting}

%% file: prompts/evaluation/logic.tex
\begin{lstlisting}
## Examples
Please identify the type of logical fallacy. There is only **one** correct answer. After selecting a correct answer, explain why you selected that option. 

For reference, we provide below seven examples that have already been solved for you.

{%
**Example {{loop.index}}**
{{example}}
{%

## Exercise

Statement: {source_article}

Question: Which type of logical fallacy is this an example of?

Options:
A. Faulty generalisation  
B. False causality  
C. Circular claim  
D. Appeal to emotion  
E. Deductive fallacy  
F. False dilemma  
G. Fallacy of credibility
\end{lstlisting}

%% file: prompts/rubric/base.tex
\onecolumn
\begin{lstlisting}[basicstyle=\ttfamily, frame=single]
{# Base template for rubric scoring #}
 # Explanation Judging Task

Your task is to evaluate a set of explanations in a given context. We define the context (**Task**, **Audience**, and **Purpose**) in the following way:

**Task**: you will be shown a series of multiple-choice questions relating to one of four tasks (commonsense reasoning, fallacy detection, reading comprehension and essay scoring) in the following format:
1. **Question**: The question being answered.
2. **Answer Choices**: The possible answer choices for that question.
3. **Correct Answer**: The correct answer to the question.
4. **User Answer**: The answer provided by the user.
5. **Explanation**: The explanation provided by the user to support their answer.

**Audience**: you should assume that the audience of the explanations is adult, English-proficient, and provided in a formal academic setting.

**Purpose**: the explanations should provide an understanding of why a certain answer was chosen for a given multiple-choice question. 

---

## Evaluation Criteria

For the given explanation, please answer the following questions with either **Yes** or **No**. Note that you **should not consider the correctness of the user's answer** when evaluating the explanation. Focus solely on the quality of the explanation according to the criteria provided.

1. **Action**: Does the explanation clearly indicate the decision or choice being made (e.g., specifying the selected answer)?
    - Answer **Yes** if it does. For example "The correct answer is A."
    - Answer **No** if it does not. For example "Because it is the final part of the sequence."

2. **Reason**: Does the explanation provide reasoning or insight into why the decision or choice was made, explaining the underlying logic or rationale for the **Action**?
    - Answer **Yes** if it does. For example "The right answer is C because it is the final part of the sequence."
    - Answer **No** if it does not. For example "The correct answer is A."

3. **Grammaticality**: Is the explanation grammatically correct and free of lexical or syntax errors? Small typos are acceptable, but the errors should not impede comprehension in any way.
    - Answer **Yes** if it is. For example "The correct answer is A because nowadays our society is based on consumerism and the way in which we are producing is contaminating the world."
    - Answer **No** if it is not. For example "The correct answer is A because now a day our socity it is bassed in consumer, so that become the word more contaminate to produce the products that we demanding."
    
4. **Word Choice**: Is the language used in the explanation tailored to the given context (task, audience, purpose)? And are the sentences in the explanation well-formed?
    - Answer **Yes** if they are. For example "The correct answer is A because the essay lacks fluency. There are many incorrect clauses and missing words. And while the overall meaning can be deduced, the essay does not demonstrate an accurate grasp of language (e.g., frequent spelling and punctuation errors)."
    - Answer **No** if they are not. For example "Answer A. lack of fluency, incorrect clauses and missing words, meaning can be found but does not demonstrate an accurate grasp of language"

5. **Cohesion**: Does the explanation make appropriate use of transition phrases (e.g., connectives like "because", "therefore", "consequently", overlapping words across sentences, etc.)?
    - Answer **Yes** if it does. For example "The correct answer is C because the man is on roller blades, not on a skateboard. Further, he is not talking to anyone and therefore cannot possibly 'continue speaking.'"
    - Answer **No** if it does not. For example "The correct answer is C, because the man is on roller blades, not a skateboard, and is not talking to anyone in the example so cannot 'continue speaking'".

6. **Conciseness**: Is the explanation free of any redundant, irrelevant, or excess sentences (that is, not required to understand the answer)?
    - Answer **Yes** if it is. For example "The correct answer is D because it accurately reflects the sequence of events."
    - Answer **No** if it is not. For example, given that the option D was "next she explains how to use the lawnmower and other tools and then she cuts the grass", the following explanation is not concise: "The correct answer is D because the sentence mentions that she explains how to use the lawnmower and other tools, and then she cuts the grass. Option D accurately reflects the sequence of events." 

7. **Appropriateness**: Is the explanation culturally appropriate, matching expectations for the given context?    
    - Answer **Yes** if it is. For example "The right answer is B because the tenses are properly used and the story makes sense."
    - Answer **No** if it is not. For example "The right answer is B because the tenses are properly used and (within the slightly odd context) the story makes sense."

8. **Coherence**: Does the explanation appropriately transition between ideas? That is, does the explanation make sense as a whole (e.g., good context-relatedness, semantic consistency, and inter-sentence causal and temporal dependencies, etc.)? 
    - Answer **Yes** if it does. For example "The correct answer is D, because no information about Liu's relationship to science subjects specifically is given in the passage, therefore the fact that they like chemistry is implied and ambiguous."
    - Answer **No** if it does not. For example "The correct answer is D, because no information about Liu's relationship to science subjects specifically is given in the passage, therefore the fact that they like cheese is implied and ambiguous."

9. **Evidence**: Does the explanation provide concrete evidence (can be both explicit or implicit) that supports the reasoning, such as information from the question's context or general knowledge?
    - Answer **Yes** if it does. For example "The right answer is C, because it finishes the sequence, describing the effect of bowling the ball and what happens as a result."
    - Answer **No** if it does not. For example "The right answer is C, because is is the final part of the sequence."

10. **Plausibility (of the evidence)**: Is the provided evidence plausible and consistent with human reasoning, considering the context and general world knowledge?
    - Answer **Yes** if it is. For example "The correct answer is A ('Jack picks the cheese') because we are told that he enjoys eating 'mozzarella' in the morning."
    - Answer **No** if it is not. For example "The correct answer is A ('Jack picks the cheese') because my name is also Jack and I personally love cheese for breakfast."

11. **Affective Appeals**: Does the explanation use vivid, or emotionally charged language (e.g., metaphors) to evoke feelings in the audience?
    - Answer **Yes** if it does. For example "The expression in the final section is very heartfelt; the tone is excitable and keen throughout."
    - Answer **No** if it does not. For example "The final section reflects the writer's strong feelings on this issue."

12. **Qualifiers**: Does the explanation make use of hedges, boosters, attitude markers, self-mentions, or engagement markers to clarify the writer's stance (i.e., the explainer's personal feelings towards the task)? Note that the stance can be implicit unlike the **Action**.
    - Answer **Yes** if it does. For example "The right answer is B, because the text is keeping with what is presumably a tour guide's voice: intentionally using clunky and overly expressive words."
    - Answer **No** if it does not. For example "The right answer is B, because the text is keeping with the original tour guide's voice."

13. **Stance Clarity**: Is the explainer's stance (their personal feelings towards the task) clearly and unambiguously conveyed through affective appeals or qualifiers? Note that the stance can be implicit unlike the Action.
    - Answer **Yes** if it is. For example "The correct answer is A (beginner) because this text is undeniably of a low English level."
    - Answer **No** if it is not. For example "The correct answer is A (beginner) because this text is clearly of a low English level although the final section is incredibly well written."

---

## Expected Output

Your answers should be formatted as follows:

1. Action: **Yes** or **No**
2. Reason: **Yes** or **No**
3. Grammaticality: **Yes** or **No**
4. Word Choice: **Yes** or **No**
5. Cohesion: **Yes** or **No**
6. Conciseness: **Yes** or **No**
7. Appropriateness: **Yes** or **No** 
8. Coherence: **Yes** or **No**
9. Evidence: **Yes** or **No**
10. Plausibility: **Yes** or **No**
11. Affective Appeals: **Yes** or **No**
12. Qualifiers: **Yes** or **No**
13. Stance Clarity: **Yes** or **No**

---

## Question

{%

{{ task_question }}

{%

## Answer Choices

{%

{%
{{ 'ABCDEFG'[loop.index0] }}) {{ choice }}
{%

{%

## Correct Answer
{{correct_answer}}

## User Answer
{{user_answer}}

## Explanation
{{explanation}}

\end{lstlisting}
\twocolumn

%% file: prompts/rubric/hellaswag.tex
\begin{lstlisting}[basicstyle=\ttfamily, frame=single]
{%

{%

{{ ctx_a }}

{%

{%

{%

{{ 'ABCD'[loop.index0] }}) {{ctx_b}} {{ ending }}

{%

{%
\end{lstlisting}

%% file: prompts/rubric/race.tex
\begin{lstlisting}[basicstyle=\ttfamily, frame=single]
{%

{%

Article: {text}

Question: {question}

{%

\end{lstlisting}

%% file: prompts/rubric/wandi.tex
\begin{lstlisting}[basicstyle=\ttfamily, frame=single]
{%

{%
Essay: {text}
{%

{%

1. Beginner (grade A)  
2. Intermediate (grade B)  
3. Advanced (grade C)

{%
\end{lstlisting}

%% file: prompts/rubric/logic.tex
\begin{lstlisting}[basicstyle=\ttfamily, frame=single]
{%

{%

Statement: {{text}}

Question: {{question}}

{%

{%

A. Faulty generalisation  
B. False causality  
C. Circular claim  
D. Appeal to emotion  
E. Deductive fallacy  
F. False dilemma  
G. Fallacy of credibility

{%
\end{lstlisting}

%% file: acl_latex_arxiv.bbl
\begin{thebibliography}{128}
\providecommand{\natexlab}[1]{#1}

\bibitem[{Agarwal et~al.(2024)Agarwal, Tanneru, and Lakkaraju}]{agarwal_faithfulness_2024}
Chirag Agarwal, Sree~Harsha Tanneru, and Himabindu Lakkaraju. 2024.
\newblock \href {https://doi.org/10.48550/arXiv.2402.04614} {Faithfulness vs. {Plausibility}: {On} the ({Un}){Reliability} of {Explanations} from {Large} {Language} {Models}}.
\newblock \emph{arXiv preprint}.
\newblock ArXiv:2402.04614 [cs].

\bibitem[{Alley(1996)}]{alley_language_1996}
Michael Alley. 1996.
\newblock \href {https://doi.org/10.1007/978-1-4757-2482-0_8} {Language: {Being} {Concise}}.
\newblock In Michael Alley, editor, \emph{The {Craft} of {Scientific} {Writing}}, pages 119--127. Springer, New York, NY.

\bibitem[{Amiryousefi and Barati(2011)}]{amiryousefi2011metadiscourse}
Mohammad Amiryousefi and Hossein Barati. 2011.
\newblock Metadiscourse: exploring interaction in writing, ken hyland.
\newblock \emph{Continuum, London. Elixir Literature}, 40:5245--5250.

\bibitem[{Anthropic(2024)}]{anthropic_claude_2024}
Anthropic. 2024.
\newblock \href {https://www.anthropic.com/news/3-5-models-and-computer-use} {Introducing computer use, a new claude 3.5 sonnet, and claude 3.5 haiku}.
\newblock Accessed: February 2025.

\bibitem[{Arnold et~al.(2020)Arnold, Chauncey, and Gajos}]{arnold_predictive_2020}
Kenneth~C. Arnold, Krysta Chauncey, and Krzysztof~Z. Gajos. 2020.
\newblock \href {https://doi.org/10.1145/3377325.3377523} {Predictive text encourages predictable writing}.
\newblock In \emph{Proceedings of the 25th {International} {Conference} on {Intelligent} {User} {Interfaces}}, {IUI} '20, pages 128--138, New York, NY, USA. Association for Computing Machinery.

\bibitem[{Arnold(2023)}]{arnold_ielts_2023}
Paris Arnold. 2023.
\newblock {IELTS} {Writing} {Band} {Descriptors}.

\bibitem[{Balta et~al.(2025)Balta, Javidan, Walser, Arntfield, and Prager}]{balta_evaluating_2025}
Kaan~Y. Balta, Arshia~P. Javidan, Eric Walser, Robert Arntfield, and Ross Prager. 2025.
\newblock \href {https://doi.org/10.1177/08850666241267871} {Evaluating the {Appropriateness}, {Consistency}, and {Readability} of {ChatGPT} in {Critical} {Care} {Recommendations}}.
\newblock \emph{Journal of Intensive Care Medicine}, 40(2):184--190.
\newblock Publisher: SAGE Publications Inc STM.

\bibitem[{Barbara et~al.(2024)Barbara, Afzaal, and Aldayel}]{barbara2024corpus}
Siu Wing~Yee Barbara, Muhammad Afzaal, and Hessah~Saleh Aldayel. 2024.
\newblock A corpus-based comparison of linguistic markers of stance and genre in the academic writing of novice and advanced engineering learners.
\newblock \emph{Humanities and Social Sciences Communications}, 11(1):1--10.

\bibitem[{Baur(2020)}]{baur-2020}
Dorothea Baur. 2020.
\newblock \href {https://dorotheabaur.medium.com/four-reasons-why-hyping-ai-is-an-ethical-problem-8db47b17bf43} {Four reasons why hyping {AI} is an ethical problem}.
\newblock Accessed: February 14, 2025.

\bibitem[{Beaugrande and Dressler(1981)}]{beaugrande_introduction_1981}
Robert~De Beaugrande and Wolfgang~U. Dressler. 1981.
\newblock \emph{Introduction to {Text} {Linguistics}}.
\newblock Longman.
\newblock Google-Books-ID: TmhiAAAAMAAJ.

\bibitem[{Berland and Mcneill(2012)}]{berland_for_2012}
Leema Berland and Katherine Mcneill. 2012.
\newblock \href {https://doi.org/10.1002/sce.21000} {For whom is argument and explanation a necessary distinction? {A} response to {Osborne} and {Patterson}}.
\newblock \emph{Science Education}, 96:808--813.

\bibitem[{Bojar et~al.(2016)Bojar, Chatterjee, Federmann, Graham, Haddow, Huck, Jimeno~Yepes, Koehn, Logacheva, Monz, Negri, Névéol, Neves, Popel, Post, Rubino, Scarton, Specia, Turchi, Verspoor, and Zampieri}]{bojar_findings_2016}
Ondřej Bojar, Rajen Chatterjee, Christian Federmann, Yvette Graham, Barry Haddow, Matthias Huck, Antonio Jimeno~Yepes, Philipp Koehn, Varvara Logacheva, Christof Monz, Matteo Negri, Aurélie Névéol, Mariana Neves, Martin Popel, Matt Post, Raphael Rubino, Carolina Scarton, Lucia Specia, Marco Turchi, Karin Verspoor, and Marcos Zampieri. 2016.
\newblock \href {https://doi.org/10.18653/v1/W16-2301} {Findings of the 2016 {Conference} on {Machine} {Translation}}.
\newblock In \emph{Proceedings of the {First} {Conference} on {Machine} {Translation}: {Volume} 2, {Shared} {Task} {Papers}}, pages 131--198, Berlin, Germany. Association for Computational Linguistics.

\bibitem[{Braet(1992)}]{braet1992ethos}
Antoine~C Braet. 1992.
\newblock Ethos, pathos and logos in aristotle's rhetoric: A re-examination.
\newblock \emph{Argumentation}, 6:307--320.

\bibitem[{Brassard et~al.(2024)Brassard, Heinzerling, Kudo, Sakaguchi, and Inui}]{brassard_acorn_2024}
Ana Brassard, Benjamin Heinzerling, Keito Kudo, Keisuke Sakaguchi, and Kentaro Inui. 2024.
\newblock \href {https://doi.org/10.48550/arXiv.2405.04818} {{ACORN}: {Aspect}-wise {Commonsense} {Reasoning} {Explanation} {Evaluation}}.
\newblock \emph{arXiv preprint}.
\newblock ArXiv:2405.04818 [cs].

\bibitem[{Brigandt(2016)}]{brigandt_why_2016}
Ingo Brigandt. 2016.
\newblock \href {https://doi.org/10.1007/s11191-016-9826-6} {Why the {Difference} {Between} {Explanation} and {Argument} {Matters} to {Science} {Education}}.
\newblock \emph{Science \& Education}, 25.

\bibitem[{Bromberger(1992)}]{bromberger1992we}
Sylvain Bromberger. 1992.
\newblock \emph{On what we know we don't know: Explanation, theory, linguistics, and how questions shape them}.
\newblock University of Chicago Press.

\bibitem[{Brown(2010)}]{brown_validity_2010}
Gavin Brown. 2010.
\newblock \href {https://doi.org/10.1111/j.1468-2273.2010.00460.x} {The {Validity} of {Examination} {Essays} in {Higher} {Education}: {Issues} and {Responses}}.
\newblock \emph{Higher Education Quarterly}, 64:276--291.

\bibitem[{Brust-Renck et~al.(2021)Brust-Renck, Weldon, and Reyna}]{brust2021judgment}
Priscila~G Brust-Renck, Rebecca~B Weldon, and Valerie~F Reyna. 2021.
\newblock Judgment and decision making.

\bibitem[{Bryant et~al.(2019)Bryant, Felice, Andersen, and Briscoe}]{bryant_bea-2019_2019}
Christopher Bryant, Mariano Felice, Øistein~E. Andersen, and Ted Briscoe. 2019.
\newblock \href {https://doi.org/10.18653/v1/W19-4406} {The {BEA}-2019 {Shared} {Task} on {Grammatical} {Error} {Correction}}.
\newblock In \emph{Proceedings of the {Fourteenth} {Workshop} on {Innovative} {Use} of {NLP} for {Building} {Educational} {Applications}}, pages 52--75, Florence, Italy. Association for Computational Linguistics.

\bibitem[{Bryant et~al.(2023)Bryant, Yuan, Qorib, Cao, Ng, and Briscoe}]{bryant_grammatical_2023}
Christopher Bryant, Zheng Yuan, Muhammad~Reza Qorib, Hannan Cao, Hwee~Tou Ng, and Ted Briscoe. 2023.
\newblock \href {https://doi.org/10.1162/coli_a_00478} {Grammatical {Error} {Correction}: {A} {Survey} of the {State} of the {Art}}.
\newblock \emph{Computational Linguistics}, pages 643--701.
\newblock Place: Cambridge, MA Publisher: MIT Press.

\bibitem[{Burchardt(2013)}]{burchardt-2013-multidimensional}
Aljoscha Burchardt. 2013.
\newblock \href {https://aclanthology.org/2013.tc-1.6} {Multidimensional quality metrics: a flexible system for assessing translation quality}.
\newblock In \emph{Proceedings of Translating and the Computer 35}, London, UK. Aslib.

\bibitem[{Callison-Burch et~al.(2007)Callison-Burch, Fordyce, Koehn, Monz, and Schroeder}]{callison-burch_meta-_2007}
Chris Callison-Burch, Cameron Fordyce, Philipp Koehn, Christof Monz, and Josh Schroeder. 2007.
\newblock \href {https://aclanthology.org/W07-0718/} {({Meta}-) {Evaluation} of {Machine} {Translation}}.
\newblock In \emph{Proceedings of the {Second} {Workshop} on {Statistical} {Machine} {Translation}}, pages 136--158, Prague, Czech Republic. Association for Computational Linguistics.

\bibitem[{Canale(1983)}]{canale_communicative_1983}
Michael Canale. 1983.
\newblock From communicative competence to communicative language pedagogy 1.
\newblock In \emph{Language and {Communication}}. Routledge.
\newblock Num Pages: 26.

\bibitem[{Cao and Zhuge(2022)}]{cao_automatic_2022}
Mengyun Cao and Hai Zhuge. 2022.
\newblock \href {https://doi.org/10.1016/j.eswa.2022.117777} {Automatic evaluation of summary on fidelity, conciseness and coherence for text summarization based on semantic link network}.
\newblock \emph{Expert Systems with Applications}, 206:117777.

\bibitem[{Chin and Brown(2000)}]{chin_learning_2000}
Christine Chin and David~E. Brown. 2000.
\newblock \href {https://doi.org/10.1002/(SICI)1098-2736(200002)37:2<109::AID-TEA3>3.0.CO;2-7} {Learning in {Science}: {A} {Comparison} of {Deep} and {Surface} {Approaches}}.
\newblock \emph{Journal of Research in Science Teaching}, 37(2):109--138.

\bibitem[{Chomsky(1965)}]{chomsky_aspects_1965}
Noam Chomsky. 1965.
\newblock \href {https://www.jstor.org/stable/j.ctt17kk81z} {\emph{Aspects of the {Theory} of {Syntax}}}, 50 edition.
\newblock The MIT Press.

\bibitem[{Clark et~al.(2021)Clark, August, Serrano, Haduong, Gururangan, and Smith}]{clark2021all}
Elizabeth Clark, Tal August, Sofia Serrano, Nikita Haduong, Suchin Gururangan, and Noah~A Smith. 2021.
\newblock All that's' human'is not gold: Evaluating human evaluation of generated text.
\newblock \emph{arXiv preprint arXiv:2107.00061}.

\bibitem[{Cohen(1960)}]{Cohen1960ACO}
Jacob Cohen. 1960.
\newblock \href {https://api.semanticscholar.org/CorpusID:15926286} {A coefficient of agreement for nominal scales}.
\newblock \emph{Educational and Psychological Measurement}, 20:37 -- 46.

\bibitem[{{Cohere for AI}(2024)}]{cohere_c4ai_2024}
{Cohere for AI}. 2024.
\newblock Introducing command r plus on microsoft azure.
\newblock \url{https://cohere.com/blog/command-r-plus-microsoft-azure}.
\newblock Accessed: 2025-02-14.

\bibitem[{Collins(1998)}]{collins_strategies_1998}
James Collins. 1998.
\newblock \href {https://doi.org/10.2307/358940} {Strategies for {Struggling} {Writers}}.
\newblock \emph{College Composition and Communication}, 49:298.

\bibitem[{Connor(1990)}]{connor_linguisticrhetorical_1990}
Ulla Connor. 1990.
\newblock \href {https://doi.org/10.58680/rte199015501} {Linguistic/{Rhetorical} {Measures} for {International} {Persuasive} {Student} {Writing}}.
\newblock \emph{Research in the Teaching of English}, 24(1):67--87.
\newblock Publisher: ncte.org.

\bibitem[{Crossley et~al.(2024)Crossley, Tian, Baffour, Franklin, Kim, Morris, Benner, Picou, and Boser}]{crossley_english_2024}
Scott Crossley, Yu~Tian, Perpetual Baffour, Alex Franklin, Youngmeen Kim, Wesley Morris, Meg Benner, Aigner Picou, and Ulrich Boser. 2024.
\newblock The {English} {Language} {Learner} {Insight}, {Proficiency} and {Skills} {Evaluation} ({ELLIPSE}) {Corpus}.
\newblock \emph{International Journal of Learner Corpus Research}.
\newblock Status: forthcoming.

\bibitem[{Dawson(2017)}]{dawson2017assessment}
Phillip Dawson. 2017.
\newblock Assessment rubrics: towards clearer and more replicable design, research and practice.
\newblock \emph{Assessment \& Evaluation in Higher Education}, 42(3):347--360.

\bibitem[{Devillez(2003)}]{devillez_writing_2003}
Randy Devillez. 2003.
\newblock \emph{Writing: {Step} by {Step}}.
\newblock Kendall Hunt Publishing Company.
\newblock Google-Books-ID: 79oAePQ7Of0C.

\bibitem[{Dewaele(2008)}]{dewaele_appropriateness_2008}
Jean-Marc Dewaele. 2008.
\newblock \href {https://doi.org/10.1515/IRAL.2008.011} {“{Appropriateness}” in foreign language acquisition and use: {Some} theoretical, methodological and ethical considerations}.
\newblock 46(3):245--265.
\newblock Publisher: De Gruyter Mouton Section: International Review of Applied Linguistics in Language Teaching.

\bibitem[{Doshi-Velez and Kim(2017)}]{doshi-velez_towards_2017}
Finale Doshi-Velez and Been Kim. 2017.
\newblock \href {https://doi.org/10.48550/arXiv.1702.08608} {Towards {A} {Rigorous} {Science} of {Interpretable} {Machine} {Learning}}.
\newblock \emph{arXiv preprint}.
\newblock ArXiv:1702.08608.

\bibitem[{Dubey et~al.(2024)Dubey, Jauhri, Pandey, Kadian, Al-Dahle, Letman, Mathur, Schelten, Yang, Fan et~al.}]{dubey_llama3_2024}
Abhimanyu Dubey, Abhinav Jauhri, Abhinav Pandey, Abhishek Kadian, Ahmad Al-Dahle, Aiesha Letman, Akhil Mathur, Alan Schelten, Amy Yang, Angela Fan, et~al. 2024.
\newblock The llama 3 herd of models.
\newblock \emph{arXiv preprint arXiv:2407.21783}.

\bibitem[{Dunn et~al.(2003)Dunn, Morgan, O'Reilly, and Parry}]{dunn_student_2003}
Lee Dunn, Chris Morgan, Meg O'Reilly, and Sharon Parry. 2003.
\newblock \href {https://doi.org/10.4324/9780203416518} {\emph{The {Student} {Assessment} {Handbook}: {New} {Directions} in {Traditional} and {Online} {Assessment}}}.
\newblock Routledge, London.

\bibitem[{Expósito-Ruiz et~al.(2010)Expósito-Ruiz, Pérez-Vicente, and Rivas-Ruiz}]{exposito-ruiz_statistical_2010}
M.~Expósito-Ruiz, S.~Pérez-Vicente, and F.~Rivas-Ruiz. 2010.
\newblock \href {https://doi.org/10.1016/j.aller.2010.06.003} {Statistical inference: {Hypothesis} testing}.
\newblock \emph{Allergologia et Immunopathologia}, 38(5):266--277.
\newblock Publisher: Elsevier.

\bibitem[{Fel et~al.(2022)Fel, Vigouroux, Cad{\`e}ne, and Serre}]{fel2022good}
Thomas Fel, David Vigouroux, R{\'e}mi Cad{\`e}ne, and Thomas Serre. 2022.
\newblock How good is your explanation? algorithmic stability measures to assess the quality of explanations for deep neural networks.
\newblock In \emph{Proceedings of the IEEE/CVF Winter Conference on Applications of Computer Vision}, pages 720--730.

\bibitem[{Feng et~al.(2020)Feng, Xie, Gu, Shao, Zhang, Yang, and Yu}]{feng_modeling_2020}
Yang Feng, Wanying Xie, Shuhao Gu, Chenze Shao, Wen Zhang, Zhengxin Yang, and Dong Yu. 2020.
\newblock \href {https://doi.org/10.1609/aaai.v34i01.5334} {Modeling {Fluency} and {Faithfulness} for {Diverse} {Neural} {Machine} {Translation}}.
\newblock In \emph{Proceedings of the {AAAI} {Conference} on {Artificial} {Intelligence}}, volume~34, pages 59--66.
\newblock ISSN: 2374-3468, 2159-5399 Issue: 01 Journal Abbreviation: AAAI.

\bibitem[{Fetzer(2012)}]{jaszczolt_textual_2012}
Anita Fetzer. 2012.
\newblock \href {https://doi.org/10.1017/CBO9781139022453.024} {Textual coherence as a pragmatic phenomenon}.
\newblock In Kasia~M. Jaszczolt and Keith Allan, editors, \emph{The {Cambridge} {Handbook} of {Pragmatics}}, Cambridge {Handbooks} in {Language} and {Linguistics}, pages 447--468. Cambridge University Press, Cambridge.

\bibitem[{Fetzer(2018)}]{fetzer_appropriateness_2018}
Anita Fetzer. 2018.
\newblock Appropriateness in context.

\bibitem[{Freeman et~al.(2016)Freeman, Richard, Lewis, Development, and Humberside)}]{freeman_planning_2016}
Freeman, Richard, Lewis, Roger (BP Professor of~Learning Development, and University~of Humberside). 2016.
\newblock \href {https://doi.org/10.4324/9781315041858} {\emph{Planning and {Implementing} {Assessment}}}.
\newblock Routledge, London.

\bibitem[{Freitag et~al.(2021)Freitag, Foster, Grangier, Ratnakar, Tan, and Macherey}]{freitag-etal-2021-experts}
Markus Freitag, George Foster, David Grangier, Viresh Ratnakar, Qijun Tan, and Wolfgang Macherey. 2021.
\newblock \href {https://doi.org/10.1162/tacl_a_00437} {Experts, errors, and context: A large-scale study of human evaluation for machine translation}.
\newblock \emph{Transactions of the Association for Computational Linguistics}, 9:1460--1474.

\bibitem[{Friedman(1940)}]{friedman_comparison_1940}
Milton Friedman. 1940.
\newblock \href {https://www.jstor.org/stable/2235971} {A {Comparison} of {Alternative} {Tests} of {Significance} for the {Problem} of m {Rankings}}.
\newblock \emph{The Annals of Mathematical Statistics}, 11(1):86--92.
\newblock Publisher: Institute of Mathematical Statistics.

\bibitem[{Garc{\'\i}a-M{\'e}ndez et~al.(2024)Garc{\'\i}a-M{\'e}ndez, de~Arriba-P{\'e}rez, and Somoza-L{\'o}pez}]{garcia2024review}
Silvia Garc{\'\i}a-M{\'e}ndez, Francisco de~Arriba-P{\'e}rez, and Mar{\'\i}a del~Carmen Somoza-L{\'o}pez. 2024.
\newblock A review on the use of large language models as virtual tutors.
\newblock \emph{Science \& Education}, pages 1--16.

\bibitem[{Gilpin et~al.(2018)Gilpin, Bau, Yuan, Bajwa, Specter, and Kagal}]{gilpin2018explaining}
Leilani~H Gilpin, David Bau, Ben~Z Yuan, Ayesha Bajwa, Michael Specter, and Lalana Kagal. 2018.
\newblock Explaining explanations: An overview of interpretability of machine learning.
\newblock In \emph{2018 IEEE 5th International Conference on data science and advanced analytics (DSAA)}, pages 80--89. IEEE.

\bibitem[{Graham et~al.(2013)Graham, Baldwin, Moffat, and Zobel}]{graham_continuous_2013}
Yvette Graham, Timothy Baldwin, Alistair Moffat, and Justin Zobel. 2013.
\newblock \href {https://aclanthology.org/W13-2305/} {Continuous {Measurement} {Scales} in {Human} {Evaluation} of {Machine} {Translation}}.
\newblock In \emph{Proceedings of the 7th {Linguistic} {Annotation} {Workshop} and {Interoperability} with {Discourse}}, pages 33--41, Sofia, Bulgaria. Association for Computational Linguistics.

\bibitem[{Granger et~al.(2009)Granger, Dagneaux, Meunier, and Paquot}]{granger_international_2009}
Sylviane Granger, Estelle Dagneaux, Fanny Meunier, and Magali Paquot. 2009.
\newblock \emph{International {Corpus} of {Learner} {English}. {Version} 2. {Handbook} and {CD}-{ROM}}.

\bibitem[{Halliday and Hasan(2014)}]{halliday_cohesion_2014}
M.~A.~K. Halliday and Ruqaiya Hasan. 2014.
\newblock \href {https://doi.org/10.4324/9781315836010} {\emph{Cohesion in {English}}}.
\newblock Routledge, London.

\bibitem[{Hart(2006)}]{hart2006nasa}
Sandra~G Hart. 2006.
\newblock Nasa-task load index (nasa-tlx); 20 years later.
\newblock In \emph{Proceedings of the human factors and ergonomics society annual meeting}, volume~50, pages 904--908. Sage publications Sage CA: Los Angeles, CA.

\bibitem[{Hart(1988)}]{hart1988development}
SG~Hart. 1988.
\newblock Development of nasa-tlx (task load index): Results of empirical and theoretical research.
\newblock \emph{Human mental workload/Elsevier}.

\bibitem[{Higgins et~al.(2004)Higgins, Burstein, Marcu, and Gentile}]{higgins_evaluating_2004}
Derrick Higgins, Jill Burstein, Daniel Marcu, and Claudia Gentile. 2004.
\newblock \href {https://aclanthology.org/N04-1024} {Evaluating {Multiple} {Aspects} of {Coherence} in {Student} {Essays}}.
\newblock In \emph{Proceedings of the {Human} {Language} {Technology} {Conference} of the {North} {American} {Chapter} of the {Association} for {Computational} {Linguistics}: {HLT}-{NAACL} 2004}, pages 185--192, Boston, Massachusetts, USA. Association for Computational Linguistics.

\bibitem[{Howcroft et~al.(2020)Howcroft, Belz, Clinciu, Gkatzia, Hasan, Mahamood, Mille, van Miltenburg, Santhanam, and Rieser}]{howcroft-etal-2020-twenty}
David~M. Howcroft, Anya Belz, Miruna-Adriana Clinciu, Dimitra Gkatzia, Sadid~A. Hasan, Saad Mahamood, Simon Mille, Emiel van Miltenburg, Sashank Santhanam, and Verena Rieser. 2020.
\newblock \href {https://doi.org/10.18653/v1/2020.inlg-1.23} {Twenty years of confusion in human evaluation: {NLG} needs evaluation sheets and standardised definitions}.
\newblock In \emph{Proceedings of the 13th International Conference on Natural Language Generation}, pages 169--182, Dublin, Ireland. Association for Computational Linguistics.

\bibitem[{Hu et~al.(2024)Hu, Gao, Hu, Zhang, Chen, Xu, and Wan}]{hu_are_2024}
Xinyu Hu, Mingqi Gao, Sen Hu, Yang Zhang, Yicheng Chen, Teng Xu, and Xiaojun Wan. 2024.
\newblock \href {https://doi.org/10.48550/arXiv.2402.12055} {Are {LLM}-based {Evaluators} {Confusing} {NLG} {Quality} {Criteria}?}
\newblock \emph{arXiv preprint}.
\newblock ArXiv:2402.12055 [cs].

\bibitem[{Huang and Chang(2023)}]{huang-chang-2023-towards}
Jie Huang and Kevin Chen-Chuan Chang. 2023.
\newblock \href {https://doi.org/10.18653/v1/2023.findings-acl.67} {Towards reasoning in large language models: A survey}.
\newblock In \emph{Findings of the Association for Computational Linguistics: ACL 2023}, pages 1049--1065, Toronto, Canada. Association for Computational Linguistics.

\bibitem[{Huang et~al.(2025)Huang, Yu, Ma, Zhong, Feng, Wang, Chen, Peng, Feng, Qin, and Liu}]{huang_survey_2025}
Lei Huang, Weijiang Yu, Weitao Ma, Weihong Zhong, Zhangyin Feng, Haotian Wang, Qianglong Chen, Weihua Peng, Xiaocheng Feng, Bing Qin, and Ting Liu. 2025.
\newblock \href {https://doi.org/10.1145/3703155} {A {Survey} on {Hallucination} in {Large} {Language} {Models}: {Principles}, {Taxonomy}, {Challenges}, and {Open} {Questions}}.
\newblock \emph{ACM Trans. Inf. Syst.}, 43(2):42:1--42:55.

\bibitem[{Huba and Freed(2000)}]{huba_learner-centered_2000}
Mary Huba and Jann Freed. 2000.
\newblock Learner-{Centered} {Assessment} on {College} {Campuses}: {Sifting} the {Focus} from {Teaching} to {Learning}.
\newblock \emph{Community College Journal of Research and Practice}, 24.

\bibitem[{Hymes(1972)}]{hymes_communicative_1972}
Dell Hymes. 1972.
\newblock On {Communicative} {Competence}.
\newblock In \emph{Sociolinguistics}, pages 269--293. Harmondsworth: Penguin.

\bibitem[{Jacovi and Goldberg(2021)}]{jacovi_aligning_2021}
Alon Jacovi and Yoav Goldberg. 2021.
\newblock \href {https://doi.org/10.1162/tacl_a_00367} {Aligning {Faithful} {Interpretations} with their {Social} {Attribution}}.
\newblock \emph{Transactions of the Association for Computational Linguistics}, 9:294--310.
\newblock Place: Cambridge, MA Publisher: MIT Press.

\bibitem[{Javidan et~al.(2024)Javidan, Feridooni, Gordon, and Crawford}]{javidan_evaluating_2024}
Arshia~P. Javidan, Tiam Feridooni, Lauren Gordon, and Sean~A. Crawford. 2024.
\newblock \href {https://doi.org/10.1016/j.jvsvi.2023.100049} {Evaluating the progression of artificial intelligence and large language models in medicine through comparative analysis of {ChatGPT}-3.5 and {ChatGPT}-4 in generating vascular surgery recommendations}.
\newblock \emph{JVS-Vascular Insights}, 2:100049.

\bibitem[{Jiang et~al.(2024)Jiang, Sablayrolles, Roux, Mensch, Savary, Bamford, Chaplot, Casas, Hanna, Bressand et~al.}]{jiang_mixtral_2024}
Albert~Q Jiang, Alexandre Sablayrolles, Antoine Roux, Arthur Mensch, Blanche Savary, Chris Bamford, Devendra~Singh Chaplot, Diego de~las Casas, Emma~Bou Hanna, Florian Bressand, et~al. 2024.
\newblock Mixtral of experts.
\newblock \emph{arXiv preprint arXiv:2401.04088}.

\bibitem[{Jin et~al.(2022)Jin, Lalwani, Vaidhya, Shen, Ding, Lyu, Sachan, Mihalcea, and Schoelkopf}]{jin_logical_2022}
Zhijing Jin, Abhinav Lalwani, Tejas Vaidhya, Xiaoyu Shen, Yiwen Ding, Zhiheng Lyu, Mrinmaya Sachan, Rada Mihalcea, and Bernhard Schoelkopf. 2022.
\newblock \href {https://doi.org/10.18653/v1/2022.findings-emnlp.532} {Logical {Fallacy} {Detection}}.
\newblock In \emph{Findings of the {Association} for {Computational} {Linguistics}: {EMNLP} 2022}, pages 7180--7198, Abu Dhabi, United Arab Emirates. Association for Computational Linguistics.

\bibitem[{Kabir et~al.(2024)Kabir, Udo-Imeh, Kou, and Zhang}]{kabir_is_2024}
Samia Kabir, David~N. Udo-Imeh, Bonan Kou, and Tianyi Zhang. 2024.
\newblock \href {https://doi.org/10.1145/3613904.3642596} {Is {Stack} {Overflow} {Obsolete}? {An} {Empirical} {Study} of the {Characteristics} of {ChatGPT} {Answers} to {Stack} {Overflow} {Questions}}.
\newblock In \emph{Proceedings of the 2024 {CHI} {Conference} on {Human} {Factors} in {Computing} {Systems}}, {CHI} '24, pages 1--17, New York, NY, USA. Association for Computing Machinery.

\bibitem[{Karpinska and Iyyer(2023)}]{karpinska-iyyer-2023-large}
Marzena Karpinska and Mohit Iyyer. 2023.
\newblock \href {https://doi.org/10.18653/v1/2023.wmt-1.41} {Large language models effectively leverage document-level context for literary translation, but critical errors persist}.
\newblock In \emph{Proceedings of the Eighth Conference on Machine Translation}, pages 419--451, Singapore. Association for Computational Linguistics.

\bibitem[{Ke and Ng(2019)}]{ke_automated_2019}
Zixuan Ke and Vincent Ng. 2019.
\newblock \href {https://doi.org/10.24963/ijcai.2019/879} {Automated {Essay} {Scoring}: {A} {Survey} of the {State} of the {Art}}.
\newblock pages 6300--6308.

\bibitem[{Keil(2006)}]{keil2006explanation}
Frank~C Keil. 2006.
\newblock Explanation and understanding.
\newblock \emph{Annu. Rev. Psychol.}, 57(1):227--254.

\bibitem[{Kim et~al.(2024)Kim, Lee, Kim, Park, and Kim}]{kim2024understanding}
Yoonsu Kim, Jueon Lee, Seoyoung Kim, Jaehyuk Park, and Juho Kim. 2024.
\newblock Understanding users’ dissatisfaction with chatgpt responses: Types, resolving tactics, and the effect of knowledge level.
\newblock In \emph{Proceedings of the 29th International Conference on Intelligent User Interfaces}, pages 385--404.

\bibitem[{Krippendorff(2011)}]{Krippendorff2011ComputingKA}
Klaus Krippendorff. 2011.
\newblock \href {https://api.semanticscholar.org/CorpusID:59901023} {Computing krippendorff's alpha-reliability}.

\bibitem[{Kristoffersen(2019)}]{kristoffersen_where_2019}
Cherise Kristoffersen. 2019.
\newblock \href {https://doi.org/10.17011/apples/urn.201907163639} {Where do my words come from? {Towards} methods for analyzing word choice in primary level writing}.
\newblock \emph{Apples - Journal of Applied Language Studies}, 13(3):59--75.
\newblock Number: 3.

\bibitem[{Kyle et~al.(2018)Kyle, Crossley, and Berger}]{kyle_tool_2018}
Kristopher Kyle, Scott Crossley, and Cynthia Berger. 2018.
\newblock \href {https://doi.org/10.3758/s13428-017-0924-4} {The tool for the automatic analysis of lexical sophistication ({TAALES}): version 2.0}.
\newblock \emph{Behavior Research Methods}, 50(3):1030--1046.

\bibitem[{Kyle and Crossley(2015)}]{kyle_automatically_2015}
Kristopher Kyle and Scott~A. Crossley. 2015.
\newblock \href {https://doi.org/10.1002/tesq.194} {Automatically {Assessing} {Lexical} {Sophistication}: {Indices}, {Tools}, {Findings}, and {Application}}.
\newblock \emph{TESOL Quarterly}, 49(4):757--786.

\bibitem[{Lai et~al.(2017)Lai, Xie, Liu, Yang, and Hovy}]{lai-etal-2017-race}
Guokun Lai, Qizhe Xie, Hanxiao Liu, Yiming Yang, and Eduard Hovy. 2017.
\newblock \href {https://doi.org/10.18653/v1/D17-1082} {{RACE}: Large-scale {R}e{A}ding comprehension dataset from examinations}.
\newblock In \emph{Proceedings of the 2017 Conference on Empirical Methods in Natural Language Processing}, pages 785--794, Copenhagen, Denmark. Association for Computational Linguistics.

\bibitem[{Li and Ng(2024)}]{li_icle_2024}
Shengjie Li and Vincent Ng. 2024.
\newblock \href {https://aclanthology.org/2024.naacl-long.468} {{ICLE}++: {Modeling} {Fine}-{Grained} {Traits} for {Holistic} {Essay} {Scoring}}.
\newblock In \emph{Proceedings of the 2024 {Conference} of the {North} {American} {Chapter} of the {Association} for {Computational} {Linguistics}: {Human} {Language} {Technologies} ({Volume} 1: {Long} {Papers})}, pages 8465--8486, Mexico City, Mexico. Association for Computational Linguistics.

\bibitem[{Li et~al.(2025)Li, Suzuki, Morishita, Abe, and Inui}]{li-etal-2025-mqm}
Yunmeng Li, Jun Suzuki, Makoto Morishita, Kaori Abe, and Kentaro Inui. 2025.
\newblock \href {https://aclanthology.org/2025.coling-main.221/} {{MQM}-chat: Multidimensional quality metrics for chat translation}.
\newblock In \emph{Proceedings of the 31st International Conference on Computational Linguistics}, pages 3283--3299, Abu Dhabi, UAE. Association for Computational Linguistics.

\bibitem[{Lombrozo(2006)}]{lombrozo2006structure}
Tania Lombrozo. 2006.
\newblock The structure and function of explanations.
\newblock \emph{Trends in cognitive sciences}, 10(10):464--470.

\bibitem[{Long(2007)}]{long_college_2007}
Elizabeth~Cloninger Long. 2007.
\newblock \href {http://archive.org/details/collegewritingre0000long} {\emph{College writing resources with readings}}.
\newblock New York : Pearson/Longman.

\bibitem[{Lunsford et~al.(2008)Lunsford, Wilson, and Eberly}]{lunsford2008sage}
Andrea~A Lunsford, Kirt~H Wilson, and Rosa~A Eberly. 2008.
\newblock \emph{The SAGE handbook of rhetorical studies}.
\newblock Sage Publications.

\bibitem[{Mariana(2014)}]{mariana2014multidimensional}
Valerie~R Mariana. 2014.
\newblock \emph{The Multidimensional Quality Metric (MQM) framework: A new framework for translation quality assessment}.
\newblock Brigham Young University.

\bibitem[{Martindale et~al.(2019)Martindale, Carpuat, Duh, and McNamee}]{martindale_identifying_2019}
Marianna Martindale, Marine Carpuat, Kevin Duh, and Paul McNamee. 2019.
\newblock \href {https://aclanthology.org/W19-6623/} {Identifying {Fluently} {Inadequate} {Output} in {Neural} and {Statistical} {Machine} {Translation}}.
\newblock In \emph{Proceedings of {Machine} {Translation} {Summit} {XVII}: {Research} {Track}}, pages 233--243, Dublin, Ireland. European Association for Machine Translation.

\bibitem[{Mathias and Bhattacharyya(2018)}]{mathias_asap_2018}
Sandeep Mathias and Pushpak Bhattacharyya. 2018.
\newblock \href {https://aclanthology.org/L18-1187} {{ASAP}++: {Enriching} the {ASAP} {Automated} {Essay} {Grading} {Dataset} with {Essay} {Attribute} {Scores}}.
\newblock In \emph{Proceedings of the {Eleventh} {International} {Conference} on {Language} {Resources} and {Evaluation} ({LREC} 2018)}, Miyazaki, Japan. European Language Resources Association (ELRA).

\bibitem[{McCarthy and Jarvis(2007)}]{mccarthy_vocd_2007}
Philip~M. McCarthy and Scott Jarvis. 2007.
\newblock \href {https://doi.org/10.1177/0265532207080767} {vocd: {A} theoretical and empirical evaluation}.
\newblock \emph{Language Testing}, 24(4):459--488.

\bibitem[{McNamara and {Com}(2010)}]{mcnamara_cohesion_2010}
Danielle McNamara and {Com}. 2010.
\newblock \emph{Cohesion, coherence, and expert evaluations of writing proficiency}.
\newblock Journal Abbreviation: Proceedings of the 32nd Annual Conference of the Cognitive Science Society Publication Title: Proceedings of the 32nd Annual Conference of the Cognitive Science Society.

\bibitem[{McNamara et~al.(2014)McNamara, Graesser, McCarthy, and Cai}]{mcnamara_automated_2014}
Danielle~S. McNamara, Arthur~C. Graesser, Philip~M. McCarthy, and Zhiqiang Cai. 2014.
\newblock \emph{Automated {Evaluation} of {Text} and {Discourse} with {Coh}-{Metrix}}.
\newblock Cambridge University Press.
\newblock Google-Books-ID: xSPeAgAAQBAJ.

\bibitem[{McNeill and Krajcik(2007)}]{mcneill_middle_2007}
Katharine~L. McNeill and Joseph Krajcik. 2007.
\newblock Middle school students' use of appropriate and inappropriate evidence in writing scientific explanations.
\newblock In \emph{Thinking with data}, Carnegie {Mellon} symposia on cognition, pages 233--265. Lawrence Erlbaum Associates Publishers, Mahwah, NJ, US.

\bibitem[{Mcneill et~al.(2006)Mcneill, Lizotte, Krajcik, and Marx}]{mcneill_supporting_2006}
Katherine Mcneill, David Lizotte, Joseph Krajcik, and Ronald Marx. 2006.
\newblock \href {https://doi.org/10.1207/s15327809jls1502_1} {Supporting {Students}' {Construction} of {Scientific} {Explanations} by {Fading} {Scaffolds} in {Instructional} {Materials}}.
\newblock \emph{Journal of the Learning Sciences}, 15:153--191.

\bibitem[{McNeill and Krajcik(2008)}]{mcneill_scientific_2008}
Katherine~L. McNeill and Joseph Krajcik. 2008.
\newblock \href {https://doi.org/10.1002/tea.20201} {Scientific explanations: {Characterizing} and evaluating the effects of teachers' instructional practices on student learning}.
\newblock \emph{Journal of Research in Science Teaching}, 45(1):53--78.

\bibitem[{Miller(2019{\natexlab{a}})}]{miller2019explanation}
Tim Miller. 2019{\natexlab{a}}.
\newblock Explanation in artificial intelligence: Insights from the social sciences.
\newblock \emph{Artificial intelligence}, 267:1--38.

\bibitem[{Miller(2019{\natexlab{b}})}]{miller_explanation_2019}
Tim Miller. 2019{\natexlab{b}}.
\newblock \href {https://doi.org/10.1016/j.artint.2018.07.007} {Explanation in artificial intelligence: {Insights} from the social sciences}.
\newblock \emph{Artificial Intelligence}, 267:1--38.

\bibitem[{Miltsakaki(2004)}]{miltsakaki_evaluation_2004}
Eleni Miltsakaki. 2004.
\newblock \href {https://doi.org/10.1017/S1351324903003206} {Evaluation of text coherence for electronic essay scoring systems}.
\newblock \emph{Natural Language Engineering}, 10:25--55.

\bibitem[{North and Piccardo(2020)}]{north_common_2020}
Brian North and Enrica Piccardo. 2020.
\newblock \emph{Common {European} {Framework} of {Reference} for {Languages}: {Learning}, {Teaching}, {Assessment} {Common} {European} {Framework} of {Reference} for {Languages}: {Learning}, {Teaching}, {Assessment}. {Companion} volume {Language} {Policy} {Programme} {Education} {Policy} {Division} {Education} {Department} {Council} of {Europe}}.

\bibitem[{Ohlsson(2002)}]{ohlsson_generating_2002}
Stellan Ohlsson. 2002.
\newblock Generating and understanding qualitative explanations.
\newblock In \emph{The psychology of science text comprehension}, pages 91--128. Lawrence Erlbaum Associates Publishers, Mahwah, NJ, US.

\bibitem[{OpenAI(2024)}]{openai_gpt4o_2024}
OpenAI. 2024.
\newblock \href {https://openai.com/index/hello-gpt-4o/} {Hello gpt-4o}.
\newblock Accessed: February 2025.

\bibitem[{Panickssery et~al.(2024)Panickssery, Bowman, and Feng}]{panickssery2024llm}
Arjun Panickssery, Samuel Bowman, and Shi Feng. 2024.
\newblock Llm evaluators recognize and favor their own generations.
\newblock \emph{Advances in Neural Information Processing Systems}, 37:68772--68802.

\bibitem[{Park and Pad{\'o}(2024)}]{park-pado-2024-multi}
Dojun Park and Sebastian Pad{\'o}. 2024.
\newblock \href {https://aclanthology.org/2024.lrec-main.1024/} {Multi-dimensional machine translation evaluation: Model evaluation and resource for {K}orean}.
\newblock In \emph{Proceedings of the 2024 Joint International Conference on Computational Linguistics, Language Resources and Evaluation (LREC-COLING 2024)}, pages 11723--11744, Torino, Italia. ELRA and ICCL.

\bibitem[{Park et~al.(2018)Park, Hendricks, Akata, Rohrbach, Schiele, Darrell, and Rohrbach}]{park2018multimodal}
Dong~Huk Park, Lisa~Anne Hendricks, Zeynep Akata, Anna Rohrbach, Bernt Schiele, Trevor Darrell, and Marcus Rohrbach. 2018.
\newblock Multimodal explanations: Justifying decisions and pointing to the evidence.
\newblock In \emph{Proceedings of the IEEE conference on computer vision and pattern recognition}, pages 8779--8788.

\bibitem[{Persing and Ng(2013)}]{persing_modeling_2013}
Isaac Persing and Vincent Ng. 2013.
\newblock \href {https://aclanthology.org/P13-1026} {Modeling {Thesis} {Clarity} in {Student} {Essays}}.
\newblock In \emph{Proceedings of the 51st {Annual} {Meeting} of the {Association} for {Computational} {Linguistics} ({Volume} 1: {Long} {Papers})}, pages 260--269, Sofia, Bulgaria. Association for Computational Linguistics.

\bibitem[{Persing and Ng(2015)}]{persing_modeling_2015}
Isaac Persing and Vincent Ng. 2015.
\newblock \href {https://doi.org/10.3115/v1/P15-1053} {Modeling {Argument} {Strength} in {Student} {Essays}}.
\newblock In \emph{Proceedings of the 53rd {Annual} {Meeting} of the {Association} for {Computational} {Linguistics} and the 7th {International} {Joint} {Conference} on {Natural} {Language} {Processing} ({Volume} 1: {Long} {Papers})}, pages 543--552, Beijing, China. Association for Computational Linguistics.

\bibitem[{Peyrard(2019)}]{peyrard_simple_2019}
Maxime Peyrard. 2019.
\newblock \href {https://doi.org/10.18653/v1/P19-1101} {A {Simple} {Theoretical} {Model} of {Importance} for {Summarization}}.
\newblock In \emph{Proceedings of the 57th {Annual} {Meeting} of the {Association} for {Computational} {Linguistics}}, pages 1059--1073, Florence, Italy. Association for Computational Linguistics.

\bibitem[{Pride(1972)}]{pride_sociolinguistics_1972}
J.~B. Pride. 1972.
\newblock \href {http://archive.org/details/sociolinguistics0000unse_n0z7} {\emph{Sociolinguistics : selected readings}}.
\newblock Harmondsworth, Penguin.

\bibitem[{Quinn and Zhai(2016)}]{quinn_cost-benefit_2016}
Philip Quinn and Shumin Zhai. 2016.
\newblock \href {https://doi.org/10.1145/2858036.2858305} {A {Cost}-{Benefit} {Study} of {Text} {Entry} {Suggestion} {Interaction}}.
\newblock In \emph{Proceedings of the 2016 {CHI} {Conference} on {Human} {Factors} in {Computing} {Systems}}, {CHI} '16, pages 83--88, New York, NY, USA. Association for Computing Machinery.

\bibitem[{Sallam(2023)}]{sallam2023chatgpt}
Malik Sallam. 2023.
\newblock Chatgpt utility in healthcare education, research, and practice: systematic review on the promising perspectives and valid concerns.
\newblock In \emph{Healthcare}, volume~11, page 887. MDPI.

\bibitem[{Sandoval(2003)}]{sandoval_conceptual_2003}
William~A. Sandoval. 2003.
\newblock \href {https://www.jstor.org/stable/1466633} {Conceptual and {Epistemic} {Aspects} of {Students}' {Scientific} {Explanations}}.
\newblock \emph{The Journal of the Learning Sciences}, 12(1):5--51.
\newblock Publisher: Taylor \& Francis, Ltd.

\bibitem[{Saxena et~al.(2024)Saxena, Chopra, and Tripathi}]{saxena2024evaluating}
Yash Saxena, Sarthak Chopra, and Arunendra~Mani Tripathi. 2024.
\newblock Evaluating consistency and reasoning capabilities of large language models.
\newblock \emph{arXiv preprint arXiv:2404.16478}.

\bibitem[{Song et~al.(2014)Song, Heilman, Beigman~Klebanov, and Deane}]{song_applying_2014}
Yi~Song, Michael Heilman, Beata Beigman~Klebanov, and Paul Deane. 2014.
\newblock \href {https://doi.org/10.3115/v1/W14-2110} {Applying {Argumentation} {Schemes} for {Essay} {Scoring}}.
\newblock In \emph{Proceedings of the {First} {Workshop} on {Argumentation} {Mining}}, pages 69--78, Baltimore, Maryland. Association for Computational Linguistics.

\bibitem[{Sottana et~al.(2023)Sottana, Liang, Zou, and Yuan}]{sottana-etal-2023-evaluation}
Andrea Sottana, Bin Liang, Kai Zou, and Zheng Yuan. 2023.
\newblock \href {https://doi.org/10.18653/v1/2023.emnlp-main.543} {Evaluation metrics in the era of {GPT}-4: Reliably evaluating large language models on sequence to sequence tasks}.
\newblock In \emph{Proceedings of the 2023 Conference on Empirical Methods in Natural Language Processing}, pages 8776--8788, Singapore. Association for Computational Linguistics.

\bibitem[{Spitale et~al.(2024)Spitale, Axelsson, and Gunes}]{spitale_appropriateness_2024}
Micol Spitale, Minja Axelsson, and Hatice Gunes. 2024.
\newblock \href {https://doi.org/10.48550/arXiv.2401.14935} {Appropriateness of {LLM}-equipped {Robotic} {Well}-being {Coach} {Language} in the {Workplace}: {A} {Qualitative} {Evaluation}}.
\newblock \emph{arXiv preprint}.
\newblock ArXiv:2401.14935 [cs].

\bibitem[{Stab and Gurevych(2014)}]{stab_annotating_2014}
Christian Stab and Iryna Gurevych. 2014.
\newblock \href {https://aclanthology.org/C14-1142} {Annotating {Argument} {Components} and {Relations} in {Persuasive} {Essays}}.
\newblock In \emph{Proceedings of {COLING} 2014, the 25th {International} {Conference} on {Computational} {Linguistics}: {Technical} {Papers}}, pages 1501--1510, Dublin, Ireland. Dublin City University and Association for Computational Linguistics.

\bibitem[{Stede(2002)}]{stede_lexical_2002}
Manfred Stede. 2002.
\newblock \href {https://doi.org/10.3115/976744.976799} {Lexical {Choice} {Criteria} in {Language} {Generation}}.

\bibitem[{Stevens and Levi(2004)}]{stevens_introduction_2004}
Dannelle~D. Stevens and Antonia~J. Levi. 2004.
\newblock \emph{Introduction to {Rubrics}: {An} {Assessment} {Tool} to {Save} {Grading} {Time}, {Convey} {Effective} {Feedback} and {Promote} {Student} {Learning}}.
\newblock Stylus Publishing, LLC.
\newblock Publication Title: Stylus Publishing, LLC ERIC Number: ED515062.

\bibitem[{Strauss and Feiz(2013)}]{strauss_discourse_2013}
Susan Strauss and Parastou Feiz. 2013.
\newblock \href {https://doi.org/10.4324/9780203121559} {Discourse analysis: {Putting} our worlds into words}.
\newblock \emph{Discourse Analysis: Putting our Worlds into Words}, pages 1--411.

\bibitem[{Team et~al.(2024)Team, Riviere, Pathak, Sessa, Hardin, Bhupatiraju, Hussenot, Mesnard, Shahriari, Ram{\'e} et~al.}]{team_gemma2_2024}
Gemma Team, Morgane Riviere, Shreya Pathak, Pier~Giuseppe Sessa, Cassidy Hardin, Surya Bhupatiraju, L{\'e}onard Hussenot, Thomas Mesnard, Bobak Shahriari, Alexandre Ram{\'e}, et~al. 2024.
\newblock Gemma 2: Improving open language models at a practical size.
\newblock \emph{arXiv preprint arXiv:2408.00118}.

\bibitem[{Toral and Sánchez-Cartagena(2017)}]{toral_multifaceted_2017}
Antonio Toral and Víctor~M. Sánchez-Cartagena. 2017.
\newblock \href {https://aclanthology.org/E17-1100/} {A {Multifaceted} {Evaluation} of {Neural} versus {Phrase}-{Based} {Machine} {Translation} for 9 {Language} {Directions}}.
\newblock In \emph{Proceedings of the 15th {Conference} of the {European} {Chapter} of the {Association} for {Computational} {Linguistics}: {Volume} 1, {Long} {Papers}}, pages 1063--1073, Valencia, Spain. Association for Computational Linguistics.

\bibitem[{Toulmin(1958)}]{toulmin-1958-arguments}
Stephen Toulmin. 1958.
\newblock \emph{The Uses of Arguments}, 1 edition.
\newblock Cambridge University Press.

\bibitem[{Virtanen et~al.(2020)Virtanen, Gommers, Oliphant, Haberland, Reddy, Cournapeau, Burovski, Peterson, Weckesser, Bright, van~der Walt, Brett, Wilson, Millman, Mayorov, Nelson, Jones, Kern, Larson, Carey, Polat, Feng, Moore, VanderPlas, Laxalde, Perktold, Cimrman, Henriksen, Quintero, Harris, Archibald, Ribeiro, Pedregosa, and van Mulbregt}]{virtanen_scipy_2020}
Pauli Virtanen, Ralf Gommers, Travis~E. Oliphant, Matt Haberland, Tyler Reddy, David Cournapeau, Evgeni Burovski, Pearu Peterson, Warren Weckesser, Jonathan Bright, Stéfan~J. van~der Walt, Matthew Brett, Joshua Wilson, K.~Jarrod Millman, Nikolay Mayorov, Andrew R.~J. Nelson, Eric Jones, Robert Kern, Eric Larson, C.~J. Carey, İlhan Polat, Yu~Feng, Eric~W. Moore, Jake VanderPlas, Denis Laxalde, Josef Perktold, Robert Cimrman, Ian Henriksen, E.~A. Quintero, Charles~R. Harris, Anne~M. Archibald, Antônio~H. Ribeiro, Fabian Pedregosa, and Paul van Mulbregt. 2020.
\newblock \href {https://doi.org/10.1038/s41592-019-0686-2} {{SciPy} 1.0: fundamental algorithms for scientific computing in {Python}}.
\newblock \emph{Nature Methods}, 17(3):261--272.
\newblock Publisher: Nature Publishing Group.

\bibitem[{Walvoord and Anderson(1998)}]{walvoord_effective_1998}
Barbara~E. Walvoord and Virginia~Johnson Anderson. 1998.
\newblock \emph{Effective {Grading}: {A} {Tool} for {Learning} and {Assessment}}.
\newblock Jossey-Bass Publishers, 350 Sansome St.
\newblock ERIC Number: ED416810.

\bibitem[{Wan et~al.(2007)Wan, Yang, and Xiao}]{wan_manifold-ranking_2007}
Xiaojun Wan, Jianwu Yang, and Jianguo Xiao. 2007.
\newblock Manifold-ranking based topic-focused multi-document summarization.
\newblock In \emph{Proceedings of the 20th international joint conference on {Artifical} intelligence}, {IJCAI}'07, pages 2903--2908, San Francisco, CA, USA. Morgan Kaufmann Publishers Inc.

\bibitem[{Wei et~al.(2018)Wei, Pham, Dillon, and O'Connor}]{wei_evaluating_2018}
Johnny Tian-Zheng Wei, Khiem Pham, Brian Dillon, and Brendan O'Connor. 2018.
\newblock \href {https://doi.org/10.48550/arXiv.1809.02035} {Evaluating {Syntactic} {Properties} of {Seq2seq} {Output} with a {Broad} {Coverage} {HPSG}: {A} {Case} {Study} on {Machine} {Translation}}.
\newblock \emph{arXiv preprint}.
\newblock ArXiv:1809.02035 [cs].

\bibitem[{Wiegreffe and Marasovi{\'c}(2021)}]{Wiegreffe2021TeachMT}
Sarah Wiegreffe and Ana Marasovi{\'c}. 2021.
\newblock \href {https://api.semanticscholar.org/CorpusID:232035689} {Teach me to explain: A review of datasets for explainable natural language processing}.
\newblock In \emph{NeurIPS Datasets and Benchmarks}.

\bibitem[{Wu and Hu(2018)}]{wu_learning_2018}
Yuxiang Wu and Baotian Hu. 2018.
\newblock \href {https://doi.org/10.1609/aaai.v32i1.11987} {Learning to {Extract} {Coherent} {Summary} via {Deep} {Reinforcement} {Learning}}.
\newblock \emph{Proceedings of the AAAI Conference on Artificial Intelligence}, 32(1).
\newblock Number: 1.

\bibitem[{Yannakoudakis et~al.(2018)Yannakoudakis, Andersen, Geranpayeh, Briscoe, and Nicholls}]{yannakoudakis_developing_2018}
Helen Yannakoudakis, Øistein Andersen, Ardeshir Geranpayeh, Ted Briscoe, and Diane Nicholls. 2018.
\newblock \href {https://doi.org/10.1080/08957347.2018.1464447} {Developing an automated writing placement system for {ESL} learners}.
\newblock \emph{Applied Measurement in Education}, 31.

\bibitem[{Yao et~al.(2017)Yao, Wan, and Xiao}]{yao_recent_2017}
Jin-ge Yao, Xiaojun Wan, and Jianguo Xiao. 2017.
\newblock \href {https://doi.org/10.1007/s10115-017-1042-4} {Recent advances in document summarization}.
\newblock \emph{Knowledge and Information Systems}, 53(2):297--336.

\bibitem[{Zangori et~al.(2013)Zangori, Forbes, and Biggers}]{zangori_fostering_2013}
Laura Zangori, Cory~T. Forbes, and Mandy Biggers. 2013.
\newblock \href {https://doi.org/10.1002/tea.21104} {Fostering student sense making in elementary science learning environments: {Elementary} teachers' use of science curriculum materials to promote explanation construction}.
\newblock \emph{Journal of Research in Science Teaching}, 50(8):989--1017.

\bibitem[{Zellers et~al.(2019)Zellers, Holtzman, Bisk, Farhadi, and Choi}]{zellers-etal-2019-hellaswag}
Rowan Zellers, Ari Holtzman, Yonatan Bisk, Ali Farhadi, and Yejin Choi. 2019.
\newblock \href {https://doi.org/10.18653/v1/P19-1472} {{H}ella{S}wag: Can a machine really finish your sentence?}
\newblock In \emph{Proceedings of the 57th Annual Meeting of the Association for Computational Linguistics}, pages 4791--4800, Florence, Italy. Association for Computational Linguistics.

\bibitem[{Zhang(2006)}]{zhang_text-based_2006}
Jiegen Zhang. 2006.
\newblock \href {https://doi.org/10.13140/2.1.3336.5124} {\emph{A {Text}-based {Approach} to {Cohesion} and {Coherence}}}.
\newblock Ph.D. thesis.

\bibitem[{Zhang et~al.(2023{\natexlab{a}})Zhang, Press, Merrill, Liu, and Smith}]{zhang2023language}
Muru Zhang, Ofir Press, William Merrill, Alisa Liu, and Noah~A Smith. 2023{\natexlab{a}}.
\newblock How language model hallucinations can snowball.
\newblock \emph{arXiv preprint arXiv:2305.13534}.

\bibitem[{Zhang et~al.(2023{\natexlab{b}})Zhang, Li, Cui, Cai, Liu, Fu, Huang, Zhao, Zhang, Chen, Wang, Luu, Bi, Shi, and Shi}]{zhang_sirens_2023}
Yue Zhang, Yafu Li, Leyang Cui, Deng Cai, Lemao Liu, Tingchen Fu, Xinting Huang, Enbo Zhao, Yu~Zhang, Yulong Chen, Longyue Wang, Anh~Tuan Luu, Wei Bi, Freda Shi, and Shuming Shi. 2023{\natexlab{b}}.
\newblock \href {https://doi.org/10.48550/arXiv.2309.01219} {Siren's {Song} in the {AI} {Ocean}: {A} {Survey} on {Hallucination} in {Large} {Language} {Models}}.
\newblock \emph{arXiv preprint}.
\newblock ArXiv:2309.01219 [cs].

\end{thebibliography}
